\documentclass[12pt,letterpaper,twoside]{article}

    \usepackage{fontspec}
    \defaultfontfeatures{Ligatures=TeX}
    \usepackage[british]{babel}
    \usepackage{xeCJK}
    \setCJKfamilyfont{zhfs}{FandolFang-Regular.otf}
    \usepackage{indentfirst}

    \usepackage{multirow}
    \usepackage{comment}
    \usepackage{lipsum}

    \usepackage[
      includefoot,
      top=2.5cm,
      bottom=2.5cm,
      inner=2.6cm,  
      outer=2.3cm,
      heightrounded     
    ]{geometry}

        \usepackage{fancyhdr}
        \usepackage{lastpage} 
    \usepackage{mathtools}    
    \usepackage{amssymb}      
    \usepackage{amsthm}       
    \usepackage{mathrsfs}     

    \usepackage{algorithm}
    \usepackage{algpseudocode}
    \floatname{algorithm}{Algorithm}
    \algrenewcommand\algorithmicrequire{\textbf{Input:}}
    \algrenewcommand\algorithmicensure{\textbf{Output:}}
    \usepackage{bm}
    
    \usepackage{graphicx}
    \usepackage[font=small,labelfont=bf,labelsep=colon]{caption}
    \usepackage{subcaption} 
    \usepackage{float}

    \usepackage{microtype}
    \microtypesetup{expansion=false,tracking=true,protrusion=true,final}
    \UseMicrotypeSet[protrusion]{basicmath}

    \usepackage[table,dvipsnames]{xcolor}
    \usepackage{hyperref}
    \hypersetup{
        colorlinks,
        linkcolor=RoyalBlue,
        citecolor=MidnightBlue,
        urlcolor=BlueViolet,
        bookmarksopen=true,
        pdftitle={Mamba-Assisted Non-Markovian Closure for Reduced-Order Modeling},
        pdfkeywords={reduced-order modeling, non-Markovian closure, Mori–Zwanzig formalism, selective state-space models (Mamba), data-driven closure modeling},
        pdfsubject={Mamba-based non-Markovian closure modeling for reduced-order systems},
        pdfauthor={Zhi-Feng Wei; Saad Qadeer; Panos Stinis},
    }
    \usepackage{xurl}

    \usepackage{bookmark}

    \usepackage[
        backend = biber,
        style = philosophy-modern,
        isbn = false,
        square = true,
        hyperref=true,
        url = true,
        doi = true,
        parentracker=true,
        shorthandintro = true]{biblatex}    
    \AtEveryBibitem{%
        \clearfield{abstract}
        \clearfield{keywords}
        \clearfield{addendum}
    }
    \DeclareFieldFormat{abstract}{\par\mkbibacro{ABSTRACT}: \small#1}
    \DeclareFieldFormat{addendum}{\par\mkbibacro{Note}: \small#1\nopunct}
    
    \NewBibliographyString{keyword,keywords}
    \DefineBibliographyStrings{english}{%
      keyword  = {keyword},
      keywords = {keywords},
    }
    
    \newcounter{cbx@keyword@total}
    \newcounter{cbx@keyword@count}
    \newcommand*{\keywordscount}[1]{%
      \stepcounter{cbx@keyword@total}}
    
    \newcommand*{\keywordsprint}[1]{%
      \stepcounter{cbx@keyword@count}%
      \ifnumless{\value{cbx@keyword@count}}{2}
        {}
        {\addcomma\space}%
      #1}
    \DeclareFieldFormat{keywords}{%
      \setcounter{cbx@keyword@total}{0}%
      \setcounter{cbx@keyword@count}{0}%
      \forcsvlist{\keywordscount}{#1}%
      \ifnumgreater{\value{cbx@keyword@total}}{1}
        {\par\mkbibacro{keywords}}
        {\par\mkbibacro{keyword}}%
      \addcolon\space
      \forcsvlist{\keywordsprint}{#1}%
    }
    \renewbibmacro*{finentry}{\printfield{abstract}\printfield{keywords}\finentry} 
    
    \DeclareFieldFormat{doi}{%
    \newline
    \mkbibacro{DOI}\addcolon\space
    \ifhyperref
      {\href{https://doi.org/#1}{\nolinkurl{#1}}}
      {\nolinkurl{#1}}
    }

    \DeclareFieldFormat{eprint:arxiv}{%
      \newline
      \mkbibacro{arXiv}\addcolon\space
      \href{https://arxiv.org/abs/#1}{#1}%
      \iffieldundef{eprintclass}{}{\addspace\mkbibbrackets{\printfield{eprintclass}}}%
    }
    \DeclareFieldAlias{eprint:arXiv}{eprint:arxiv} 

    \DeclareFieldFormat{url}{%
    \newline
    \mkbibacro{url}\addcolon\space
    \ifhyperref
      {\href{#1}{\nolinkurl{#1}}}
      {\nolinkurl{#1}}
    }
    
    \renewbibmacro*{doi+eprint+url}{%
      \iffieldequalstr{eprinttype}{arXiv}
        {%
          \iftoggle{bbx:eprint}{\usebibmacro{eprint}}{}%
        }%
        {%
          \printfield{doi}%
          \newunit\newblock
          \iffieldundef{doi}{\usebibmacro{url+urldate}}{}%
        }%
    }

    \usepackage{titlecaps}
    \Addlcwords{on and for a is but with of in as the etc to if its de l' von}
    
    \DeclareFieldFormat[article,inbook,incollection,inproceedings,patent,thesis,unpublished,misc]{titlecase}{\MakeSentenceCase*{#1}}
    \DeclareFieldFormat[misc]{title}{\MakeSentenceCase*{#1}}
    \DeclareFieldFormat[phdthesis,book]{titlecase}{\itshape\titlecap{#1}}
    
    \DeclareFieldFormat{jtu}{\itshape\titlecap{#1}}
    \renewbibmacro*{journal}{%
      \iffieldundef{journaltitle}
        {}
        {\printtext[journaltitle]{%
          \printfield[jtu]{journaltitle}%
          \setunit{\subtitlepunct}%
          \printfield[jtu]{journalsubtitle}}}}
    \DeclareFieldFormat[inproceedings]{booktitle}{
        \printfield[jtu]{booktitle}%
          \setunit{\subtitlepunct}%
          \printfield[jtu]{booksubtitle}
        }

    \NewBibliographyString{artno}
    \DefineBibliographyStrings{english}{artno = {article}}
    \DeclareFieldFormat[article,periodical]{eid}{\bibstring{artno}\addabbrvspace #1}

    \DefineBibliographyStrings{english}{jourvol = {vol\adddot}}
    \DefineBibliographyStrings{english}{number = {issue}}

    \DeclareFieldFormat[article]{volume}{\bibstring{jourvol}\addnbspace #1}
    \DeclareFieldFormat[article]{number}{\bibstring{number}\addnbspace #1}
    \renewbibmacro*{volume+number+eid}{%
      \printfield{volume}%
      \setunit{\addcomma\space}
      \printfield{number}%
      \setunit{\addcomma\space}%
      \printfield{eid}}
    
    \usepackage{cleveref}
    \crefname{equation}{Eq.\@}{Eqs.\@}
    \usepackage{appendix}
    
    \usepackage{cases}

    \theoremstyle{plain}

    \theoremstyle{definition}
    
    \theoremstyle{remark}
    
    \theoremstyle{remark}

    \usepackage{booktabs}

    \DeclarePairedDelimiter\abs{\lvert}{\rvert} 
    \DeclarePairedDelimiter\norm{\lVert}{\rVert} 

    \newcommand{\ue}{\mathrm{e}} 
    
    \newcommand{\ZN}{\mathbb{Z}}
    
    \newcommand{\RNS}{\mathbb{R}}
    
    \newcommand{\ud}{\mathrm{d}} 
    \newcommand{\df}[1]{\,\mathrm{d}#1}

    \usepackage{yfonts} 
    \usepackage{calligra} 

    \usepackage{metalogo} 
    \usepackage[en-US]{datetime2}
    \newcommand{\USdate}{\DTMenglishmonthname{\month}\space\number\day, \number\year}
    \usepackage{fontawesome5} 

    \newcommand\tstamp{\thanks{Complied: \USdate.}}
    
    \usepackage{titling} 
    \pretitle{\begin{center}\LARGE\textbf}
    \posttitle{\end{center}}
    \title{Mamba-Assisted Non-Markovian Closure for Reduced-Order Modeling\tstamp}

    \preauthor{\vskip 1em\begin{center}
    \large \lineskip 0.5em%
    \begin{tabular}[t]{c}}
    \postauthor{\end{tabular}\par\end{center}}
    \usepackage{authblk} 
    \newcommand\corref[2]{\thanks{Email: \href{mailto:#1}{#1}. #2}}

    \author[1]{Zhi-Feng Wei\corref{zfwei@pnnl.gov}{}}
    \author[1]{Saad Qadeer\corref{saad.qadeer@pnnl.gov}{}}
    \author[1,2,3]{Panos Stinis\corref{panagiotis.stinis@pnnl.gov}{}}

    \affil[1]{Advanced Computing, Mathematics, and Data Division,\linebreak
        Pacific Northwest National Laboratory, Richland, WA 99354, USA}
    \affil[2]{Department of Applied Mathematics, University of Washington,
        Seattle, WA 98195, USA}
    \affil[3]{Division of Applied Mathematics, Brown University,
        Providence, RI 02912, USA}

    \predate{}
    \date{}
    \postdate{}

    \providecommand{\keywords}[1]{\flushleft\textbf{\textit{Keywords:}} #1}  
    
    \usepackage{tabularx}
    \newcolumntype{M}{>{$\displaystyle}X<{$}} 
    \usepackage{pdflscape}
    \usepackage{tikz}
    \usetikzlibrary{positioning,arrows.meta,calc}
    \newcommand\res[1]{\widehat{#1}}
    \usepackage{siunitx}
    \usepackage{tablefootnote}
    \usepackage{bbding}

\begin{document}
\pagenumbering{roman}
\setcounter{page}{1}
\maketitle\thispagestyle{empty}
\begin{abstract}
Reduced-order modeling of high-dimensional dynamical systems is often hindered by closure effects arising from unresolved variables, which can introduce non-Markovian dependence into the resolved dynamics.
Motivated by the history-dependent memory term arising in the Mori--Zwanzig formalism, we recast non-Markovian closure modeling as a sequence modeling problem and propose the Mamba-Assisted Closure (MAC) framework.
MAC employs a Mamba-based sequence model to predict the closure from the resolved trajectory and couples the learned closure with the reduced-order governing equations through a numerical integrator to advance the resolved variables in time.
During training, the selective scan mechanism in Mamba enables efficient parallel sequence processing with linear scaling in sequence length, while autoregressive inference proceeds through recurrent state updates at essentially constant per-step cost.
We evaluate MAC on four benchmark systems with complementary characteristics: the viscous Burgers' equation, the chaotic two-scale Lorenz '96 system, the 3-bus DeMarco--Zheng power-grid system, and the dispersive Korteweg--de Vries equation.
Across these benchmarks, MAC consistently improves predictive accuracy and long-time rollout stability relative to the comparison models, demonstrating an effective and computationally scalable approach to non-Markovian closure modeling.


\end{abstract}
\clearpage
\setcounter{tocdepth}{2}
{
\small
\tableofcontents
}
\thispagestyle{empty}
\clearpage
\pagenumbering{arabic}
\setcounter{page}{1}

\clearpage
\section{Introduction}\label{sec_intro}

Many physical systems of scientific and engineering interest, such as turbulent flows, climate dynamics, computational biology, and materials science, are governed by high-dimensional nonlinear dynamics that span a wide range of spatial and temporal scales \autocite{Shady-2021,Christensen-2021,Frezat-2023,Saunders-2013,Giessen-2020}.
Direct numerical simulation of such systems at full resolution remains computationally expensive, motivating the development of reduced-order models (ROMs) that evolve only a subset of the full state variables at reduced cost \autocite{Panos-2024}. 
However, the evolution of the resolved variables is in general not closed: the effects of the unresolved variables feed back onto the resolved dynamics through a closure term, whose accurate modeling is a central challenge in reduced-order modeling \autocite{Williams-2005,Demaeyer-2018,Durbin-2018}.

The Mori--Zwanzig (MZ) formalism provides a principled characterization of this closure term~\autocite{Mori-1965,Zwanzig-1973,Chorin-2000}. 
By projecting the full dynamics onto the resolved subspace, it shows that the closure generally contains a history-dependent contribution that cannot be determined from the instantaneous resolved state alone --- a defining feature of non-Markovian dynamics~\autocite{Parish-2017}.
Physically, information transferred from the resolved to the unresolved scales is not instantaneously lost, but is partially retained and fed back into the resolved dynamics at later times. The Mori--Zwanzig decomposition contains a memory term, which captures this history dependence, together with an orthogonal fluctuation term arising from dynamics intrinsic to the unresolved subspace. The memory term therefore provides a theoretical motivation for using the history of the resolved variables in closure modeling.
In the linear regime, the memory term reduces to a convolution between the resolved trajectory and a memory kernel. 
While conceptually transparent, accurately and efficiently approximating the Mori--Zwanzig memory term remains challenging in practice~\autocite{Zhu-2018,Ma-2019}.
This motivates a data-driven alternative: learning the closure's non-Markovian dependence on the resolved history directly from trajectory data, thereby recasting closure modeling as a sequence modeling problem~\autocite{Gupta-2021,Ma-2019}.

Within this sequence-modeling perspective, existing data-driven approaches differ primarily in how --- and whether --- they represent the closure's dependence on the temporal history.
One line of work couples a neural network with a numerical solver and trains the coupled system by unrolling it over multiple steps, so that the network is optimized against its own rollout trajectory~\autocite{Um-2020,Kochkov-2021}. 
Here, temporal consistency is promoted through the training procedure rather than through an explicit memory of past resolved states.
To capture non-Markovian memory effects explicitly, a second line of work incorporates the temporal history of the resolved variables into the closure. One such approach uses a fixed-length window of past resolved states as input~\autocite{Bhouri-2023}, which turns the reduced dynamics into a delay system but requires the memory depth to be prescribed a priori.
A more flexible alternative employs recurrent neural networks such as LSTMs, which encode the history in an evolving hidden state and have been used to represent closure and memory terms in reduced-order models~\autocite{Wang-2020,Maulik-2020}. 
More recently, an LSTM-based memory model has been coupled with a differentiable physics solver to learn non-Markovian closures for coarse-grained fluid transport~\autocite{Xue-2025}.
Despite their flexibility, recurrent architectures such as LSTMs remain inherently sequential along the temporal dimension during training, which limits parallelism over long trajectories.
Across these approaches to closure modeling, capturing non-Markovian memory without prescribing a fixed history window, while retaining both efficient training over long trajectories and low-cost inference, remains an open challenge.

State-space models offer a natural architectural candidate for addressing this challenge. The output of a structured SSM can be expressed as a discrete convolution of its input with a learnable kernel~\autocite{Gu-2021b,Gu-2021}. This provides a structural correspondence with the Mori--Zwanzig memory integral. Mamba extends structured SSMs with an input-dependent selective mechanism that provides dynamic control over what temporal information is retained and propagated based on the current input, without requiring a prescribed history window~\autocite{Gu-2023}. 
Mamba-based state-space models have already shown strong performance in scientific machine learning settings --- including PDE operator learning~\autocite{Hu-2025a} and dynamical systems modeling~\autocite{Hu-2025b} --- providing additional motivation for their application to the closure modeling problem.

In this work, we propose the Mamba-Assisted Closure (MAC) framework, which recasts non-Markovian closure modeling as a sequence modeling problem.
Motivated by the structural correspondence between SSM convolutions and the Mori--Zwanzig memory integral, we use a Mamba-based sequence model to predict the closure from the temporal history of the resolved trajectory and couple the predicted closure with the reduced-order governing equations through a numerical integrator to advance the resolved variables in time.
During training, the framework exploits the selective scan mechanism in Mamba for efficient parallel sequence processing over long training trajectories with linear scaling in sequence length, while during autoregressive inference it advances through recurrent state updates at essentially constant per-step cost.
We validate the MAC framework on four benchmark systems with complementary characteristics --- the viscous Burgers' equation in Fourier space, the chaotic two-scale Lorenz '96 system, the 3-bus DeMarco--Zheng power-grid system, and the dispersive Korteweg--de Vries equation --- and compare it with Markovian and data-driven closure baselines appropriate to each problem. 
Across these benchmarks, MAC consistently improves predictive accuracy and long-time rollout stability relative to the comparison models.

The remainder of this paper is organized as follows.
\Cref{sec_technical} presents the reduced-order modeling formulation, the associated memory effects, and the MAC framework.
\Cref{sec_results} reports numerical experiments on the viscous Burgers' equation, the two-scale Lorenz '96 system, the 3-bus DeMarco--Zheng power-grid system, and the Korteweg--de Vries equation.
\Cref{sec_conclusion} summarizes the main findings and concludes the paper.
Additional numerical results, computational cost comparisons, implementation details, and a summary of notation are provided in the appendices. 

\section{Technical Approach}\label{sec_technical}
    In this section, we describe the technical framework used for closure modeling in reduced-order models. We first introduce the reduced-order modeling formulation and the associated closure effects arising from unresolved variables. We also discuss why the Mamba-based architecture is well suited for closure modeling with memory effects. 
    We then present the training and inference framework in which sequence models are used to model the closure term and are coupled with the reduced-order governing equations to advance the resolved variables in time and predict their temporal evolution.
    
    \subsection{Reduced-Order Modeling and Memory Effects}\label{sec_rom}
        Many high-dimensional dynamical systems admit effective low-dimensional descriptions, with their essential dynamics evolving on lower-dimensional subspaces despite the large number of degrees of freedom in the full system.
        This observation motivates the construction of reduced-order models (ROMs) that evolve only a subset of the full state variables while aiming to faithfully reproduce the dynamics of interest~\autocite{Panos-2024}.
        
        To formalize this idea, consider a general dynamical system
        \begin{equation}\label{eq_ds}
            \frac{\ud {\varphi}}{\ud {t}} = R(\varphi),
        \end{equation}
        where $\varphi$ denotes the vector of full state variables.
        In reduced-order modeling, we decompose $\varphi$ into resolved and unresolved components,
        \begin{displaymath}
            \varphi = (\widehat{\varphi}, \widetilde{\varphi}),
        \end{displaymath}
        where $\widehat{\varphi}$ collects the variables retained in the ROM and $\widetilde{\varphi}$ collects the degrees of freedom excluded from the reduced-order representation.
        A naive truncation that retains only the interactions among resolved variables yields the approximate dynamics $\df{\widehat{\varphi}}/\df{t} \approx \res{R}(\widehat{\varphi})$. However, this approximation is generally inadequate: the unresolved variables continue to influence the resolved dynamics through their coupling in the original system, and neglecting this influence introduces systematic errors.
        The exact reduced-order dynamics therefore takes the form
        \begin{equation}\label{eq_rom1}
            \frac{\ud {\widehat{\varphi}}}{\ud {t}}
            =
            \res{R}(\widehat{\varphi}) + \mathcal{C},
        \end{equation}
        where $\res{R}(\widehat{\varphi})$ represents the retained interactions among the resolved variables and $\mathcal{C}$ --- the closure term --- represents the net effect of the unresolved variables on the resolved dynamics.
        
        A central question is how to characterize the closure term $\mathcal{C}$. The Mori--Zwanzig formalism provides a principled answer by making the role of memory explicit~\autocite{Mori-1965,Zwanzig-1973,Chorin-2000}. In the full system \Cref{eq_ds}, the resolved and unresolved variables continuously exchange information: information transferred from $\widehat{\varphi}$ to $\widetilde{\varphi}$ is generally not instantaneously lost, but is partially retained by the unresolved variables and subsequently fed back into the resolved dynamics over time. Although $\widetilde{\varphi}$ is not explicitly evolved in the reduced-order model, its influence is therefore implicitly carried forward through memory effects. Consequently, memory effects can make the closure depend not only on the instantaneous state of the resolved variables, but also on their temporal history --- a defining feature of non-Markovian dynamics.
        
        Formally, the Mori--Zwanzig formalism expresses the memory contribution as a functional of the history of the resolved variables. This motivates using the resolved trajectory history to model the closure term,
        \begin{displaymath}
            \mathcal{C}(t)
            \approx
            \mathcal{F}\bigl(\widehat{\varphi}(s):0\leqslant s\leqslant t\bigr).
        \end{displaymath}
        
        In the linear case, the Mori--Zwanzig memory contribution takes the convolution form
        \begin{displaymath}
            \int_{0}^{t}
            K(t-s)\,\widehat{\varphi}(s)\df{s},
        \end{displaymath}
        with $K$ denoting a matrix of memory kernel functions quantifying the influence of past resolved states on the present memory contribution. More generally, the past values of the resolved variables appear in terms of a basis of all functions of the resolved space, e.g., Hermite polynomials, modulated by the memory kernel functions \autocite{Saad-2025}. Dropping all but the linear polynomials then yields the linear regime identified above.
        We note that the full Mori--Zwanzig decomposition also includes an orthogonal fluctuation term arising from dynamics intrinsic to the unresolved subspace.    
        While conceptually transparent, this memory-integral representation is rarely tractable in practice: explicitly estimating the memory kernel and evaluating the resulting convolution integral at every time step can become computationally expensive for long-time rollouts and high-dimensional systems \autocite{Saad-2025}.
           
        These difficulties motivate a data-driven alternative. Rather than estimating the memory kernel explicitly, 
        one may directly approximate the closure term from the temporal history of the resolved variables, thereby recasting closure modeling as a sequence modeling problem. This perspective motivates the sequence-modeling framework developed in the next subsection.
          
    \subsection{Mamba for Non-Markovian Closure Modeling}\label{sec_mamba}
        As discussed in \Cref{sec_rom}, memory effects provide a theoretical motivation for using the temporal history of the resolved variables in closure modeling. The central modeling question is therefore how to use this temporal information to approximate the closure in a way that is both expressive and computationally tractable. In this subsection, we review several candidate approaches and explain why a Mamba-based architecture is particularly well suited to this task.
              
        The simplest baseline is a Markovian, closure-free reduced model, obtained by neglecting the closure term entirely and evolving only the resolved dynamics. This baseline provides a reference for quantifying the effect of unresolved-variable corrections, but does not model either instantaneous or history-dependent closure effects.
        
        A more expressive alternative is to use recurrent neural networks, such as gated recurrent units (GRUs), which encode temporal information in evolving hidden states and can in principle represent non-Markovian dependencies~\autocite{Cho-2014}. 
        However, GRUs remain inherently sequential along the temporal dimension during training, which limits parallelism over long trajectories.
        Learning long-range temporal dependencies with recurrent architectures is also known to pose optimization challenges associated with gradient propagation over extended time horizons~\autocite{Bengio-1994,Pascanu-2013}. 
        These computational and optimization limitations motivate consideration of alternative sequence-modeling architectures that enable efficient training over long trajectories while retaining information from the distant past when it remains dynamically relevant.
        
        State-space models (SSMs) offer an alternative sequence-modeling framework that is particularly well aligned with the closure modeling problem, since they describe sequences through an underlying latent dynamical system --- much like the resolved variables themselves evolve under a (history-dependent) dynamical system~\autocite{Gu-2021b}. A continuous SSM takes the form
        \begin{displaymath}
            \frac{\ud {h}(t)}{\ud {t}} = A h(t) + B x(t),
            \qquad
            y(t) = C h(t).
        \end{displaymath}
        After temporal discretization, this becomes
        \begin{displaymath}
            h_t = \overline{A} h_{t-1} + \overline{B} x_t,
            \qquad
            y_t = \overline{C} h_t,
        \end{displaymath}
        where $x_t$ denotes the input at discrete time step $t$, $h_t$ is the hidden state, $y_t$ is the corresponding output, and $\overline{A}$, $\overline{B}$, $\overline{C}$ are the discretized state-space matrices. This recurrent structure naturally enables temporal information from earlier inputs and states to influence future predictions over long time horizons, making SSMs well suited for reduced-order models with long memory effects.
    
        In structured state-space models such as S4, the matrices $\overline{A}$, $\overline{B}$, $\overline{C}$ are fixed across all time steps~\autocite{Gu-2021}. Under this assumption, the recurrence may be written out explicitly. Starting from $h_0 = \overline{B} x_0$, successive hidden states satisfy
        \begin{displaymath}
            h_1 = \overline{A}\overline{B} x_0 + \overline{B} x_1,
            \quad
            h_2 = \overline{A}^2 \overline{B} x_0 + \overline{A}\overline{B} x_1 + \overline{B} x_2,
            \quad
            \ldots,
        \end{displaymath}
        with corresponding outputs
        \begin{displaymath}
            y_0 = \overline{C}\overline{B} x_0,
            \quad
            y_1 = \overline{C}\overline{A}\overline{B} x_0 + \overline{C}\overline{B} x_1,
            \quad
            y_2 = \overline{C}\overline{A}^2 \overline{B} x_0 + \overline{C}\overline{A}\overline{B} x_1 + \overline{C}\overline{B} x_2,
            \quad
            \ldots
        \end{displaymath}
        More generally,
        \begin{displaymath}
            y_k
            =
            \overline{C}\overline{A}^{k}\overline{B} x_0
            +
            \overline{C}\overline{A}^{k-1}\overline{B} x_1
            +
            \cdots
            +
            \overline{C}\overline{A}\overline{B} x_{k-1}
            +
            \overline{C}\overline{B} x_k.
        \end{displaymath}
        Equivalently, the SSM admits the convolutional representation
        \begin{displaymath}
            y = K * x,
        \end{displaymath}
        with input sequence $x = (x_0, x_1, \ldots, x_L)$, output sequence $y = (y_0, y_1, \ldots, y_L)$, and convolution kernel
        \begin{displaymath}
            K
            =
            \bigl(
            \overline{C}\overline{B},\ 
            \overline{C}\overline{A}\overline{B},\ 
            \ldots,\ 
            \overline{C}\overline{A}^{L}\overline{B}
            \bigr).
        \end{displaymath}
        This convolutional representation closely mirrors the memory-integral structure that arises in the Mori--Zwanzig formalism, in which the memory term depends on the temporal history of the resolved variables through a convolution-like memory operator. The analogy is particularly clear in this convolutional setting: the SSM kernel $K$ plays a role analogous to that of the memory kernel, while the input sequence $x$ corresponds to the history of the resolved variables. This structural parallel provides a principled motivation for using SSM-based sequence models for reduced-order closure modeling, beyond their empirical success on long-sequence tasks.

        However, in conventional SSMs such as S4, the matrices $\overline{A}, \overline{B}, \overline{C}$ remain fixed after training and are shared across all input sequences. Consequently, the same state-space dynamics --- and hence the same convolution kernel --- are applied regardless of the current input, limiting the model's ability to adaptively retain or discard temporal information according to the evolving sequence. To overcome this rigidity, we employ the Mamba architecture for reduced-order closure modeling in this work~\autocite{Gu-2023}. Mamba preserves the SSM framework while introducing an input-dependent selective mechanism, allowing the model to dynamically retain, propagate, or discard temporal information according to the current input. This input dependence prevents the use of the fixed convolutional representation available to S4; instead, Mamba enables parallel sequence processing through the selective scan during training while retaining recurrent state updates during autoregressive inference.
        
        Beyond the selective mechanism, Mamba offers two further properties well suited to closure modeling. First, it does not require a prescribed memory window: information from earlier inputs is carried forward through the recurrent state, while the selective mechanism provides input-dependent control over how temporal information is retained and propagated. Second, Mamba scales linearly in sequence length, enabling efficient processing of the long temporal trajectories that arise in closure modeling problems with extended memory effects.

        Motivated by these properties, we employ Mamba-based sequence models to 
        predict the closure term from the temporal history of the resolved variables.

    \subsection{Training and Inference Framework}\label{sec_framework}

        Having motivated the use of Mamba-based sequence models for closure modeling, we now describe how such models are trained and subsequently coupled with the reduced-order system during inference to evolve the resolved variables. The framework presented below is applied consistently to all sequence model architectures considered in this work, including both Mamba-based and GRU-based sequence models; their detailed architectures are deferred to \Cref{apdx_neural_network_architectures}.
        
        Recall from \Cref{sec_rom} that the reduced-order system takes the form
        \begin{equation}\label{eq_rom2}
            \frac{\ud\widehat{\varphi}}{\ud t}
            =
            \res{R}(\widehat{\varphi}) + \mathcal{C},
        \end{equation}
        where $\widehat{\varphi}$ denotes the resolved variables, $\res{R}$ represents the retained interactions among them and is typically known from the full dynamical system \Cref{eq_ds}, and $\mathcal{C}$ denotes the unknown closure term associated with the unresolved variables. The role of the sequence model is to approximate $\mathcal{C}$ using the history of $\widehat{\varphi}$, so that the resulting closed system can be advanced in time to predict the evolution of the resolved variables.

        During training, the sequence model is supplied with ground-truth resolved trajectories and learns to predict the corresponding closure terms from their temporal histories. Specifically, given a resolved trajectory
        \begin{displaymath}
            \{\widehat{\varphi}_0,\widehat{\varphi}_1,\dots,\widehat{\varphi}_T\},
        \end{displaymath}
        the sequence model produces the associated trajectory of closure terms
        \begin{displaymath}
            \{\mathcal{C}_0,\mathcal{C}_1,\dots,\mathcal{C}_T\}.
        \end{displaymath}
        For most experiments, the model is trained directly against the reference closure trajectories under teacher forcing, with the ground-truth resolved trajectory supplied as the input sequence. For the Lorenz '96 system, we instead employ a multi-step resolved-variable rollout loss, as described in \Cref{sec_results_l96} and \Cref{apdx_training_losses}.
        
\begin{figure}[t]
    \centering
    \begin{tikzpicture}
        \colorlet{resolvedcolor}{blue!50!black}
        \colorlet{closurecolor}{red!60!black}
        \colorlet{integratorcolor}{black!70}

        \node[
            draw=resolvedcolor,
            fill=blue!5,
            rounded corners,
            thick,
            minimum width=1.8cm,
            minimum height=0.9cm
        ] (phi0) {$\widehat{\varphi}_0$};

        \node[
            draw=resolvedcolor,
            fill=blue!5,
            rounded corners,
            thick,
            minimum width=1.8cm,
            minimum height=0.9cm,
            right=2.5cm of phi0
        ] (phi1) {$\widehat{\varphi}_1$};

        \node[
            draw=resolvedcolor,
            fill=blue!5,
            rounded corners,
            thick,
            minimum width=1.8cm,
            minimum height=0.9cm,
            right=2.5cm of phi1
        ] (phi2) {$\widehat{\varphi}_2$};

        \node[
            right=2.5cm of phi2,
            scale=1.6
        ] (dots1) {$\cdots$};

        \node[
            draw=closurecolor,
            fill=red!5,
            rounded corners,
            thick,
            minimum width=1.8cm,
            minimum height=0.9cm,
            below=1.8cm of phi0
        ] (c0) {$\mathcal{C}_0$};

        \node[
            draw=closurecolor,
            fill=red!5,
            rounded corners,
            thick,
            minimum width=1.8cm,
            minimum height=0.9cm,
            below=1.8cm of phi1
        ] (c1) {$\mathcal{C}_1$};

        \node[
            draw=closurecolor,
            fill=red!5,
            rounded corners,
            thick,
            minimum width=1.8cm,
            minimum height=0.9cm,
            below=1.8cm of phi2
        ] (c2) {$\mathcal{C}_2$};

        \node[
            below=1.8cm of dots1,
            scale=1.6
        ] (dots2) {$\cdots$};

        \node[
            draw=integratorcolor,
            fill=black!5,
            circle,
            thick,
            inner sep=1pt,
            minimum size=0.75cm
        ] (int01) at ($(phi0)!0.5!(phi1) + (0,-0.6)$) {\scriptsize RK4};

        \node[
            draw=integratorcolor,
            fill=black!5,
            circle,
            thick,
            inner sep=1pt,
            minimum size=0.75cm
        ] (int12) at ($(phi1)!0.5!(phi2) + (0,-0.6)$) {\scriptsize RK4};

        \node[
            draw=integratorcolor,
            fill=black!5,
            circle,
            thick,
            inner sep=1pt,
            minimum size=0.75cm
        ] (int23) at ($(phi2)!0.5!(dots1) + (0,-0.6)$) {\scriptsize RK4};

        \draw[->, thick, closurecolor]
            (phi0) -- node[left] {\small Sequence model} (c0);
        \draw[->, thick, closurecolor] (phi1) -- (c1);
        \draw[->, thick, closurecolor] (phi2) -- (c2);
        \draw[->, thick, closurecolor] (dots1) -- (dots2);

        \draw[->, thick, integratorcolor] (phi0) -- (int01);
        \draw[->, thick, integratorcolor] (phi1) -- (int12);
        \draw[->, thick, integratorcolor] (phi2) -- (int23);

        \draw[->, thick, integratorcolor]
            (c0.north east) to[out=30,in=-110] (int01.south);
        \draw[->, thick, integratorcolor]
            (c1.north east) to[out=30,in=-110] (int12.south);
        \draw[->, thick, integratorcolor]
            (c2.north east) to[out=30,in=-110] (int23.south);

        \draw[->, thick, integratorcolor]
            (int01) -- node[above, sloped] {\scriptsize\itshape} (phi1);
        \draw[->, thick, integratorcolor] (int12) -- (phi2);
        \draw[->, thick, integratorcolor] (int23) -- (dots1);

    \end{tikzpicture}

    \caption{
    Schematic illustration of the autoregressive rollout procedure for reduced-order model evolution with predicted closure terms. At each step, the sequence model predicts the closure term $\mathcal{C}_t$ from the current resolved state $\widehat{\varphi}_t$ (red arrows) together with the recurrent state carrying the preceding history. The resolved state is then advanced by one integrator step that takes both $\widehat{\varphi}_t$ and $\mathcal{C}_t$ as inputs to produce $\widehat{\varphi}_{t+1}$ (gray arrows). See \Cref{apdx_rk4_zero_order_holding} for details of the time-stepping procedure.
    }

    \label{fig_autoregressive_rollout}
\end{figure}
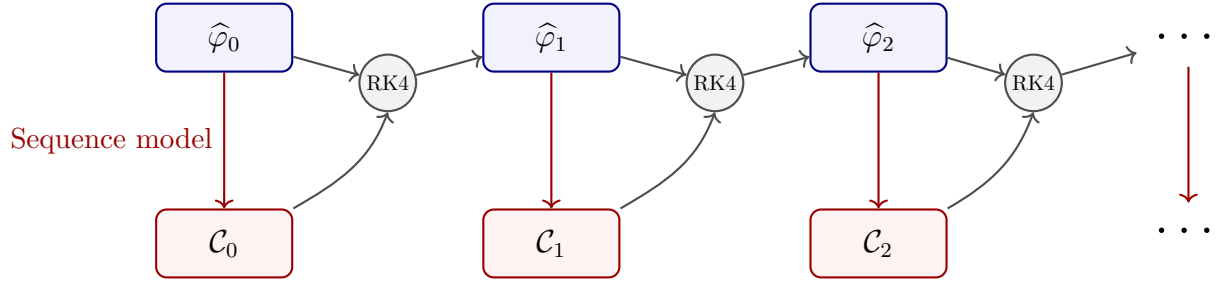

        During inference, in contrast, the resolved and closure trajectories are advanced autoregressively in tandem with the reduced-order dynamics. Starting from an initial resolved state $\widehat{\varphi}_0$, the sequence model first predicts the corresponding closure term $\mathcal{C}_0$. The reduced-order system \Cref{eq_rom2} is then advanced by one numerical integrator step that takes both $\widehat{\varphi}_0$ and $\mathcal{C}_0$ as inputs, yielding the next resolved state $\widehat{\varphi}_1$. This updated state is fed back into the sequence model to predict $\mathcal{C}_1$, which together with $\widehat{\varphi}_1$ drives the next integrator step to produce $\widehat{\varphi}_2$, and so on. The procedure is illustrated in \Cref{fig_autoregressive_rollout}. Iterating in this manner rolls out the full resolved trajectory, with both the predicted resolved states and the predicted closure terms recorded for subsequent evaluation. In our implementation, the known reduced-order dynamics are advanced using a fourth-order Runge--Kutta scheme, while the predicted closure is evaluated once at the beginning of each time step and held fixed over the Runge--Kutta stages; details are provided in \Cref{apdx_rk4_zero_order_holding}.
        
        A practical advantage of this autoregressive procedure is that, at each step, only the current resolved state $\widehat{\varphi}_t$ needs to be supplied to the sequence model: the temporal history of the resolved variables is implicitly carried in the recurrent state. Recurrent sequence models such as Mamba and GRU therefore enable efficient autoregressive rollout without reprocessing the entire trajectory at every step. In contrast, Transformer-based architectures must apply attention over an ever-growing input sequence during rollout, incurring substantially higher computational cost for long trajectories~\autocite{Vaswani-2017}.
        
        Notably, this training--inference design aligns naturally with the computational structure of state-space models discussed in \Cref{sec_mamba}. During training, the sequence model is applied in a sequence-to-sequence fashion, allowing an entire resolved trajectory to be processed in parallel. For conventional structured SSMs such as S4, this parallelism can be achieved through the convolutional representation, whereas Mamba employs the selective scan to accommodate its input-dependent state-space parameters. During inference, by contrast, the model is advanced step by step in tandem with the reduced-order dynamics through recurrent state updates, which provide essentially constant per-step cost and avoid reprocessing the entire temporal history.        
        This combination of parallel sequence processing during training and recurrent state updates during inference is well suited for achieving efficient training and low-cost autoregressive rollout in reduced-order closure modeling.   
        

        We refer to the resulting framework --- in which a Mamba-based sequence model is coupled with the reduced-order dynamics to perform autoregressive closure prediction and trajectory evolution --- as the Mamba-Assisted Closure (MAC) framework, and use \emph{MAC model} interchangeably to denote the resulting closed reduced-order system. The MAC framework serves as the basis for all numerical experiments in the remainder of this paper. 

\clearpage
\section{Numerical Results}\label{sec_results}

In this section, we report our experimental results on the viscous Burgers' equation, the two-scale Lorenz '96 system, the 3-bus DeMarco--Zheng power-grid system, and the Korteweg--de Vries (KdV) equation.
These problems provide complementary settings for closure modeling, including dissipative and dispersive PDE dynamics, chaotic multiscale dynamics, and a system with long-memory effects.
For Burgers' equation, the KdV equation, and the power-grid system, we compare the MAC model with the corresponding Markovian closure-free reduced-order model without closure correction and the GRU-based sequence model.
For the Lorenz '96 system, we compare the MAC model with the GRU-based sequence model and the classical Wilks method~\autocite{Cho-2014,Wilks-2005}.
For Burgers' equation, we additionally compare MAC with the linear and cubic memory models in~\autocite{Saad-2025} for structured out-of-distribution initial conditions. Throughout this section, we focus primarily on the resolved variables; additional results and plots for the closure terms are reported in \Cref{apdx_closure_plots}. Implementation details are provided in \Cref{apdx_implementation}.

    \subsection{Viscous Burgers' Equation}\label{sec_results_burgers}
        Consider the viscous Burgers' equation with periodic boundary conditions
            \begin{displaymath}
            u_t = \partial_x\Bigl(-\frac{1}{2}u^2\Bigr)+\nu u_{xx},
            \qquad
            u(0,x)=u_0(x),
            \qquad
            x\in[0,2\pi).
            \end{displaymath}
        Throughout our experiments, the viscosity coefficient is set to $\nu=0.1$.
        Expanding the solution $u(t,x)$ in terms of Fourier modes
            \begin{displaymath}
            u(t,x)=\sum_{k\in\ZN}\theta_k(t)e^{ikx},
            \end{displaymath}
        we can rewrite the viscous Burgers' equation as a system of ODEs
            \begin{equation}\label{eq_burgers_resolved}
            \theta'_k(t)
            =
            -\frac{ik}{2}
            \sum_{\substack{p,q\in\mathbb{Z}\\ p+q=k}}
            \theta_p(t)\theta_q(t)
            -
            \nu k^2\theta_k(t)
            \eqcolon
            R_k\bigl(\theta(t)\bigr), \quad\text{for\quad}  k\in\ZN.
            \end{equation}
        Here
            \begin{displaymath}
            \theta_k(0)=\bigl(\widehat{u_0}\bigr)_k
            =
            \frac{1}{2\pi}
            \int_0^{2\pi}e^{-ikx}u_0(x)\df{x},
            \end{displaymath}
        and the reality of $u$ implies $\theta_{-k}(t)=\overline{\theta_k(t)}$ for $k\in\ZN$, so that the negatively indexed Fourier modes are completely determined by the positively indexed ones.
        Since $\theta'_0(t)=0$, the zero Fourier mode remains constant in time. Therefore, without loss of generality, we restrict attention to problems with a vanishing zero Fourier mode, i.e., $\theta_0(0)=0$, so that $\theta_0(t)=0$ for all nonnegative $t$.
        To construct a reduced-order system, we retain the first $M$ positively indexed complex Fourier modes as resolved modes and treat the remaining positively indexed modes as unresolved. The corresponding negatively indexed modes are determined by the reality constraint.
        For notational convenience, we also represent the complex Fourier modes in real coordinates and write
            \begin{displaymath}
                \theta_k(t)=\phi_k(t)+i\psi_k(t), \quad\text{for\quad}  k\in\ZN.
            \end{displaymath}
        In our experiments, we retain $M=3$ positively indexed complex Fourier modes as resolved variables, corresponding to $6$ real variables.
        
        The evolution of the resolved modes alone is not closed, since the unresolved Fourier modes continue to influence the low-frequency dynamics through nonlinear interactions. Following our discussion in \Cref{sec_technical}, this unresolved contribution is represented by a closure term. For each resolved mode $k=1,2,\ldots,M$, we define
            \begin{displaymath}
            \mathcal{M}_k(t)
            =
            R_k\bigl(\theta(t)\bigr)
            -
            R_k\bigl(\widetilde{\theta}(t)\bigr),
            \end{displaymath}
        where $\widetilde{\theta}(t)$ denotes the truncated Fourier state obtained by retaining only the resolved positively indexed modes and enforcing the corresponding reality constraint. The first term $R_k(\theta(t))$ represents the exact evolution of the resolved mode computed from the full Fourier state, while the second term $R_k(\widetilde{\theta}(t))$ corresponds to the reduced dynamics obtained after truncation in the reduced-order model. We further decompose the closure term into real and imaginary parts,
            \begin{displaymath}
            \mathcal{M}_k(t)
            =
            m_k^\phi(t)+i m_k^\psi(t).
            \end{displaymath}

        In our numerical experiments, the sequence models are trained to learn
            \begin{displaymath}
            \{m_k^\phi,m_k^\psi\}_{k=1}^M
            \end{displaymath}
        from the trajectory of the real representation
            \begin{displaymath}
            \{\phi_k,\psi_k\}_{k=1}^M
            \end{displaymath}
        of the resolved Fourier modes. During inference, as described in \Cref{sec_framework}, the trained sequence model is coupled with the reduced-order dynamics through an autoregressive rollout procedure to evolve the resolved modes from the initial condition.
        
        In data preparation, the initial conditions are sampled in the resolved Fourier modes by drawing both $\phi_k(0)$ and $\psi_k(0)$ uniformly from the interval $(-\ue^{-k},\ue^{-k})$ for
        $k=1, 2, \ldots, M$. The negative Fourier modes are determined by conjugate symmetry, while the zero mode and all higher modes are initially set to zero.
        To prepare training data, Burgers' equation is first solved on the time interval $[0,1]$ using a pseudospectral method and a fourth-order Runge--Kutta integrator with time step $\Delta t=10^{-4}$.
        During training, each sequence model directly predicts the closure terms from the resolved trajectories, and the loss is computed as the mean-squared error between the predicted and reference closures. 
        
        We now present the experimental results for the MAC model and compare its predictive performance with that of the corresponding Markovian reduced-order model without closure correction and that of the GRU-based sequence model.
        
\begin{figure}[htbp]
            \centering
            \includegraphics[width=16cm]{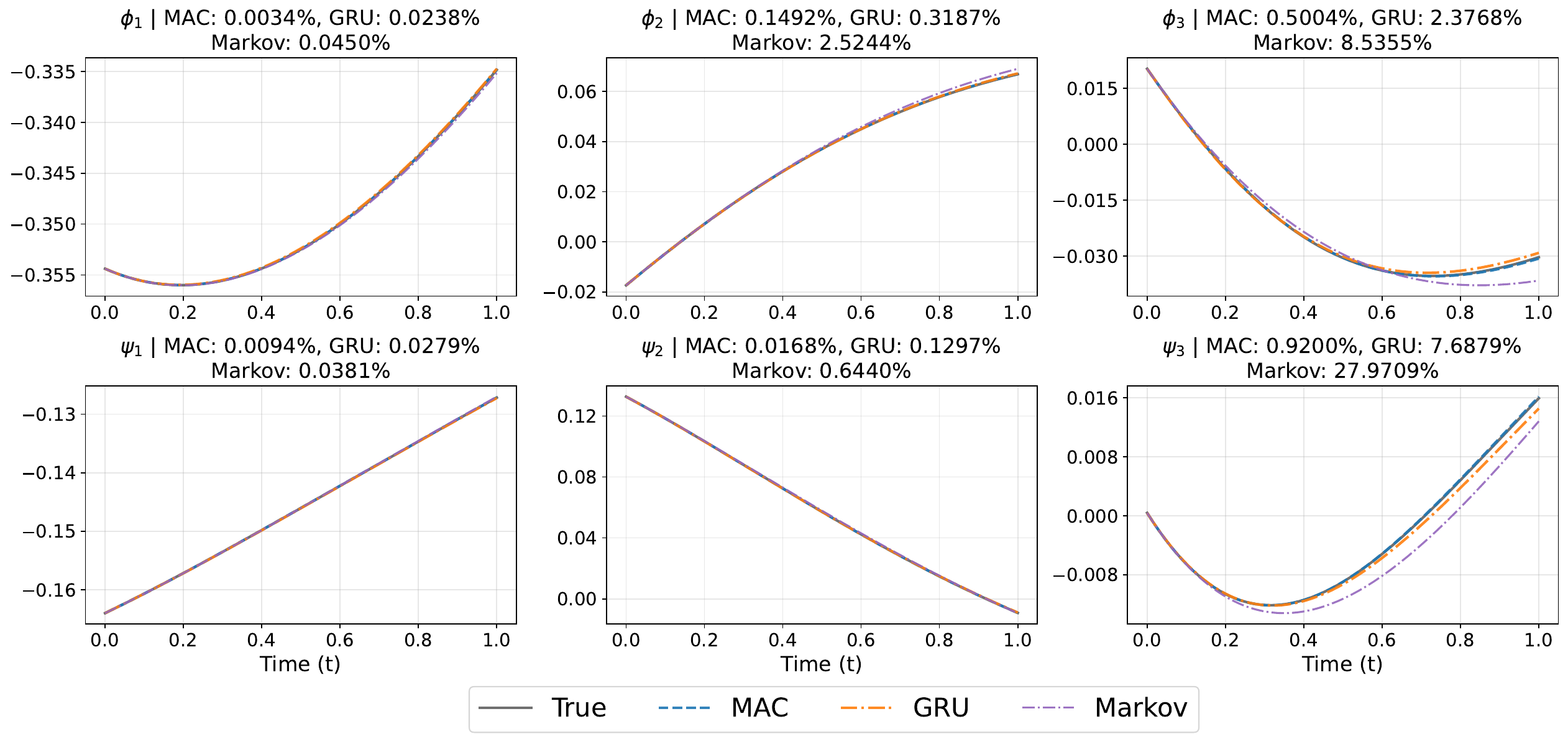}
            \caption{\label{fig_burgers_interp_resolved}Comparison of resolved-mode predictions from the MAC model, the GRU-based model, and the Markovian reduced-order model on one representative test initial condition over the temporal interpolation regime $[0,1]$. For each resolved Fourier mode, the relative $L^2$ error over the time interval $[0,1]$ is also reported.}
        \end{figure}

        We first examine the prediction accuracy within the training time interval $[0,1]$, corresponding to a temporal interpolation regime.
        \Cref{fig_burgers_interp_resolved} compares the predicted trajectories of the resolved Fourier modes against the true solutions for one representative test initial condition.
        From \Cref{fig_burgers_interp_resolved}, we observe that the MAC model matches the true solution more closely than the GRU-based sequence model and the Markovian model, especially for higher-frequency modes. The mean relative $L^2$ error for each Fourier mode across all initial conditions in the test dataset, reported in \Cref{tab_burgers_interp_resolved}, further confirms this observation. 
        Both the mean relative $L^2$ error and the corresponding standard deviation are roughly one order of magnitude smaller for the MAC model than for the Markovian model, indicating substantially improved prediction accuracy.
        The MAC model also yields lower errors than the GRU-based sequence model for all resolved Fourier modes.
\begin{table}[htbp]
    \centering
    \caption{\label{tab_burgers_interp_resolved}Relative $L^2$ error statistics for each resolved Fourier mode over the temporal interpolation regime $[0,1]$. The mean and standard deviation are computed across all initial conditions in the test dataset and over 5 random seeds for the MAC model, the GRU-based model, and the Markovian reduced-order model.}
    \begin{tabular}{c|ccc}
    \hline
    Mode & MAC (\%) & GRU (\%) & Markov (\%) \\
    \hline
    $\phi_1$ & $0.0184 \pm 0.0294$ & $0.0253 \pm 0.0393$ & $0.1113 \pm 0.1663$ \\
    $\phi_2$ & $0.1574 \pm 0.1638$ & $0.2187 \pm 0.2240$ & $1.3557 \pm 0.8234$ \\
    $\phi_3$ & $1.0438 \pm 1.0495$ & $1.7570 \pm 1.6960$ & $12.1529 \pm 8.1459$ \\
    \hline
    $\psi_1$ & $0.0270 \pm 0.0433$ & $0.0381 \pm 0.0688$ & $0.1451 \pm 0.2384$ \\
    $\psi_2$ & $0.1384 \pm 0.1338$ & $0.2251 \pm 0.2371$ & $1.3002 \pm 1.3278$ \\
    $\psi_3$ & $1.4144 \pm 1.4368$ & $2.2016 \pm 2.8572$ & $13.2198 \pm 11.6266$ \\
    \hline
    \end{tabular}
\end{table}

\begin{figure}[htbp]
                \centering
                \includegraphics[width=16cm]{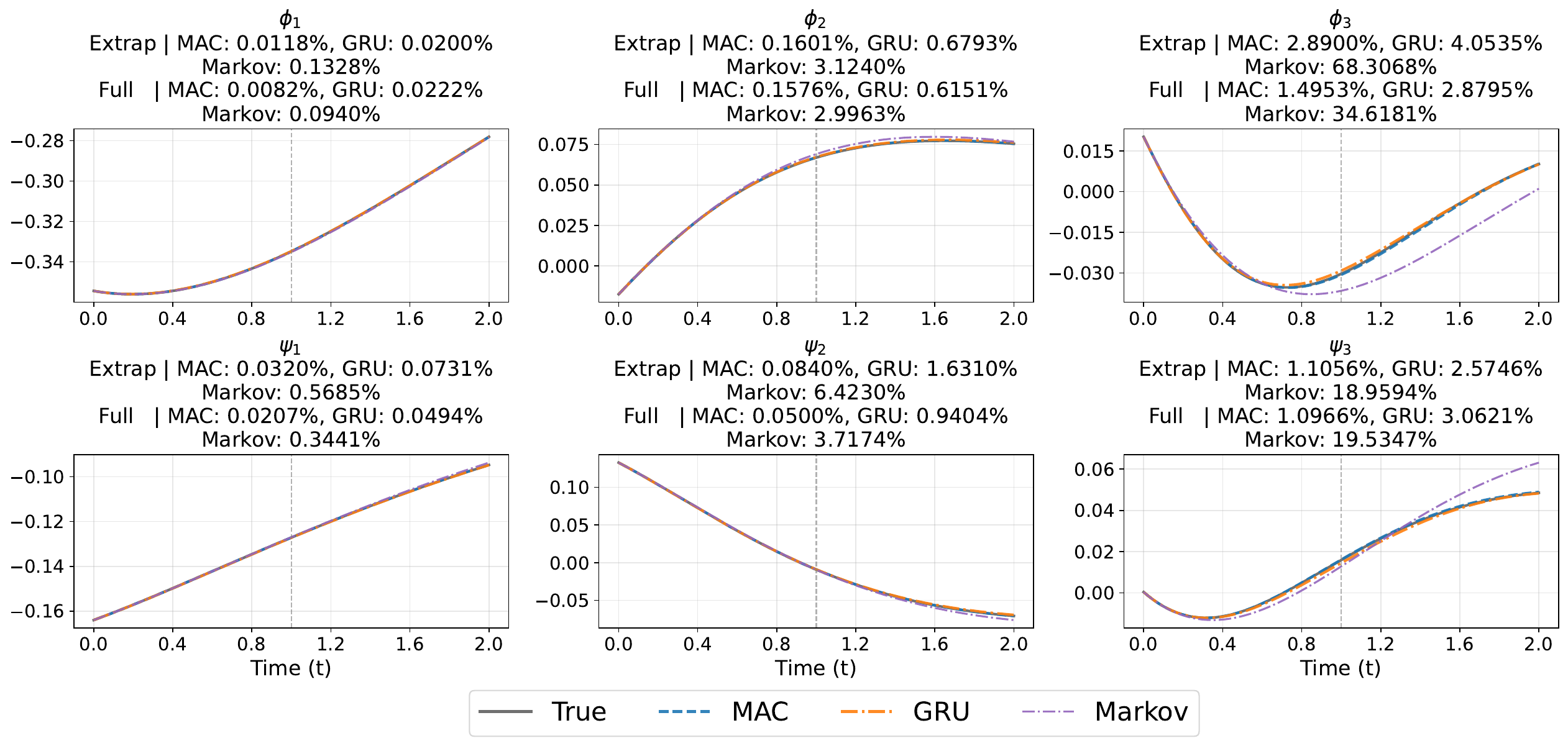}
                \caption{\label{fig_burgers_extrap_resolved}Comparison of resolved-mode predictions from the MAC model, the GRU-based model, and the Markovian reduced-order model on the same representative test initial condition as in \Cref{fig_burgers_interp_resolved} over the temporal extrapolation regime $[0,2]$. For each resolved Fourier mode, the relative $L^2$ errors over the intervals $[1,2]$ and $[0,2]$ are also reported.}
            \end{figure}

        We then examine the prediction accuracy over the longer time interval $[0,2]$. Recall that the MAC and GRU-based models are trained using data only over the time interval $[0,1]$; therefore, this test corresponds to a temporal extrapolation regime. \Cref{fig_burgers_extrap_resolved} compares the predicted trajectories of the resolved Fourier modes against the true solutions on the same representative test initial condition as in \Cref{fig_burgers_interp_resolved}.
        From \Cref{fig_burgers_extrap_resolved}, we observe that the MAC model continues to match the true solution closely even on the temporal extrapolation interval $[1,2]$, while the GRU-based model shows moderate degradation and the Markovian reduced-order model rapidly deteriorates and deviates significantly from the true dynamics. The mean relative $L^2$ error for each Fourier mode across all initial conditions in the test dataset, reported in \Cref{tab_burgers_extrap_resolved}, further confirms this observation. Compared with the temporal interpolation regime, all three models exhibit increased prediction error on the extrapolation interval $[1,2]$, reflecting the accumulation of rollout error outside the training horizon. Nevertheless, the MAC model maintains stable long-time prediction accuracy across all resolved modes, with the GRU-based model producing roughly twice the error of the MAC model, while the Markovian reduced-order model suffers substantial error growth during extrapolation. The performance gap becomes increasingly pronounced for higher-frequency resolved modes. In particular, for the third resolved Fourier mode on the extrapolation interval $[1,2]$, the mean relative $L^2$ error of the Markovian model reaches approximately $30\%$ to $40\%$, the GRU-based model yields errors around $4\%$ to $5\%$, while the corresponding error produced by the MAC model remains around $2\%$ to $3\%$. 
        This observation suggests that unresolved nonlinear interactions become increasingly important for higher-frequency resolved dynamics, where the MAC model maintains more accurate long-time predictions.
        Although the GRU-based model also incorporates temporal information through its hidden state, its prediction errors remain larger than those of the MAC model, particularly for higher-frequency modes and longer rollout horizons.
        
\begingroup
\renewcommand{\thefootnote}{\FiveStarOpen}
\begin{table}[htbp]
\centering
\caption{\label{tab_burgers_extrap_resolved}
Relative $L^2$ error statistics for each resolved Fourier mode over the temporal extrapolation regime. Results are reported on both the full rollout interval $[0,2]$ and the pure extrapolation interval $[1,2]$. The mean and standard deviation are computed across all initial conditions in the test dataset and over 5 random seeds for the MAC model, the GRU-based model, and the Markovian reduced-order model.
}

\resizebox{\textwidth}{!}{
\begin{tabular}{c|ccc|ccc}
\hline
\multirow{2}{*}{Mode}
& \multicolumn{3}{c|}{$[0,2]$}
& \multicolumn{3}{c}{$[1,2]$} \\
& MAC (\%) & GRU (\%) & Markov (\%)
& MAC (\%) & GRU (\%) & Markov (\%) \\
\hline

$\phi_1$
& $0.0373 \pm 0.0491$
& $0.0533 \pm 0.0645$
& $0.2590 \pm 0.4670$
& $0.0508 \pm 0.0641$
& $0.0736 \pm 0.0834$
& $0.3463 \pm 0.5644$ \\

$\phi_2$
& $0.3023 \pm 0.2978$
& $0.4911 \pm 0.4762$
& $2.4248 \pm 1.7619$
& $0.5630 \pm 0.7592$
& $0.9224 \pm 1.1639$
& $4.0607 \pm 3.7503$ \\

$\phi_3$
& $1.5970 \pm 1.2594$
& $2.9286 \pm 2.1488$
& $24.9455 \pm 16.1813$
& $2.4487 \pm 2.4552$
& $4.5312 \pm 4.2860$
& $38.1344 \pm 30.1155$\\

\hline

$\psi_1$
& $0.0558 \pm 0.0772$
& $0.0840 \pm 0.1148$
& $0.4131 \pm 0.6710$
& $0.0810 \pm 0.1154$
& $0.1244 \pm 0.1773$
& $0.6897 \pm 1.2885$ \\

$\psi_2$
& $0.3139 \pm 0.4123$
& $0.5564 \pm 0.6316$
& $3.7221 \pm 4.6915$
& $0.8779 \pm 2.8116$
& $1.3400 \pm 3.0364$
& $10.8559 \pm 34.1128$\tablefootnote{
The slightly elevated standard deviation for $\psi_2$ on $[1,2]$ is unsurprising: it stems from a single outlier test case in which the true resolved trajectory happens to decay close to zero over this interval. Because the relative $L^2$ error is normalized by the magnitude of the true trajectory, a near-vanishing denominator on this particular case yields a relative error far exceeding that of the remaining test cases. As the standard deviation is highly sensitive to such a single extreme value among an otherwise tightly clustered set of errors, this one outlier alone is sufficient to inflate the standard deviation beyond the mean, even though it does not reflect a general lack of robustness of the underlying models.}\\

$\psi_3$
& $1.8764 \pm 1.7066$
& $3.3037 \pm 2.8738$
& $21.5987 \pm 14.2675$
& $2.7496 \pm 2.9486$
& $5.1749 \pm 5.3727$
& $31.6570 \pm 27.4063$ \\

\hline
\end{tabular}
}
\end{table}
\endgroup
\addtocounter{footnote}{-1}

\begin{figure}[htbp]
                \centering
                \includegraphics[width=9cm]{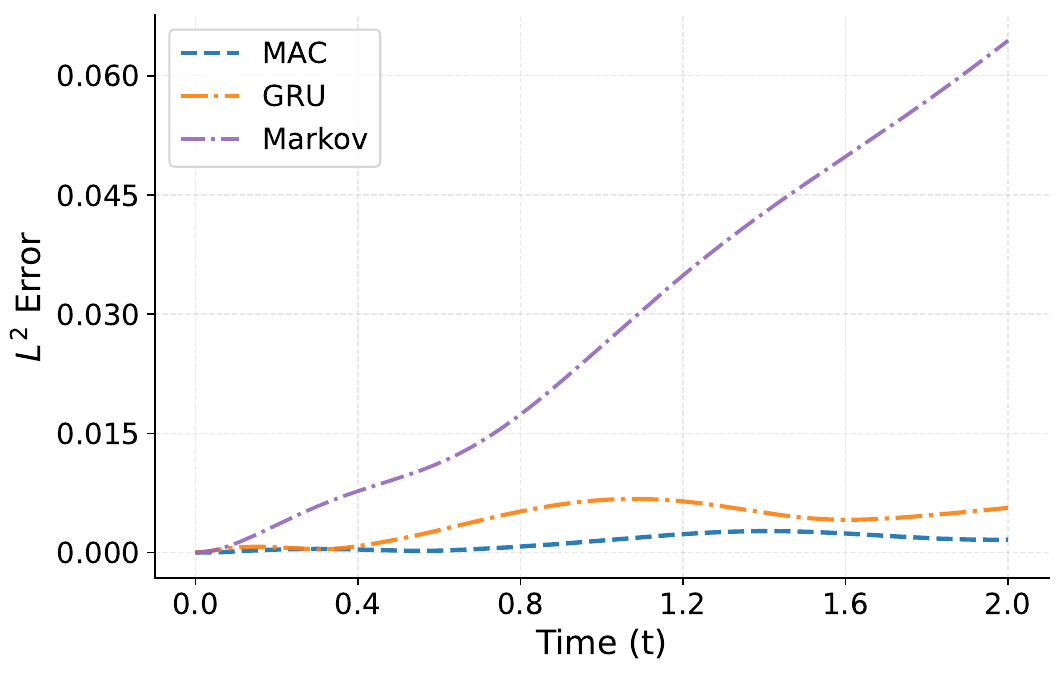}
                    \caption{\label{fig_burgers_extrap_l2}Evolution of the physical-space $L^2$ error over the temporal extrapolation interval $[0,2]$ for the same representative test initial condition as in \Cref{fig_burgers_interp_resolved}, comparing the MAC model, the GRU-based model, and the Markovian reduced-order model. }
            \end{figure}

        We further compare the evolution of the absolute $L^2$ error in physical space over the entire extrapolation interval $[0,2]$. At each time instant, the $L^2$ error on the physical-space interval $x\in[0,2\pi)$ is computed between the reconstructed physical-space field obtained from the resolved Fourier modes and the corresponding true resolved field extracted from the full-order simulation. As shown in \Cref{fig_burgers_extrap_l2}, the error produced by the MAC model remains consistently small throughout the rollout, whereas the error produced by the Markovian reduced-order model grows rapidly over time. This further demonstrates the effectiveness of the learned closure in improving the reduced-order dynamics.

        Recall that both the training and test datasets discussed above are generated from random low-frequency Fourier initial conditions, where the resolved coefficients $\phi_k(0)$ and $\psi_k(0)$ are independently sampled from the interval $(-\ue^{-k},\ue^{-k})$ for $k=1,2,\ldots,M$. Although the temporal extrapolation results above already demonstrate strong long-time stability of the MAC closure model, the corresponding initial conditions still lie within the same distributional setting as the training data.
        To further investigate out-of-distribution generalization, we consider two structured initial conditions that lie outside the training distribution: \(u_0(x)=\sin x\) and a projected \(e^{\sin x}\) initial condition. For the latter, only the first three positive Fourier modes and their corresponding negative modes are retained, with the zero Fourier mode set to zero.
            
        These two initial conditions lie outside the training distribution. In particular, selected resolved Fourier coefficients exceed the corresponding training bounds, as summarized in \Cref{tab_ood_fourier}.
\begin{table}[htbp]
    \centering
    \caption{\label{tab_ood_fourier}Comparison between selected resolved Fourier coefficients of the out-of-distribution test initial conditions and the corresponding coefficient bounds of the training initial condition distribution.}
    \begin{tabular}{cccc}
    \toprule
    Initial condition & Resolved Fourier coefficient & Training bound & Ratio to training bound \\
    \midrule
    $\sin x$
    & $|\psi_1(0)|=0.500$
    & $0.368$
    & $1.36\times$ \\

    projected $\ue^{\sin x}$
    & $|\psi_1(0)|=0.565$
    & $0.368$
    & $1.54\times$ \\

    projected $\ue^{\sin x}$
    & $|\phi_2(0)|=0.136$
    & $0.135$
    & $1.01\times$ \\
    \bottomrule
    \end{tabular}
\end{table}

\begin{figure}[htbp]
                \centering
                \includegraphics[width=10cm]{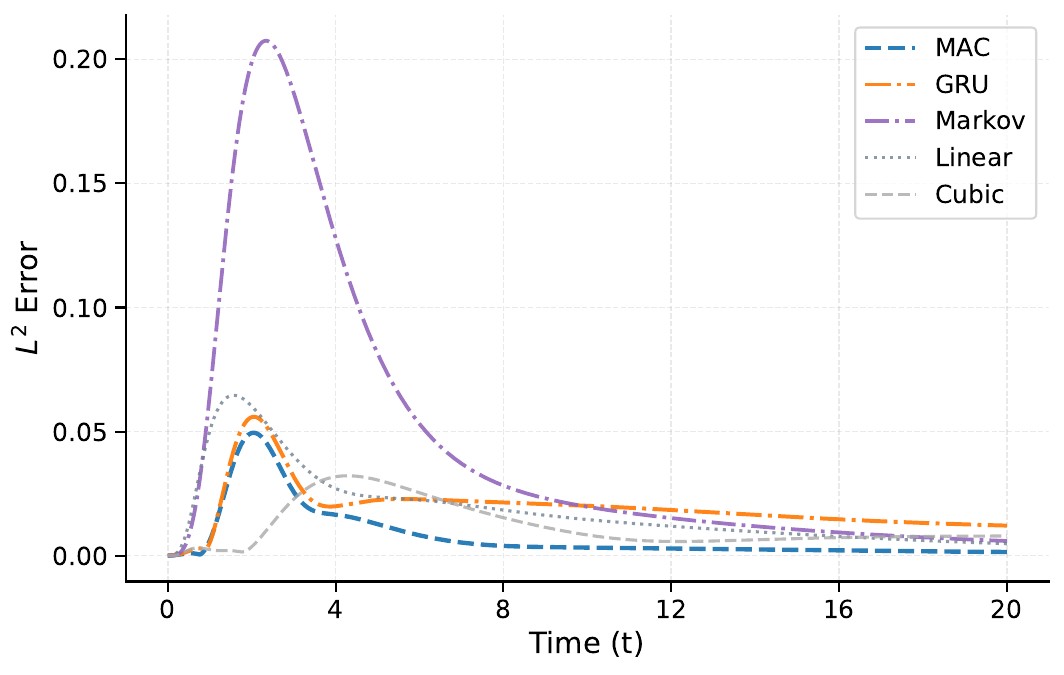}
                \caption{\label{fig_burgers_ood_case0}Evolution of the physical-space $L^2$ error for the out-of-distribution initial condition $u_0(x)=\sin x$ over the long-time rollout interval $[0,20]$, comparing the MAC model, the GRU-based model, the Markovian reduced-order model, and the linear and cubic memory models in~\autocite{Saad-2025}.}
            \end{figure}

\begin{figure}[htbp]
                \centering
                \includegraphics[width=10cm]{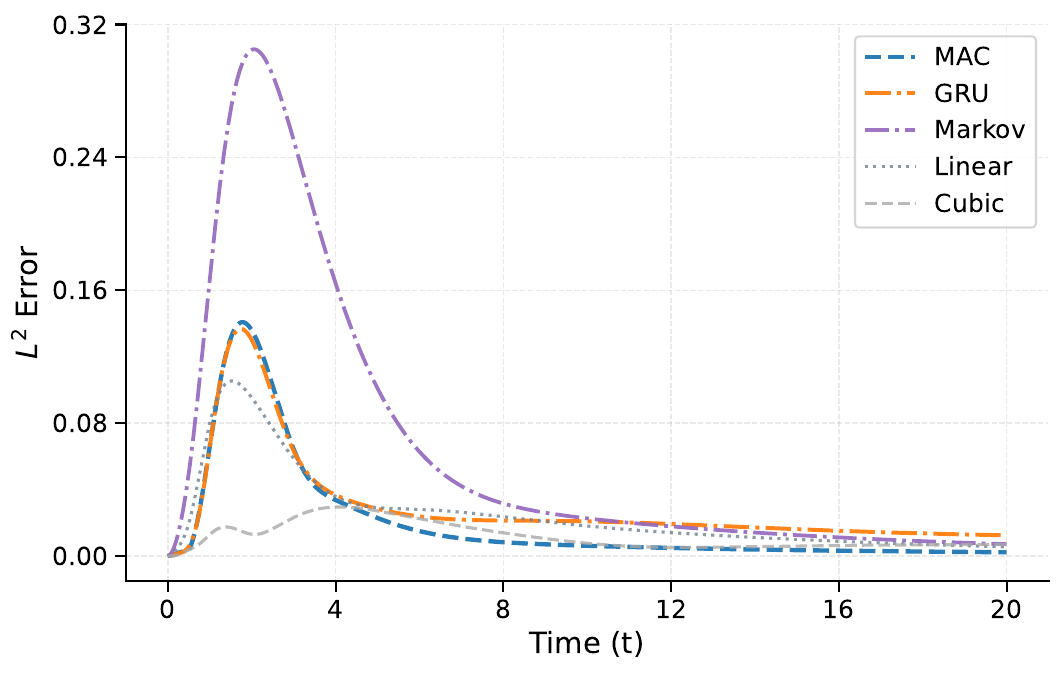}
                \caption{\label{fig_burgers_ood_case1}Evolution of the physical-space $L^2$ error 
                for the projected $\ue^{\sin x}$ out-of-distribution initial condition over the long-time rollout interval $[0,20]$, comparing the MAC model, the GRU-based model, the Markovian reduced-order model, and the linear and cubic memory models in~\autocite{Saad-2025}.}
            \end{figure}

        For these out-of-distribution initial conditions, we perform long-time rollouts over the time interval $t\in[0,20]$, whereas the MAC and GRU-based models are trained using data only over the time interval $[0,1]$. We compare the MAC model against the GRU-based model, the Markovian reduced-order model, and the linear and cubic memory models proposed in~\autocite{Saad-2025}.
        At each time instant, the absolute physical-space $L^2$ error is computed between the reconstructed resolved physical-space field and the corresponding true resolved field extracted from the full-order simulation. The resulting error evolutions for the two out-of-distribution initial conditions are shown in \Cref{fig_burgers_ood_case0,fig_burgers_ood_case1}.

        As shown in \Cref{fig_burgers_ood_case0,fig_burgers_ood_case1}, the MAC model maintains stable long-time prediction accuracy for both out-of-distribution initial conditions. The prediction accuracy of the MAC model is comparable to or better than that of the linear memory model in~\autocite{Saad-2025}, even though these initial conditions lie outside the training initial-condition distribution.
        In contrast, the GRU-based model initially produces reasonable predictions, but its error grows during the long-time rollout and eventually exceeds the errors of the MAC model and the linear and cubic memory models in~\autocite{Saad-2025}.
        We note that both the linear and cubic memory models in~\autocite{Saad-2025} require the evaluation of convolution integrals to estimate the memory during rollout. In contrast, the MAC model learns the effective non-Markovian dynamics directly from data and performs inference through a lightweight autoregressive sequence model. These comparisons suggest that the proposed MAC model achieves strong long-time predictive accuracy while maintaining favorable computational efficiency for long rollout horizons; see \Cref{apdx_runtime} for a computational cost comparison.

        We further investigate the physical consistency of the learned reduced-order dynamics through the evolution of the resolved energy. For the resolved Fourier modes, we define the resolved energy by
            \begin{displaymath}
                E_{\mathrm{res}}(t)
                =
                \sum_{k=1}^{M} \abs[\big]{\theta_k(t)}^2
                =
                \sum_{k=1}^{M} \big(\abs[\big]{\phi_k(t)}^2 + \abs[\big]{\psi_k(t)}^2\big).
            \end{displaymath}
            \begin{figure}[htbp]
                    \centering
                    \includegraphics[width=10cm]{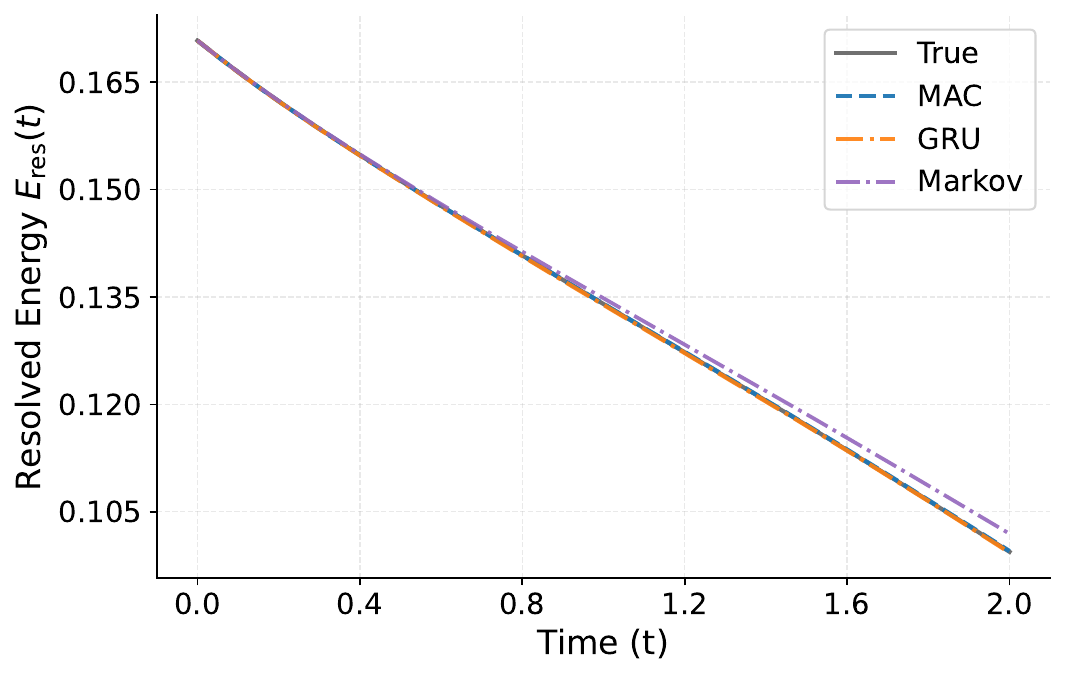}
                    \caption{\label{fig_burgers_energy}Evolution of the resolved energy over the temporal extrapolation interval $[0,2]$ for the same representative test initial condition as in \Cref{fig_burgers_interp_resolved}, comparing the MAC model, the GRU-based sequence model, and the Markovian reduced-order model.}
                \end{figure}
    
        \Cref{fig_burgers_energy} compares the evolution of the resolved energy for the same representative extrapolation test case considered above. As expected for the viscous Burgers' equation, the resolved energy decays over time due to viscous dissipation and nonlinear energy transfer to the unresolved modes.
        We observe that the MAC model accurately reproduces the evolution of the resolved energy throughout the extrapolation interval, whereas the Markovian reduced-order model exhibits noticeable deviation from the true energy trajectory. This indicates that the MAC model not only improves trajectory-level prediction accuracy, but also better captures the physically consistent long-time energy behavior induced by unresolved nonlinear interactions and closure effects.  

        Finally, we investigate the computational scalability of the proposed MAC model in both training and inference. 
        During training, we measure the average wall-clock time per training iteration for different input sequence lengths fed into the model.
        To ensure a fair comparison, a batch size of $96$ is used for all input sequence lengths during training.
        The results are reported in \Cref{tab_burgers_train_efficiency}.
        As shown in \Cref{tab_burgers_train_efficiency}, the training cost per iteration grows approximately linearly with the input sequence length for both the MAC model and the GRU-based sequence model. However, the GRU-based model is roughly $2.8$ times more expensive per iteration than the MAC model across all sequence lengths, owing to the inherently sequential nature of its recurrent computation, which prevents the parallelization across time steps that the MAC model exploits during training.     
\begin{table}[htbp]
    \centering
    \caption{\label{tab_burgers_train_efficiency} Average wall-clock time per training iteration of the MAC model and the GRU-based model for different input sequence lengths. The batch size is fixed at $96$ throughout the timing experiments.
    }
    \begin{tabular}{c|ccccc}
    \hline
    Training Interval
    & $[0,0.5]$
    & $[0,1.0]$
    & $[0,1.5]$
    & $[0,2.0]$
    & $[0,2.5]$ \\
    \hline
    Sequence Length
    & \num{5001}
    & \num{10001}
    & \num{15001}
    & \num{20001}
    & \num{25001} \\
    \hline
    MAC (s)
    & 0.044
    & 0.082
    & 0.127
    & 0.175
    & 0.219 \\
    GRU (s)
    & 0.123
    & 0.241
    & 0.364
    & 0.485
    & 0.607\\
    \hline
    \end{tabular}
\end{table}
            
\begin{figure}[htbp]
        \centering
        \includegraphics[width=0.6\textwidth]{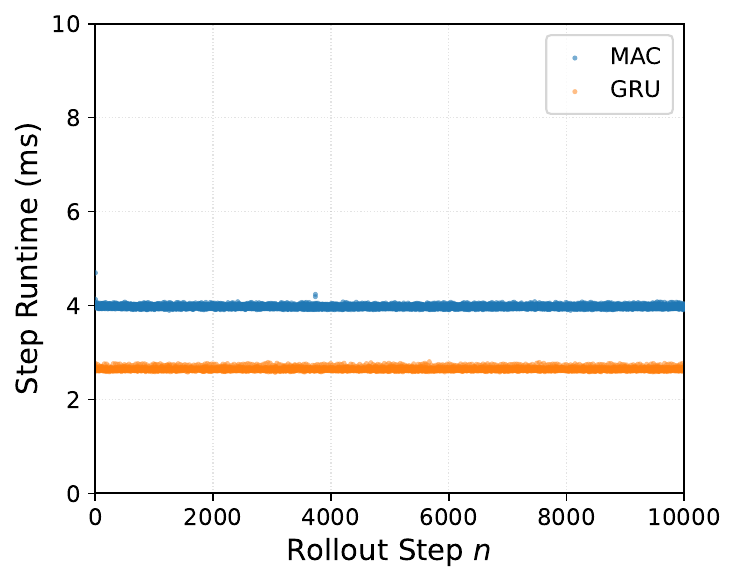}
        \caption{\label{fig_burgers_infer_efficiency}
        Wall-clock time of each inference step of the MAC model and the GRU-based model as a function of the rollout step index over the interpolation interval $[0,1]$ (\num{10000} steps). The batch size is fixed at $50$.}
        \end{figure}

        During autoregressive inference, we record the wall-clock time of advancing each rollout step for the entire batch. We note that the autoregressive rollouts are performed in parallel over the entire test batch, which contains $50$ initial conditions.
        \Cref{fig_burgers_infer_efficiency} plots the wall-clock time of each individual inference step against the rollout step index for the MAC model and the GRU-based model over the interpolation interval $[0,1]$ (\num{10000} steps at $\Delta t = 10^{-4}$, batch size $50$).
        As shown in \Cref{fig_burgers_infer_efficiency}, the wall-clock time of each step remains essentially constant throughout the rollout for both models, consistent with the autoregressive inference framework described in \Cref{sec_framework}, where temporal information is propagated through the recurrent state without repeatedly processing the entire temporal history. The GRU-based model is slightly faster at each inference step owing to its simpler recurrent update.
        This constant-cost behavior is particularly advantageous for long-time reduced-order simulation, where the dynamics must be advanced sequentially over many time steps.

        Compared with transformer-based sequence models, whose training cost typically grows rapidly with sequence length, the approximately linear scaling observed in \Cref{tab_burgers_train_efficiency} highlights the favorable scalability of the Mamba-based architecture for long training trajectories~\autocite{Vaswani-2017}.
        Similarly, during inference, the MAC model maintains essentially constant per-step rollout cost through a lightweight state-space update, in contrast to explicit memory-based approaches that require repeated evaluation of memory integrals over the trajectory history.
        Moreover, compared with GRU, whose recurrence is inherently sequential across time steps during training, the Mamba architecture achieves substantially lower measured training cost while also yielding lower prediction errors in the Burgers experiments.

\subsection{Two-Scale Lorenz '96 System}\label{sec_results_l96}

The two-scale Lorenz '96 system is a standard benchmark for studying multiscale chaotic dynamics and reduced-order closure modeling~\autocite{Lorenz-96,Bhouri-2023,Fatkullin-2004}. Originally introduced as a simplified atmospheric model, it has since become a canonical test problem in weather prediction, data assimilation, and multiscale modeling.

The system consists of a set of slow variables $X_k(t)$ coupled with a larger collection of fast variables $Y_j(t)$:
\begin{displaymath}
\frac{\ud X_k}{\ud t}
=
-X_{k-1}(X_{k-2}-X_{k+1})
-
X_k
+
F
-
\frac{hc}{b}
\sum_{j=J(k-1)+1}^{kJ}Y_j,
\qquad
k = 1, 2, \ldots, N,
\end{displaymath}
and
\begin{displaymath}
\frac{\ud Y_j}{\ud t}
=
-cbY_{j+1}(Y_{j+2}-Y_{j-1})
-
cY_j
+
\frac{hc}{b}
X_{\lfloor (j-1)/J \rfloor +1},
\qquad
j = 1, 2, \ldots, JN.
\end{displaymath}
Here the variables $\{X_k\}_{k=1}^N$ represent the slowly varying large-scale dynamics, while $\{Y_j\}_{j=1}^{JN}$ represent the rapidly varying small-scale variables.
The interaction between the fast and slow variables enters the slow dynamics through the coupling term
\begin{displaymath}
U_k(t)
\coloneq
-
\frac{hc}{b}
\sum_{j=J(k-1)+1}^{kJ}Y_j,
\qquad
k=1,2,\ldots,N.
\end{displaymath}
In our experiments, we set $N=8$, $J=32$, $F=10$, $h=1$, $c=10$, and $b=10$, corresponding to a chaotic multiscale regime commonly used in the Lorenz '96 literature~\autocite{Duben-2014}.
To verify numerically that the system is chaotic under our specific configuration, we estimated the maximum Lyapunov exponent via the Benettin renormalization method~\autocite{Benettin-1980} applied to the full state vector $(X, Y) \in \RNS^{N(1+J)}$. The running estimate converged over $t = 250$ to
\begin{displaymath}
    \lambda_{\max} \approx 6.426 > 0,
\end{displaymath}
confirming sensitive dependence on initial conditions and providing numerical evidence that the system operates in a chaotic regime.

In the reduced-order setting, the slow variables $\{X_k\}_{k=1}^{N}$ are treated as resolved variables, while the fast variables $\{Y_j\}_{j=1}^{JN}$ are regarded as unresolved degrees of freedom. The resolved dynamics may therefore be written as
\begin{displaymath}
\frac{\ud X_k}{\ud t}
=
-X_{k-1}(X_{k-2}-X_{k+1})
-
X_k
+
F
+
U_k,
\qquad
k=1,2,\ldots,N.
\end{displaymath}
The closure modeling problem then consists of learning an effective approximation of the coupling term $\{U_k\}_{k=1}^{N}$ from the temporal history of the resolved variables.

In our experiments, the MAC model is compared against the GRU-based model and the statistical method proposed by Wilks, which models the coupling term using polynomial regression together with autoregressive stochastic noise~\autocite{Cho-2014,Wilks-2005}.

To prepare training data, the full two-scale Lorenz '96 system is first simulated using the fine time step $\Delta t_{\mathrm{fine}}=0.005$. The resulting trajectories are then downsampled by a factor of two in time, yielding coarse observations with time step $\Delta t=0.01$. 
After an initial warmup stage to remove transient effects, a single long coarse trajectory of length $10001$ over the time interval $[0,100]$ is generated and used to construct the training dataset. More specifically, sliding windows of length $64$ with stride $1$ are extracted from the long trajectory and used as training sequences for the sequence models considered in this study.

Unlike the Burgers' experiments, where the sequence models are trained directly against the closure term, the sequence models for the Lorenz '96 system are trained using a multi-step resolved-variable rollout loss. More specifically, each sequence model first predicts the closure term from the temporal history of the resolved variables. 
The predicted closure is then coupled with the reduced slow dynamics, which are subsequently advanced using the time-stepping procedure described in \Cref{apdx_rk4_zero_order_holding}. The training loss is computed by comparing the resulting multi-step rollout of the resolved variables against the corresponding true resolved trajectory. In our experiments, we use a ten-step rollout loss. See \Cref{apdx_training_losses} for details.

To evaluate the MAC model on the Lorenz '96 system, we generate $100$ independent test trajectories using random initial conditions different from those used in the training trajectory. Each test trajectory contains $2001$ coarse time steps over the time interval $[0,20]$ and is evolved independently using the same numerical solver and parameter settings.

\begin{figure}[htbp]
    \centering
    \includegraphics[width=15cm]{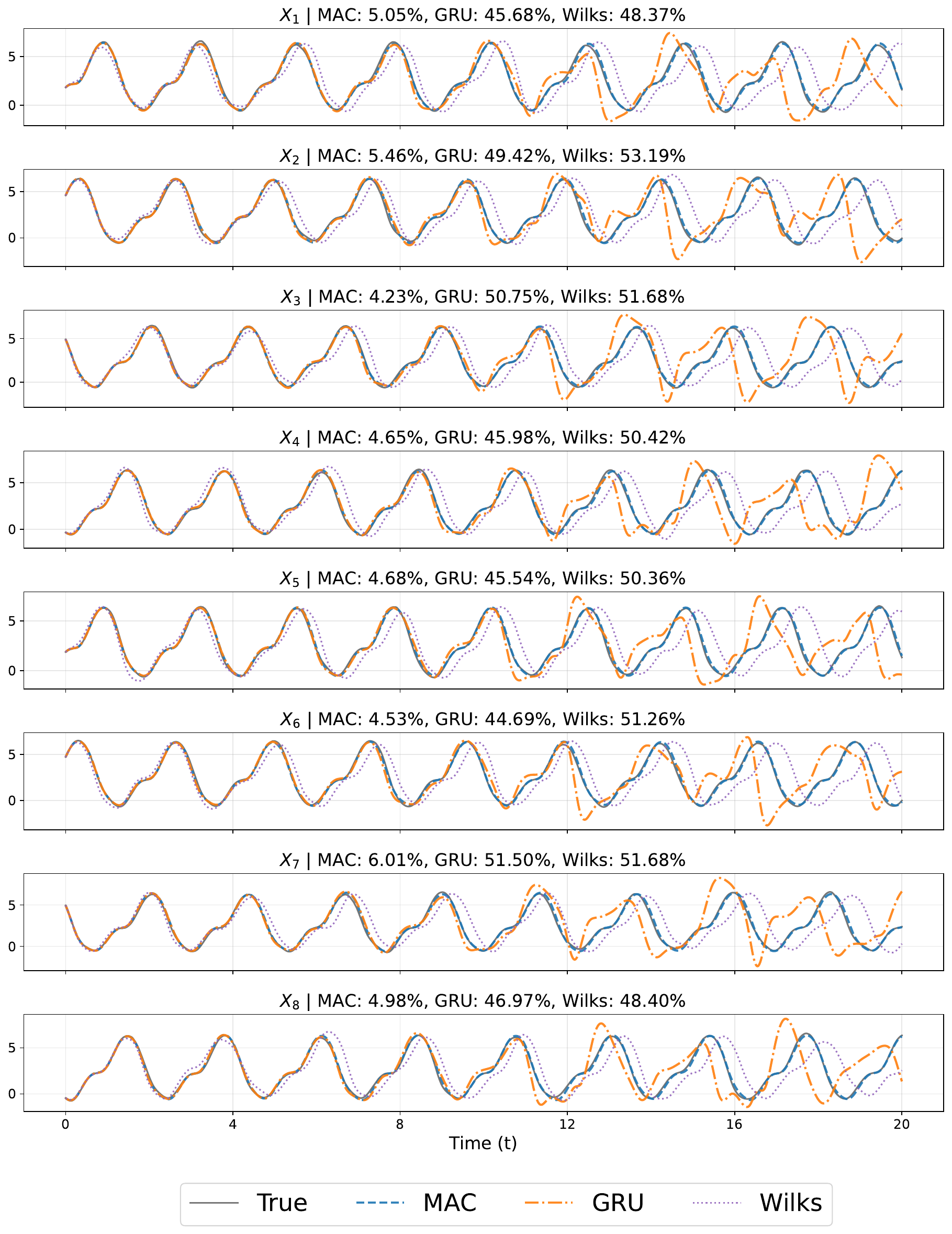}
    \caption{\label{fig_l96_unseen_resolved}Comparison of resolved slow-variable predictions from the MAC model, the GRU-based model, and the Wilks method for one representative unseen initial condition over the time interval $[0,20]$. For each resolved slow variable, the relative $L^2$ error is also reported.}
\end{figure}

We first present one representative unseen-initial-condition test trajectory to illustrate the qualitative rollout behavior of the different methods.
As shown in \Cref{fig_l96_unseen_resolved}, the MAC model accurately reproduces the chaotic oscillatory behavior of the resolved variables even for initial conditions that are not observed during training.
In contrast, the GRU-based model and the Wilks method exhibit substantial phase drift or amplitude distortion during autoregressive rollout.
During long-time autoregressive rollout, errors may gradually accumulate due to the chaotic nature of the dynamics. Nevertheless, the MAC model continues to deliver substantially improved trajectory accuracy and maintains the correct large-scale oscillatory behavior over long time horizons.

Following the evaluation procedure in~\autocite{Bhouri-2023}, we quantify the prediction accuracy using the running cumulative relative $L^2$ error. For a rollout trajectory, the cumulative relative $L^2$ error at time $t_i$ is defined by
\begin{displaymath}
\Biggl(
\frac{
\sum_{j=0}^{i}
\norm{X(t_j)-\widetilde{X}(t_j)}_2^2
}{
\sum_{j=0}^{T-1}
\norm{X(t_j)}_2^2
}
\Biggr)^{1/2},
\end{displaymath}
where $T=2001$ is the full sequence length, and $X$ and $\widetilde{X}$ denote the true and predicted vectors of resolved variables, respectively.

\begin{figure}[htbp]
    \centering
    \includegraphics[width=10cm]{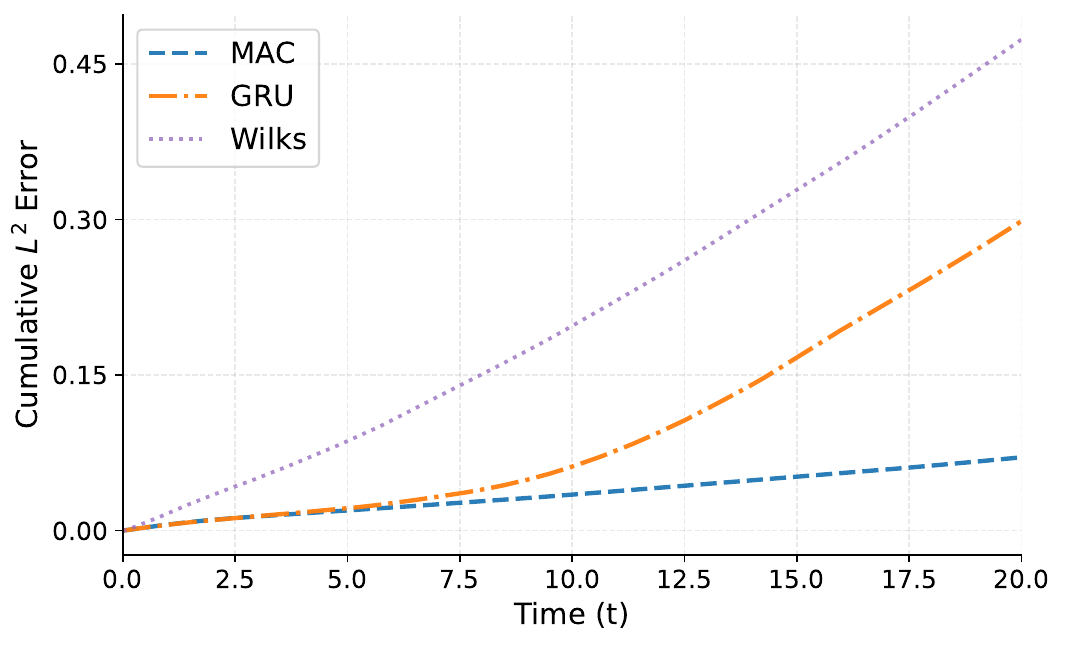}
    \caption{\label{fig_l96_unseen_rmse}Evolution of the running cumulative relative $L^2$ error averaged over $100$ unseen-initial-condition test trajectories over the time interval $[0,20]$ for the Lorenz '96 system, comparing the MAC model, the GRU-based model, and the Wilks method.
    }    
\end{figure}

We next examine the running cumulative correlation coefficient between the predicted and true resolved trajectories. The running cumulative correlation coefficient at time $t_i$ is defined by
\begin{displaymath}
\mathrm{Corr}(t_i)
=
\frac{
\sum_{j=0}^{i}
\bigl\langle X(t_j),\widetilde{X}(t_j)\bigr\rangle
}{
\Bigl(
\sum_{j=0}^{i}
\norm{X(t_j)}_2^2
\Bigr)^{1/2}
\Bigl(
\sum_{j=0}^{i}
\norm{\widetilde{X}(t_j)}_2^2
\Bigr)^{1/2}
}.
\end{displaymath}

\begin{figure}[htbp]
    \centering
    \includegraphics[width=10cm]{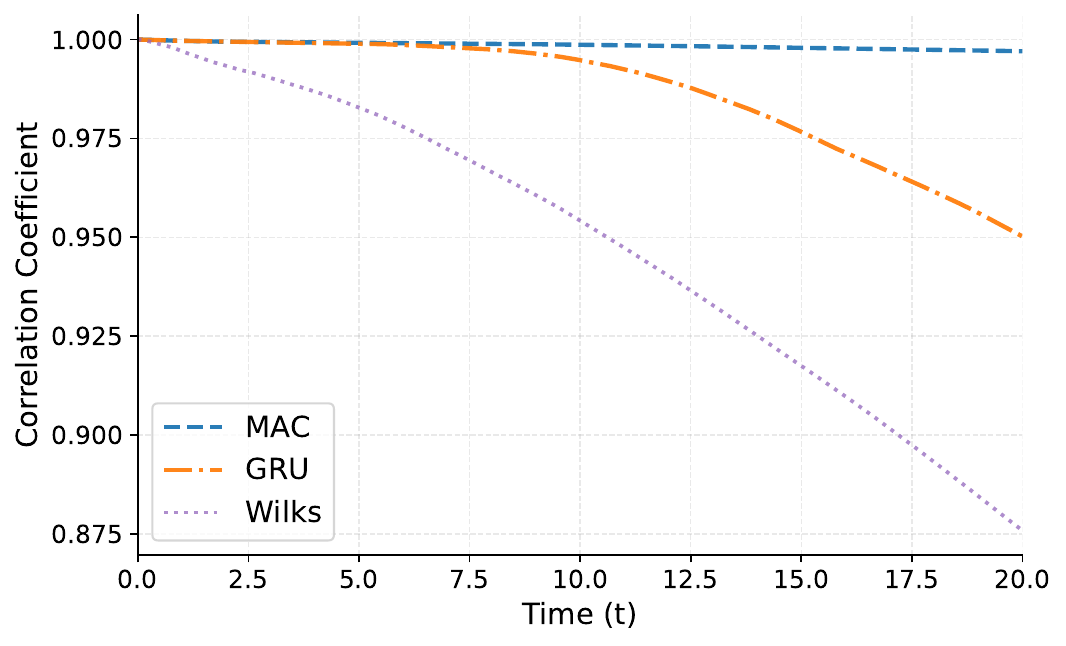}
    \caption{\label{fig_l96_unseen_corr}Evolution of the running cumulative correlation coefficient between the predicted and true resolved variables, averaged over $100$ unseen-initial-condition test trajectories over the time interval $[0,20]$ for the Lorenz '96 system, comparing the MAC model, the GRU-based model, and the Wilks method.}
\end{figure}

As shown in \Cref{fig_l96_unseen_rmse,fig_l96_unseen_corr}, the cumulative relative $L^2$ error increases over time for all three methods, but the growth is substantially slower for the MAC model, which also maintains consistently higher correlation with the true dynamics than both the GRU-based model and the Wilks method. These results indicate that the MAC closure model generalizes robustly to unseen initial conditions during long-time autoregressive rollout.

To reduce potential bias associated with the choice of random seeds, we additionally perform experiments using 5 random seeds. The cumulative resolved-variable relative $L^2$ error and correlation coefficient of the resolved variables 
are summarized in \Cref{tab_l96_statistics}. 
The mean relative $L^2$ errors for each resolved variable are reported in \Cref{tab_l96_resolved_errors}.
Overall, these results demonstrate that the MAC model provides substantially improved long-time predictive accuracy and robustness for the resolved dynamics compared with both the GRU-based model and the Wilks method.

\begin{table}[htbp]
\centering
\caption{\label{tab_l96_statistics}
Statistical diagnostics for the unseen-initial-condition experiment in the Lorenz '96 system. 
In this table, the cumulative resolved-variable relative $L^2$ error and correlation coefficient of the resolved variables
are reported. 
All values are averaged over 5 random seeds.
}
\begin{tabular}{c|ccc}
\hline
Metric & MAC & GRU & Wilks \\
\hline
Cumulative Relative $L^2$ Error & $0.0643$ & $0.2624$ & $0.4609$ \\
Correlation Coefficient & $0.9975$ & $0.9561$ & $0.8808$ \\
\hline
\end{tabular}
\end{table}

\begin{table}[htbp]
\centering
\caption{\label{tab_l96_resolved_errors}Relative $L^2$ error statistics for each resolved slow variable ($X_k$) in the unseen-initial-condition experiment. The mean and standard deviation are computed across $100$ test trajectories and averaged over 5 random seeds.}
\begin{tabular}{c|ccc}
\hline
Variable & MAC (\%) & GRU (\%) & Wilks (\%) \\
\hline
$X_1$ & $6.34 \pm 2.44$ & $25.84 \pm 10.27$ & $46.21 \pm 16.33$ \\
$X_2$ & $6.41 \pm 2.44$ & $25.75 \pm 10.46$ & $46.07 \pm 16.19$ \\
$X_3$ & $6.45 \pm 2.52$ & $26.88 \pm 10.93$ & $46.04 \pm 16.24$ \\
$X_4$ & $6.38 \pm 2.42$ & $25.89 \pm 10.52$ & $46.17 \pm 16.23$ \\
$X_5$ & $6.44 \pm 2.49$ & $25.78 \pm 10.10$ & $46.12 \pm 16.14$ \\
$X_6$ & $6.36 \pm 2.47$ & $26.97 \pm 10.69$ & $46.10 \pm 16.15$ \\
$X_7$ & $6.50 \pm 2.47$ & $25.77 \pm 10.09$ & $45.93 \pm 16.12$ \\
$X_8$ & $6.50 \pm 2.46$ & $26.56 \pm 10.69$ & $46.07 \pm 16.21$ \\
\hline
\end{tabular}
\end{table} 

\subsection{Power-Grid System}\label{sec_results_pg}

As a third benchmark problem, we consider the 3-bus DeMarco--Zheng power-grid model studied in \autocite{Li-2022}. This system provides a complementary test case to Burgers' equation and the Lorenz '96 system. In particular, the resolved and unresolved variables exhibit oscillatory dynamics without clear time-scale separation. As shown in \autocite{Li-2022}, this absence of time-scale separation leads to long-memory effects: to retain the accuracy of the reduced-order model, the memory length may need to be as long as the time interval over which the system is evolved. Therefore, this example provides a challenging benchmark for evaluating MAC in a system known to exhibit long-memory effects.

The full 3-bus dynamics used in this work are given by
\begin{displaymath}
\begin{split}
    \frac{d\omega_1}{dt}
    &=
    -\frac{D_1}{M_1}\omega_1
    +
    \frac{g_2+g_3}{M_1},\\
    \frac{d\omega_2}{dt}
    &=
    -\frac{D_2}{M_2}\omega_2
    -
    \frac{g_2}{M_2},\\
    \frac{d\alpha_2}{dt}
    &=
    -\omega_1+\omega_2,\\
    \frac{d\alpha_3}{dt}
    &=
    -\omega_1-\frac{g_3}{D_3},\\
    \frac{dV_3}{dt}
    &=
    -\frac{g_V}{\epsilon},
\end{split}
\end{displaymath}
where
\begin{displaymath}
\begin{split}
    g_2
    &=
    b_1V_1V_2\sin(\alpha_2)
    +
    b_3V_2V_3\sin(\alpha_2-\alpha_3)
    +
    P_2,\\
    g_3
    &=
    b_2V_1V_3\sin(\alpha_3)
    +
    b_3V_2V_3\sin(\alpha_3-\alpha_2)
    +
    P_3,\\
    g_V
    &=
    (b_2+b_3)V_3
    -
    b_2V_1\cos(\alpha_3)
    -
    b_3V_2\cos(\alpha_3-\alpha_2)
    +
    \frac{Q_3}{V_3}.
\end{split}
\end{displaymath}
Here the full state is
\begin{displaymath}
    (\omega_1,\omega_2,\alpha_2,\alpha_3,V_3),
\end{displaymath}
where $\omega_1$ and $\omega_2$ are the angular velocities of the two generators, $\alpha_2$ and $\alpha_3$ are voltage phase-angle differences relative to the reference bus, and $V_3$ is the voltage magnitude at the load bus. The voltage magnitudes $V_1$ and $V_2$ are fixed. The constants $M_1$, $M_2$, $D_1$, $D_2$, $D_3$, $\epsilon$, $b_1$, $b_2$, $b_3$, $P_2$, $P_3$, and $Q_3$ are system parameters, whose numerical values are taken from the 3-bus DeMarco--Zheng example in \autocite{Li-2022}.

Following the reduced-order setting in \autocite{Li-2022}, we choose the generator-related variables
\begin{displaymath}
    (\omega_1,\omega_2,\alpha_2)
\end{displaymath}
as the resolved variables, while the load-bus variables $(\alpha_3,V_3)$ are treated as unresolved variables. 

We first construct the Markovian part of the reduced-order model by fixing the unresolved variables at their reference values,
\begin{displaymath}
    \alpha_3=\alpha_3^0,
    \qquad
    V_3=V_3^0.
\end{displaymath}
In our experiments, we use
\begin{displaymath}
    \alpha_3^0=-0.3,
    \qquad
    V_3^0=0.8.
\end{displaymath}
Let $g_2^0$ and $g_3^0$ denote the quantities $g_2$ and $g_3$ evaluated with
$\alpha_3=\alpha_3^0$ and $V_3=V_3^0$. Then the Markovian reduced-order model is
\begin{displaymath}
\begin{split}
    \frac{d\omega_1}{dt}
    &=
    -\frac{D_1}{M_1}\omega_1
    +
    \frac{g_2^0+g_3^0}{M_1},\\
    \frac{d\omega_2}{dt}
    &=
    -\frac{D_2}{M_2}\omega_2
    -
    \frac{g_2^0}{M_2},\\
    \frac{d\alpha_2}{dt}
    &=
    -\omega_1+\omega_2.
\end{split}
\end{displaymath}
This model is closure-free: the unresolved variables are fixed at their reference values, so that the reduced dynamics depend only on the current resolved state.

To account for the effect of the unresolved variables, we augment the Markovian reduced-order model with a closure term. Since the equation for $\alpha_2$ depends only on $\omega_1$ and $\omega_2$, the closure has only two nontrivial components.
The closure-corrected reduced-order model is written as
\begin{displaymath}
\begin{split}
    \frac{d\omega_1}{dt}
    &=
    -\frac{D_1}{M_1}\omega_1
    +
    \frac{g_2^0+g_3^0}{M_1}
    +
    C_{\omega_1},\\
    \frac{d\omega_2}{dt}
    &=
    -\frac{D_2}{M_2}\omega_2
    -
    \frac{g_2^0}{M_2}
    +
    C_{\omega_2},\\
    \frac{d\alpha_2}{dt}
    &=
    -\omega_1+\omega_2.
\end{split}
\end{displaymath}
Under this construction, the closure terms are given by
\begin{displaymath}
\begin{split}
    C_{\omega_1}(t)
    &=
    \frac{g_2(t)+g_3(t)}{M_1}
    -
    \frac{g_2^0+g_3^0}{M_1},\\
    C_{\omega_2}(t)
    &=
    -
    \frac{g_2(t)}{M_2}
    +
    \frac{g_2^0}{M_2}.
\end{split}
\end{displaymath}
Therefore, in this experiment, the sequence models are trained to predict
\begin{displaymath}
    (C_{\omega_1}(t),C_{\omega_2}(t))
\end{displaymath}
from the temporal history of the resolved variables $(\omega_1,\omega_2,\alpha_2)$.

For data generation, we integrate the full 3-bus system on the time interval $[0,10]$ using a fine time step $\Delta t_{\rm fine}=5\times 10^{-5}$ and then downsample the trajectories every $20$ steps, yielding the data used by the sequence models with time step $\Delta t=10^{-3}$.
The unresolved variables are initialized at their reference values, $\alpha_3(0)=-0.3$ and $V_3(0)=0.8$, for all trajectories. The initial conditions of the resolved variables are sampled uniformly from
\begin{displaymath}
    \omega_1(0),\omega_2(0)\in[-0.05,0.05],
    \qquad
    \alpha_2(0)\in[-0.20,-0.12].
\end{displaymath}
We generate $512$ training trajectories, $64$ validation trajectories, and $100$
test trajectories. Along each trajectory, the reference closure terms
$(C_{\omega_1}(t),C_{\omega_2}(t))$ are computed from the expressions above at each
saved time instant.

To assess robustness, we also generate an out-of-distribution test set using a
wider range of resolved initial conditions,
\begin{displaymath}
    \omega_1(0),\omega_2(0)\in[-0.10,0.10],
    \qquad
    \alpha_2(0)\in[-0.24,-0.08],
\end{displaymath}
while keeping the same initial values for the unresolved variables. This out-of-distribution set contains $50$ trajectories. 

\begin{figure}[htbp]
    \centering
    \includegraphics[width=0.55\textwidth]{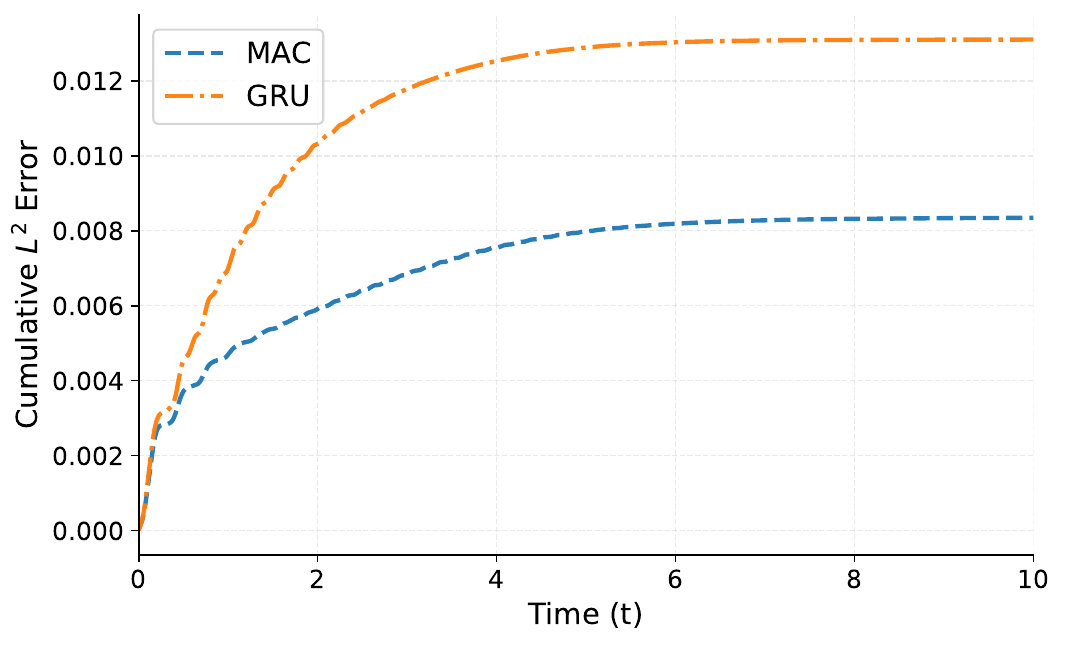}
    \caption{Evolution of the running cumulative relative $L^2$ error averaged over 100 in-distribution test trajectories over the time interval $[0,10]$ for the power-grid system, comparing the MAC model and the GRU-based model.}
    \label{fig_pg_rmse}
\end{figure}

We first evaluate the models on the in-distribution test set. \Cref{fig_pg_rmse} shows the cumulative relative $L^2$ error of the resolved variables over the rollout interval. Since the Markovian reduced-order model produces errors that are several orders of magnitude larger than those of the learned closure models, it is omitted from this plot in order to make the comparison between the MAC model and the GRU-based model visible. The Markovian baseline will be included in the trajectory comparison and in the quantitative error statistics below. As shown in \Cref{fig_pg_rmse}, the MAC model maintains a consistently smaller cumulative error than the GRU-based model over the full time interval.

\begin{figure}[htbp]
    \centering
    \includegraphics[width=\textwidth]{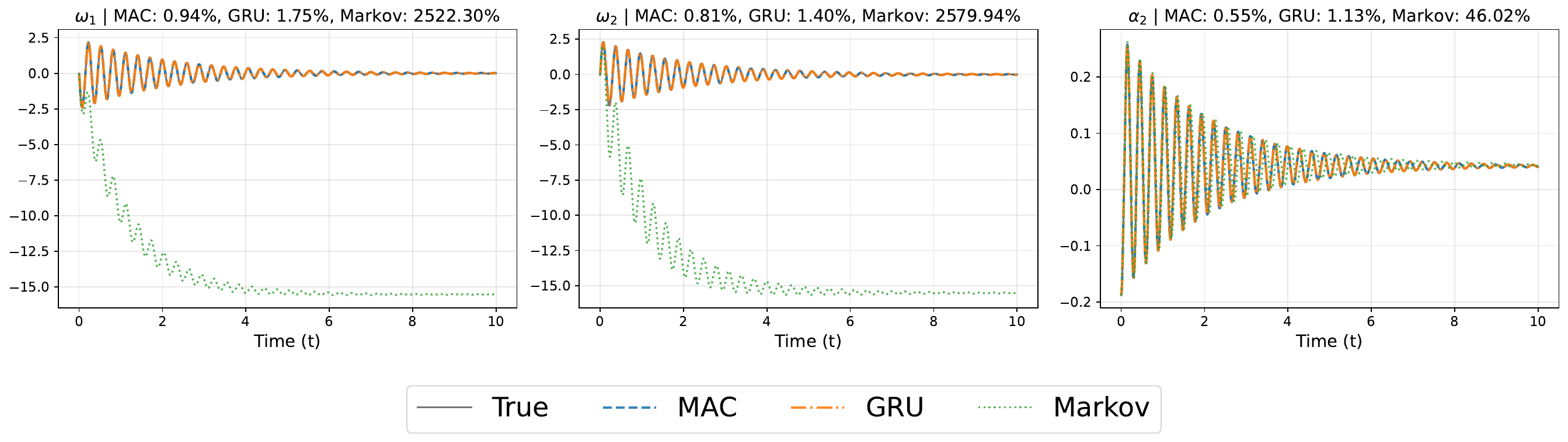}
    \caption{Comparison of resolved-variable predictions from the MAC model, the GRU-based model, and the Markovian reduced-order model on one representative in-distribution test initial condition over the time interval $[0,10]$. For each resolved variable, the relative $L^2$ error over the time interval $[0,10]$ is also reported.}
    \label{fig_pg_resolved}
\end{figure}

\Cref{fig_pg_resolved} shows the resolved variables for a representative in-distribution test trajectory. The Markovian reduced-order model rapidly deviates from the full-system trajectory, especially in the two generator angular velocities. In contrast, both the MAC model and the GRU-based model remain stable over the full rollout interval. The MAC model gives the most accurate predictions among the reduced-order models, with relative $L^2$ errors of $0.94\%$, $0.81\%$, and $0.55\%$ for $\omega_1$, $\omega_2$, and $\alpha_2$, respectively. The corresponding GRU errors are $1.75\%$, $1.40\%$, and $1.13\%$.

The relative $L^2$ errors of the resolved variables over the in-distribution test set are reported in \Cref{tab_pg_error}. The Markovian reduced-order model has errors above $2900\%$ for the two angular velocities and above $43\%$ for $\alpha_2$, confirming that the Markovian closure-free reduced-order model is inadequate for this system.
Both sequence models substantially reduce the errors, but the MAC model is more accurate than the GRU-based model for all three resolved variables. In particular, MAC reduces the mean relative $L^2$ error from $1.7111\%$ to $0.8886\%$ for $\omega_1$, from $1.7980\%$ to $0.8866\%$ for $\omega_2$, and from $1.2276\%$ to $0.5408\%$ for $\alpha_2$.

\begin{table}[htbp]
\centering
\caption{Relative $L^2$ error statistics for each resolved variable on the in-distribution test set for the power-grid system. The mean and standard deviation are computed across all initial conditions in the test dataset and over 5 random seeds for the MAC model, the GRU-based model, and the Markovian reduced-order model.}
\label{tab_pg_error}
\begin{tabular}{c|ccc}
\hline
Variable & MAC & GRU & Markov \\
\hline
$\omega_1$ & $0.8886\% \pm 0.1045\%$ & $1.7111\% \pm 0.5626\%$ & $2916.6965\% \pm 346.5254\%$ \\
$\omega_2$ & $0.8866\% \pm 0.1067\%$ & $1.7980\% \pm 0.6029\%$ & $2978.8986\% \pm 354.2861\%$ \\
$\alpha_2$ & $0.5408\% \pm 0.0841\%$ & $1.2276\% \pm 0.4082\%$ & $43.3424\% \pm 2.3344\%$ \\
\hline
\end{tabular}
\end{table}

\begin{figure}[htbp]
    \centering
    \includegraphics[width=0.55\textwidth]{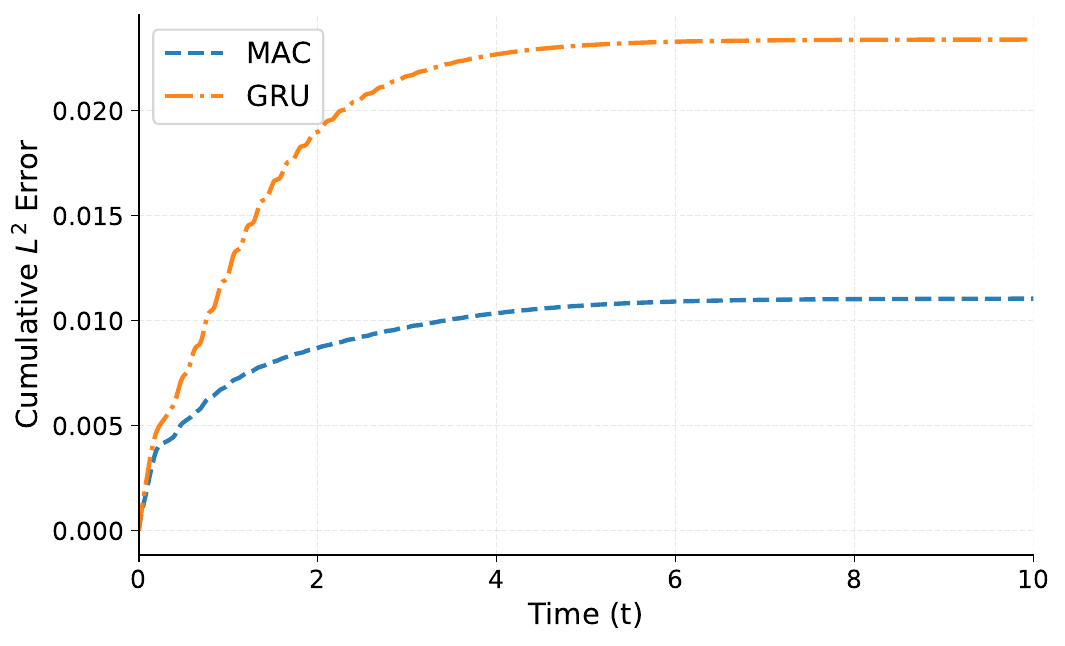}
    \caption{Evolution of the running cumulative relative $L^2$ error averaged over the out-of-distribution test trajectories over the time interval $[0,10]$ for the power-grid system, comparing the MAC model and the GRU-based model.}
    \label{fig_pg_ood_rmse}
\end{figure}

We next evaluate the models on the out-of-distribution test set. \Cref{fig_pg_ood_rmse} shows the cumulative relative $L^2$ error of the resolved variables over the rollout interval. As in the in-distribution case, the Markovian reduced-order model is omitted from this plot because its error is much larger than those of the learned closure models. As shown in \Cref{fig_pg_ood_rmse}, the MAC model maintains a consistently smaller cumulative error than the GRU-based model over the full time interval, indicating better robustness under the wider range of resolved initial conditions.

\begin{figure}[htbp]
    \centering
    \includegraphics[width=\textwidth]{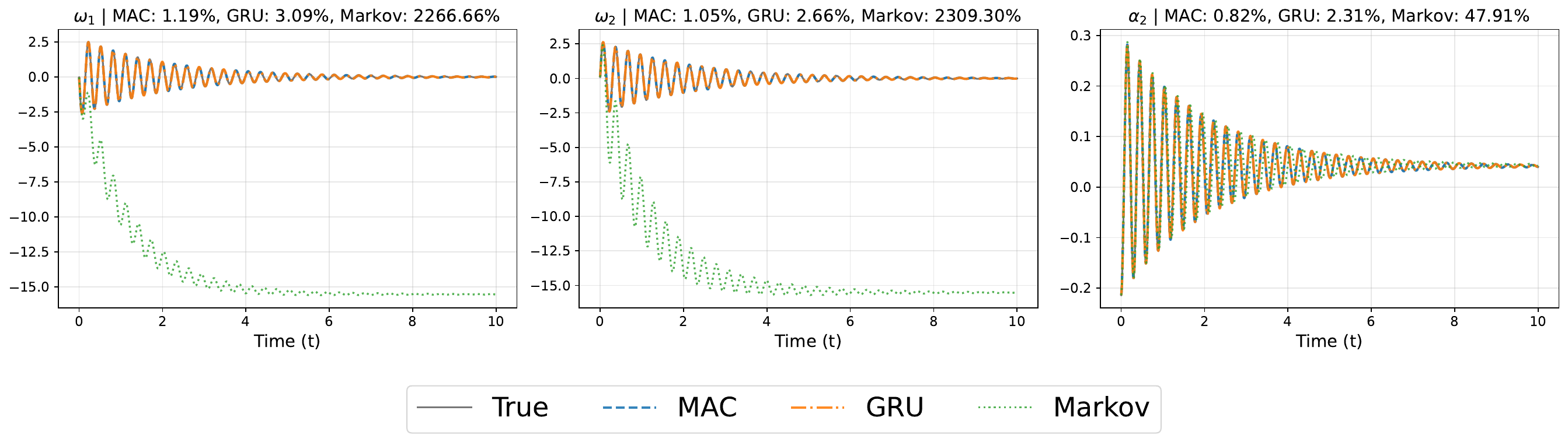}
    \caption{Comparison of resolved-variable predictions from the MAC model, the GRU-based model, and the Markovian reduced-order model on one representative out-of-distribution test initial condition over the time interval $[0,10]$. For each resolved variable, the relative $L^2$ error over the time interval $[0,10]$ is also reported.}
    \label{fig_pg_ood_resolved}
\end{figure}

\Cref{fig_pg_ood_resolved} shows the resolved variables for a representative out-of-distribution test trajectory. The Markovian reduced-order model again rapidly deviates from the full-system trajectory, while both the MAC model and the GRU-based model remain stable over the full rollout interval. The MAC model gives more accurate predictions than the GRU-based model, with relative $L^2$ errors of $1.19\%$, $1.05\%$, and $0.82\%$ for $\omega_1$, $\omega_2$, and $\alpha_2$, respectively. The corresponding GRU errors are $3.09\%$, $2.66\%$, and $2.31\%$.

The relative $L^2$ errors of the resolved variables over the out-of-distribution test set are reported in \Cref{tab_pg_ood_error}. The Markovian reduced-order model again produces very large errors, with mean relative $L^2$ errors above $3000\%$ for the two angular velocities and above $42\%$ for $\alpha_2$. Both sequence models substantially reduce the errors, but the MAC model remains more accurate than the GRU-based model for all three resolved variables. In particular, MAC reduces the mean relative $L^2$ error from $3.0965\%$ to $1.2879\%$ for $\omega_1$, from $3.1293\%$ to $1.2569\%$ for $\omega_2$, and from $2.1812\%$ to $0.7457\%$ for $\alpha_2$.

\begin{table}[htbp]
\centering
\caption{Relative $L^2$ error statistics for each resolved variable on the out-of-distribution test set for the power-grid system. The mean and standard deviation are computed across all initial conditions in the out-of-distribution test dataset and over 5 random seeds for the MAC model, the GRU-based model, and the Markovian reduced-order model.}
\label{tab_pg_ood_error}
\begin{tabular}{c|ccc}
\hline
Variable & MAC & GRU & Markov \\
\hline
$\omega_1$ & $1.2879\% \pm 0.4981\%$ & $3.0965\% \pm 1.4482\%$ & $3024.8686\% \pm 795.6225\%$ \\
$\omega_2$ & $1.2569\% \pm 0.5050\%$ & $3.1293\% \pm 1.4045\%$ & $3089.9292\% \pm 813.9286\%$ \\
$\alpha_2$ & $0.7457\% \pm 0.3063\%$ & $2.1812\% \pm 1.0938\%$ & $42.8866\% \pm 5.0633\%$ \\
\hline
\end{tabular}
\end{table}

\subsection{Korteweg--de Vries Equation}\label{sec_results_kdv}
Consider the Korteweg--de Vries (KdV) equation with periodic boundary conditions
    \begin{displaymath}
    u_t
    =
    \partial_x\Bigl(-\frac{1}{2}u^2\Bigr)
    -
    \delta^2 u_{xxx},
    \qquad
    u(0,x)=u_0(x),
    \qquad
    x\in[0,2\pi).
    \end{displaymath}
Throughout our experiments, the dispersion coefficient is set to $\delta=0.1$.
In contrast to the viscous Burgers' equation, the KdV equation is nondissipative and provides a complementary setting for evaluating closure models in dispersive dynamics.

Expanding the solution $u(t,x)$ in terms of Fourier modes
    \begin{displaymath}
    u(t,x)
    =
    \sum_{k\in\ZN}\theta_k(t)e^{ikx},
    \end{displaymath}
we can rewrite the KdV equation as a system of ODEs
    \begin{equation}\label{eq_kdv_resolved}
    \theta_k'(t)
    =
    -\frac{ik}{2}
    \sum_{\substack{p,q\in\mathbb{Z}\\p+q=k}}
    \theta_p(t)\theta_q(t)
    -
    \delta^2(ik)^3\theta_k(t)
    \eqcolon
    R_k\bigl(\theta(t)\bigr),
    \quad\text{for\quad} k\in\ZN.
    \end{equation}
As in the Burgers' equation, the reality of $u$ implies $\theta_{-k}(t)=\overline{\theta_k(t)}$, and we restrict attention to initial conditions with a vanishing zero Fourier mode.
To construct a reduced-order system, we retain the first $M=3$ positively indexed complex Fourier modes as resolved variables and write
    \begin{displaymath}
    \theta_k(t)
    =
    \phi_k(t)+i\psi_k(t),
    \qquad
    k=1,2,3,
    \end{displaymath}
corresponding to six real resolved variables.

Following the same Fourier-space construction as in the Burgers' experiment, the closure terms associated with each resolved mode are
    \begin{displaymath}
    \mathcal{M}_k(t)
    =
    R_k\bigl(\theta(t)\bigr)
    -
    R_k\bigl(\widetilde{\theta}(t)\bigr),
    \qquad
    k=1,2,\ldots,M,
    \end{displaymath}
where $\widetilde{\theta}(t)$ denotes the truncated Fourier state obtained by retaining only the resolved positively indexed modes and enforcing the corresponding reality constraint.
We further write
    \begin{displaymath}
    \mathcal{M}_k(t)
    =
    m_k^\phi(t)+i m_k^\psi(t).
    \end{displaymath}

As in the Burgers' experiment, the sequence models are trained to predict the closure terms
    \begin{displaymath}
    \{m_k^\phi,m_k^\psi\}_{k=1}^M
    \end{displaymath}
from the trajectories of the resolved modes
    \begin{displaymath}
    \{\phi_k,\psi_k\}_{k=1}^M.
    \end{displaymath}
During inference, the predicted closure is coupled with the reduced-order KdV dynamics to evolve the resolved modes autoregressively from the initial conditions.

In data preparation, the initial conditions are sampled in the resolved Fourier modes by drawing both $\phi_k(0)$ and $\psi_k(0)$ uniformly from $(-\ue^{-k},\ue^{-k})$,  $k=1,2,\ldots,M$.
The negative Fourier modes are determined by conjugate symmetry, while the zero mode and all higher modes are initially set to zero.

We generate training trajectories on the time interval $[0,1]$ using a Fourier pseudospectral method and a fourth-order time integrator with time step $\Delta t=10^{-4}$.
During training, each sequence model directly predicts the closure terms from the resolved trajectories, and the loss is computed as the mean-squared error between the predicted and reference closures. 

\begin{figure}[htbp]
    \centering
    \includegraphics[width=16cm]{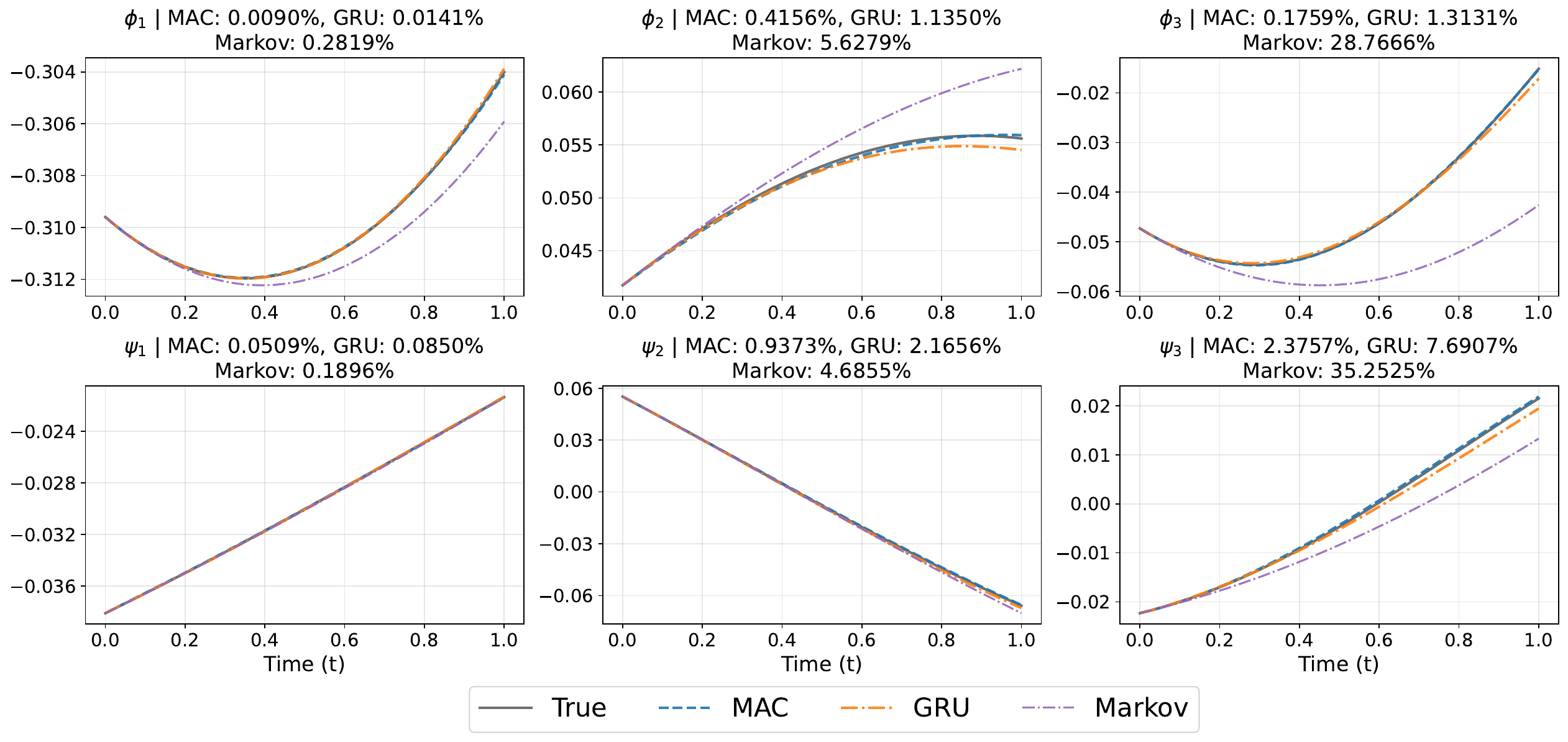}
    \caption{\label{fig_kdv_interp_resolved}Comparison of resolved-mode predictions from the MAC model, the GRU-based model, and the Markovian reduced-order model on one representative test initial condition over the temporal interpolation regime $[0,1]$. For each resolved Fourier mode, the relative $L^2$ error over the time interval $[0,1]$ is also reported.}
\end{figure}

We first examine the prediction accuracy within the training time interval $[0,1]$, corresponding to a temporal interpolation regime.
\Cref{fig_kdv_interp_resolved} compares the predicted trajectories of the resolved Fourier modes against the true solutions for one representative test initial condition.

From \Cref{fig_kdv_interp_resolved}, we observe that both sequence models reproduce the true resolved trajectories substantially more accurately than the Markovian model, with the MAC model providing the closest agreement overall. 
The mean relative $L^2$ error for each Fourier mode across all initial conditions in the test dataset, reported in \Cref{tab_kdv_interp_resolved}, further confirms this observation.
The MAC model achieves the lowest mean error for every resolved mode and consistently improves upon the GRU-based model.
Compared with the Markovian model, the improvement becomes particularly pronounced for the higher-frequency modes, for which the mean error is reduced by approximately one order of magnitude.

\begin{table}[htbp]
    \centering
    \caption{\label{tab_kdv_interp_resolved}Relative $L^2$ error statistics for each resolved Fourier mode over the temporal interpolation regime $[0,1]$. The mean and standard deviation are computed across all initial conditions in the test dataset and over 5 random seeds for the MAC model, the GRU-based model, and the Markovian reduced-order model.}
    \begin{tabular}{c|ccc}
    \hline
    Mode & MAC (\%) & GRU (\%) & Markov (\%) \\
    \hline
    $\phi_1$ & $0.0726 \pm 0.0946$ & $0.0886 \pm 0.1287$ & $0.3164 \pm 0.3360$ \\
    $\phi_2$ & $0.4181 \pm 0.4040$ & $0.4980 \pm 0.5271$ & $3.1573 \pm 2.0444$ \\
    $\phi_3$ & $2.2643 \pm 2.7112$ & $2.7898 \pm 2.8836$ & $23.0732 \pm 17.7627$ \\
    \hline
    $\psi_1$ & $0.1123 \pm 0.1932$ & $0.1410 \pm 0.2532$ & $0.3737 \pm 0.5696$ \\
    $\psi_2$ & $0.4811 \pm 0.5893$ & $0.5674 \pm 0.7079$ & $3.0305 \pm 2.9979$ \\
    $\psi_3$ & $2.7115 \pm 3.3148$ & $3.8516 \pm 4.7018$ & $30.3956 \pm 32.7351$ \\
    \hline
    \end{tabular}
\end{table}

Long-time temporal extrapolation poses an additional challenge for the nonlinear, nondissipative, and dispersive KdV dynamics. We therefore focus the main discussion on prediction within the training time interval and report the temporal extrapolation results separately in \Cref{apdx_kdv_extrapolation}. 

\clearpage
\section{Conclusion}\label{sec_conclusion}

In this work, we proposed the Mamba-Assisted Closure (MAC) framework, which recasts non-Markovian closure modeling for reduced-order dynamical systems as a sequence modeling problem. Motivated by the memory effects arising in the Mori--Zwanzig formalism, MAC uses a Mamba-based sequence model to predict the closure from the temporal history of the resolved trajectory. The input-dependent selective mechanism of Mamba provides additional flexibility in retaining and propagating temporal information as the resolved state evolves. At the same time, the framework does not require a prescribed history window or explicit estimation of a memory kernel. From a computational perspective, Mamba supports parallel sequence processing through the selective scan during training and recurrent state updates during autoregressive inference. During inference, the history of the resolved variables is carried implicitly through the recurrent state, allowing the closure model to advance step by step without reprocessing the entire preceding trajectory.

We validated the MAC framework on four benchmark problems with complementary dynamical characteristics. For the viscous Burgers' equation, MAC consistently provides more accurate predictions of the resolved Fourier modes than both the parameter-matched GRU-based sequence model and the Markovian closure-free reduced-order model in the temporal interpolation and extrapolation regimes. It also preserves the expected dissipative behavior of the resolved energy and remains stable and accurate over long-time rollouts from structured initial conditions. In these long-time tests, MAC achieves accuracy comparable to or better than the linear and cubic memory models in~\autocite{Saad-2025}, while avoiding the increasing cost associated with evaluating their memory terms as the rollout progresses. 
For the chaotic two-scale Lorenz '96 system, MAC yields substantially lower errors in the resolved variables than both the GRU-based sequence model and the classical Wilks method on unseen initial conditions, while maintaining accurate long-time resolved trajectories.
For the 3-bus DeMarco--Zheng power-grid system, where prior work has demonstrated pronounced long-memory effects, MAC yields lower errors than the GRU-based sequence model for all three resolved variables in both the in-distribution and out-of-distribution tests, with substantially lower errors than the Markovian closure-free reduced-order model as well.
The Korteweg--de Vries equation provides a dispersive PDE example. MAC again yields lower errors than the GRU-based sequence model for all resolved Fourier modes, with even larger improvements over the Markovian closure-free reduced-order model.

Across all four benchmark problems, the MAC framework achieves better predictive accuracy than the parameter-matched GRU-based sequence model. 
The comparisons with the linear and cubic memory models for Burgers' equation further show that MAC can achieve competitive accuracy without explicitly evaluating the memory integral over an increasingly long trajectory history during inference.
Consistent with this advantage, the scalability experiments show that the training cost of MAC grows approximately linearly with sequence length, while its per-step inference cost remains essentially constant with respect to the rollout horizon. 
Taken together, these results demonstrate that the MAC framework provides an effective and computationally scalable approach to non-Markovian closure modeling and long-time reduced-order simulation.

A natural direction for future work is to assess generalization across variations in the governing equation parameters. Throughout this study, these parameters (e.g., the viscosity $\nu$ in Burgers' equation, the dispersion coefficient $\delta$ in the Korteweg--de Vries equation, and the parameters of the Lorenz~'96 and power-grid systems) are kept fixed. Extending the MAC framework to settings with varying physical parameters would further test its robustness and broaden its applicability to reduced-order modeling problems in which such parameters are uncertain or vary across scenarios.

\section*{Code and Data Availability}
\addcontentsline{toc}{section}{\protect\numberline {}{Code and Data Availability}}
    To support reproducibility and facilitate future research, the code and all accompanying data will be made publicly available upon publication acceptance.

\section*{Acknowledgments}
\addcontentsline{toc}{section}{\protect\numberline {}{Acknowledgments}}

The work of Zhi-Feng Wei was supported by the U.S.\ Department of Energy (DOE) Office of Advanced Scientific Computing Research (ASCR) through the ASCR Distinguished Computational Mathematics Postdoctoral Fellowship (Project Nos.\ 71268 and 83358).
The work of Saad Qadeer was supported by the U.S.\ Department of Energy, Office of Science, Scientific Discovery through Advanced Computing (SciDAC) program, via a partnership in Earth System Model Development between the ASCR and Biological and Environmental Research (BER) programs, as part of the project titled ``Physical, Accurate, and Efficient atmosphere and surface coupling across SCALes'' (PAESCAL; proposal No.\ 0000267817).
The work of Panos Stinis was partially supported by ASCR under the ``Resolution-invariant deep learning for accelerated propagation of epistemic and aleatory uncertainty in multi-scale energy storage systems, and beyond'' project (Project No.\ 81824).

The computational work was performed using the Pacific Northwest National Laboratory (PNNL) Research Computing facilities.

PNNL is a multi-program national laboratory operated for the U.S.\ Department of Energy by Battelle Memorial Institute under Contract No.\ DE-AC05-76RL01830.

\begin{appendices}
\crefalias{section}{appendix}
\crefalias{subsection}{appendix}

\counterwithin{theorem}{section}
\counterwithin{equation}{section}
\counterwithin{figure}{section}
\counterwithin{table}{section}

\renewcommand{\thetheorem}{\thesection\arabic{theorem}}
\renewcommand{\theequation}{\thesection\arabic{equation}}
\renewcommand{\thefigure}{\thesection\arabic{figure}}
\renewcommand{\thetable}{\thesection\arabic{table}}

\clearpage
\section{Additional Numerical Results and Computational Cost}\label{apdx_additional}
\subsection{Additional Closure-Term Predictions}
\label{apdx_closure_plots}

The main text focuses on the resolved-variable predictions. 
Here we provide additional results for the learned closure terms, 
including predictions for representative initial conditions and related summary statistics. 
These results complement the resolved-variable comparisons reported in the main text.

\subsubsection{Burgers' Equation}
\label{apdx_closure_burgers}

\begin{figure}[htbp]
    \centering
    \includegraphics[width=0.9\textwidth]{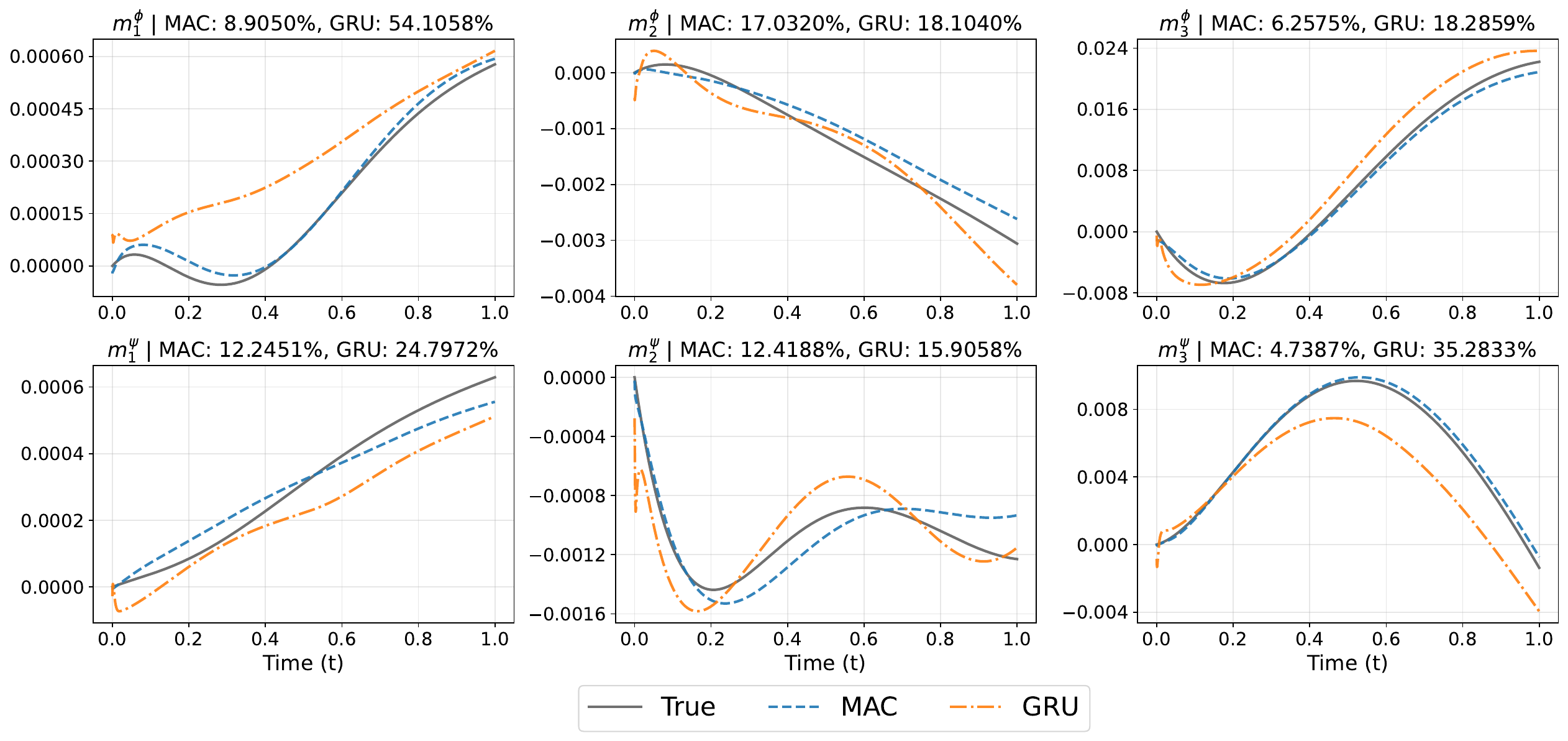}
    \caption{\label{fig_burgers_test_closure}Comparison of the closure-term predictions from the MAC model, the GRU-based model, and the true closure terms on the same representative test initial condition as in \Cref{fig_burgers_interp_resolved} over the temporal interpolation regime $[0,1]$. For each closure term, the relative $L^2$ error over the time interval $[0,1]$ is also reported.
    }
\end{figure}

\begin{figure}[htbp]
    \centering
    \includegraphics[width=0.9\textwidth]{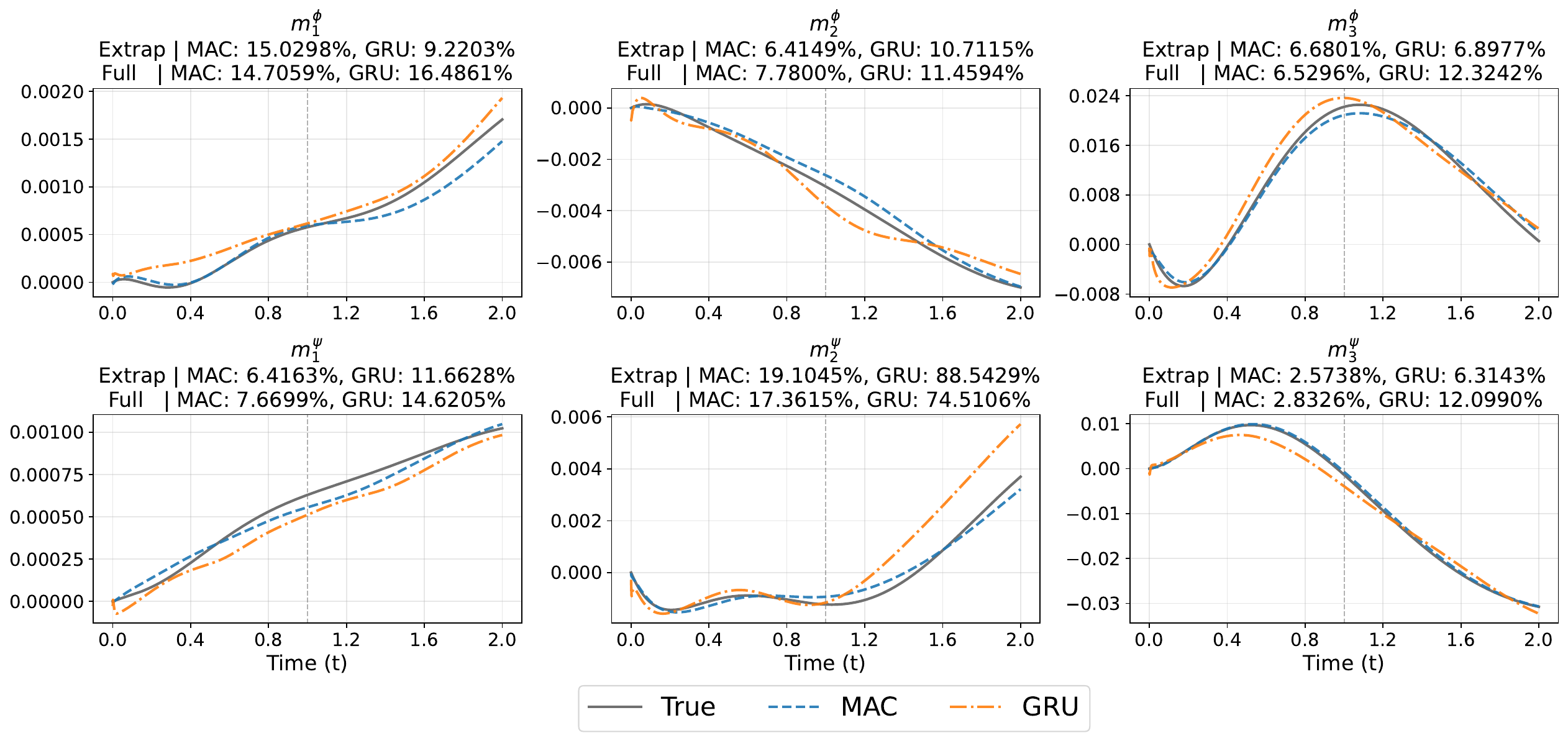}
    \caption{\label{fig_burgers_extrap_closure}Comparison of the closure-term predictions from the MAC model, the GRU-based model, and the true closure terms on the same representative test initial condition as in \Cref{fig_burgers_interp_resolved} over the temporal extrapolation regime $[0,2]$. For each closure term, the relative $L^2$ errors over the intervals $[1,2]$ and $[0,2]$ are also reported.
    }
\end{figure}

For Burgers' equation, the main text reports autoregressive rollout results for the resolved Fourier modes in \Cref{sec_results_burgers}. 
Here, in \Cref{fig_burgers_test_closure,fig_burgers_extrap_closure}, we show the corresponding closure-term predictions for the same representative test initial condition as in \Cref{fig_burgers_interp_resolved}. 
These plots directly compare the predicted closure terms produced by the MAC model and the GRU-based model with the true closure data.

To complement the closure-term predictions for the representative test initial condition, we also report summary statistics for the Burgers closure prediction errors.
\Cref{tab_burgers_closure_error} lists the mean relative $L^2$ errors of the closure terms predicted by the MAC model and the GRU-based model on the interpolation interval $[0,1]$, the extrapolation interval $[1,2]$, and the full interval $[0,2]$.
These values are computed over $50$ test cases and averaged over 5 random seeds.
Although the closure-term relative errors are larger than the resolved-variable relative errors reported in the main text, the predicted closure trajectories still capture the main trends of the true closure data.
In fact, the relative $L^2$ errors of the predicted closure terms can appear large because the closure terms themselves have small magnitudes, so that even a small absolute discrepancy may lead to a relatively large relative error.
From \Cref{tab_burgers_closure_error}, the MAC model consistently achieves lower closure prediction errors than the GRU-based model across all modes and all intervals.
The closure prediction errors from the MAC model remain comparable between the interpolation and extrapolation intervals, suggesting that the learned closure does not substantially degrade over the longer rollout. The GRU-based model, by contrast, exhibits more noticeable error growth on the extrapolation interval.
\begin{table}[htbp]
    \centering
    \caption{\label{tab_burgers_closure_error}Mean relative $L^2$ errors of the Burgers closure terms predicted by the MAC model and the GRU-based model over the interpolation interval $[0,1]$, the extrapolation interval $[1,2]$, and the full interval $[0,2]$. The values are averaged over $50$ test cases and 5 random seeds.}
    \resizebox{\textwidth}{!}{
    \begin{tabular}{l|cc|cc|cc}
        \hline
        \multirow{2}{*}{Closure}
        & \multicolumn{2}{c|}{Interpolation $[0,1]$}
        & \multicolumn{2}{c|}{Extrapolation $[1,2]$}
        & \multicolumn{2}{c}{Full $[0,2]$} \\
        & MAC & GRU & MAC & GRU & MAC & GRU \\
        \hline
        $m_1^\phi$ & $48.20\%$ & $62.51\%$ & $42.33\%$ & $65.53\%$ & $42.15\%$ & $58.56\%$ \\
        $m_2^\phi$ & $36.31\%$ & $57.89\%$ & $44.21\%$ & $67.51\%$ & $36.50\%$ & $56.78\%$ \\
        $m_3^\phi$ & $28.57\%$ & $35.32\%$ & $40.61\%$ & $50.75\%$ & $32.08\%$ & $40.06\%$ \\
        $m_1^\psi$ & $42.86\%$ & $77.27\%$ & $40.64\%$ & $75.54\%$ & $30.71\%$ & $51.26\%$ \\
        $m_2^\psi$ & $23.24\%$ & $41.46\%$ & $23.54\%$ & $55.69\%$ & $18.89\%$ & $39.88\%$ \\
        $m_3^\psi$ & $35.27\%$ & $58.58\%$ & $46.28\%$ & $91.18\%$ & $39.67\%$ & $70.78\%$ \\
        \hline
    \end{tabular}
    }
\end{table}

\subsubsection{Lorenz '96 System}
\label{apdx_closure_l96}

\begin{figure}[htbp]
    \centering
    \includegraphics[width=15cm]{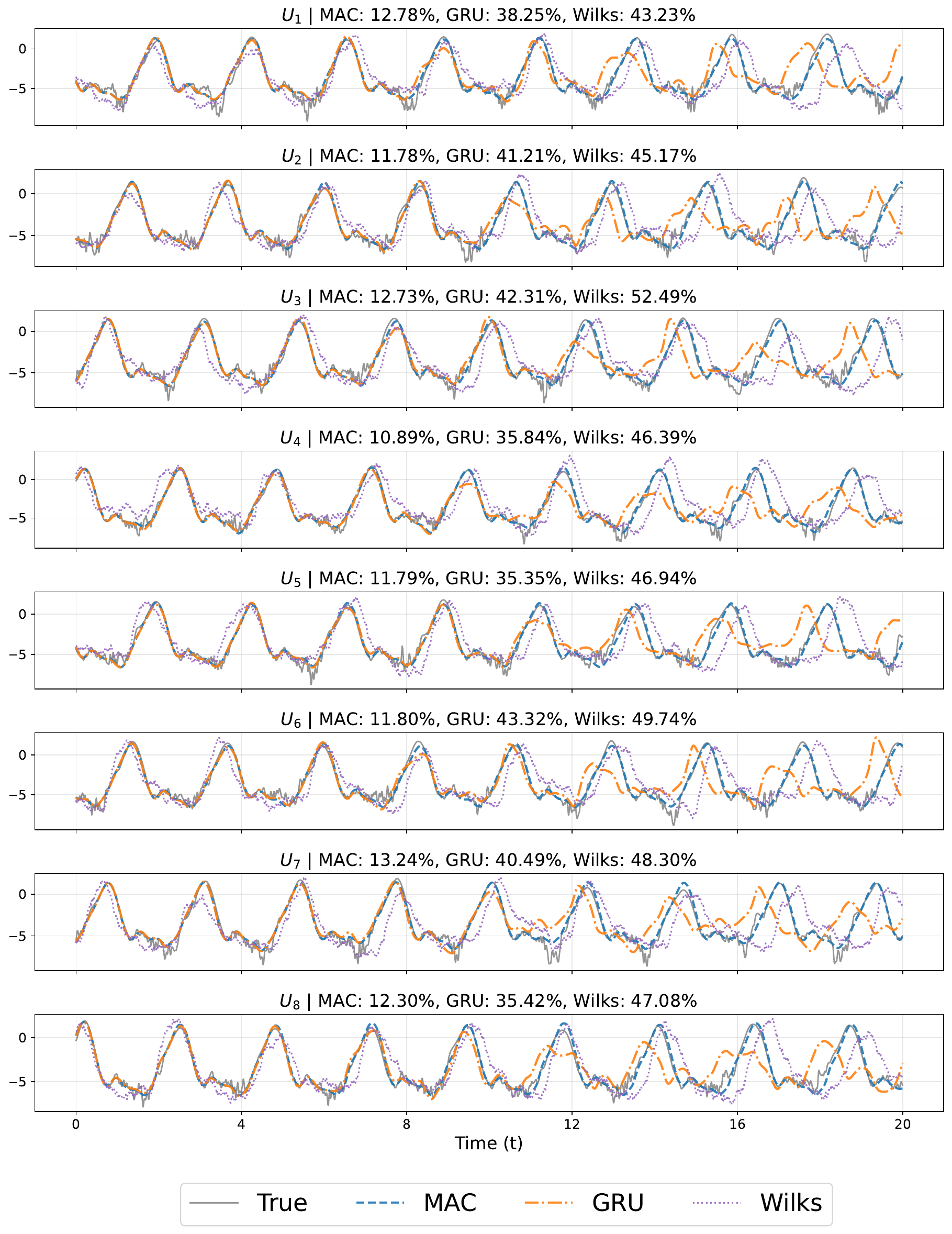}
    \caption{\label{fig_l96_unseen_closure}Comparison of closure-term predictions from the MAC model, the GRU-based model, and the Wilks method for the same representative unseen initial condition as in \Cref{fig_l96_unseen_resolved} over the time interval $[0,20]$. For each closure term, the relative $L^2$ error is also reported.}
\end{figure}

For the Lorenz '96 system, the main text reports resolved-variable rollout results for unseen initial conditions in \Cref{sec_results_l96}. 
Here in \Cref{fig_l96_unseen_closure}, we show the corresponding closure-term predictions, comparing the true closure terms with the predictions from the MAC model, the GRU-based model, and the Wilks method.

To complement the resolved-variable predictions reported in the main text for Lorenz '96, we also report summary statistics for the Lorenz '96 closure prediction errors.
\Cref{tab_l96_closure_errors} lists the mean relative $L^2$ errors of the closure terms predicted by the MAC model, the GRU-based model, and the Wilks method for each closure term $U_k$.
These values are computed across $100$ unseen-initial-condition test trajectories and averaged over 5 random seeds.
As shown in \Cref{tab_l96_closure_errors}, the MAC model achieves substantially lower mean errors than both the GRU-based model and the Wilks method for all eight closure terms.
\begin{table}[htbp]
\centering
\caption{\label{tab_l96_closure_errors}Relative $L^2$ error statistics for each closure term ($U_k$) in the unseen-initial-condition experiment. The mean and standard deviation are computed across $100$ test trajectories and averaged over 5 random seeds.}
\begin{tabular}{c|ccc}
\hline
Variable & MAC (\%) & GRU (\%) & Wilks (\%) \\
\hline
$U_1$ & $12.86 \pm 1.61$ & $25.81 \pm 7.39$ & $43.89 \pm 11.75$ \\
$U_2$ & $13.04 \pm 1.73$ & $25.73 \pm 7.52$ & $43.83 \pm 11.57$ \\
$U_3$ & $12.94 \pm 1.64$ & $25.36 \pm 7.48$ & $43.71 \pm 11.84$ \\
$U_4$ & $13.23 \pm 1.75$ & $25.97 \pm 7.62$ & $43.86 \pm 11.88$ \\
$U_5$ & $12.85 \pm 1.62$ & $24.60 \pm 7.25$ & $43.81 \pm 11.71$ \\
$U_6$ & $13.07 \pm 1.74$ & $25.75 \pm 7.46$ & $43.96 \pm 11.75$ \\
$U_7$ & $13.11 \pm 1.79$ & $26.01 \pm 7.71$ & $43.83 \pm 11.60$ \\
$U_8$ & $12.93 \pm 1.63$ & $25.38 \pm 7.44$ & $43.66 \pm 11.70$ \\
\hline
\end{tabular}
\end{table}

\subsubsection{Power-Grid System}
\label{apdx_closure_pg}

\begin{figure}[htbp]
    \centering
    \includegraphics[width=0.9\textwidth]{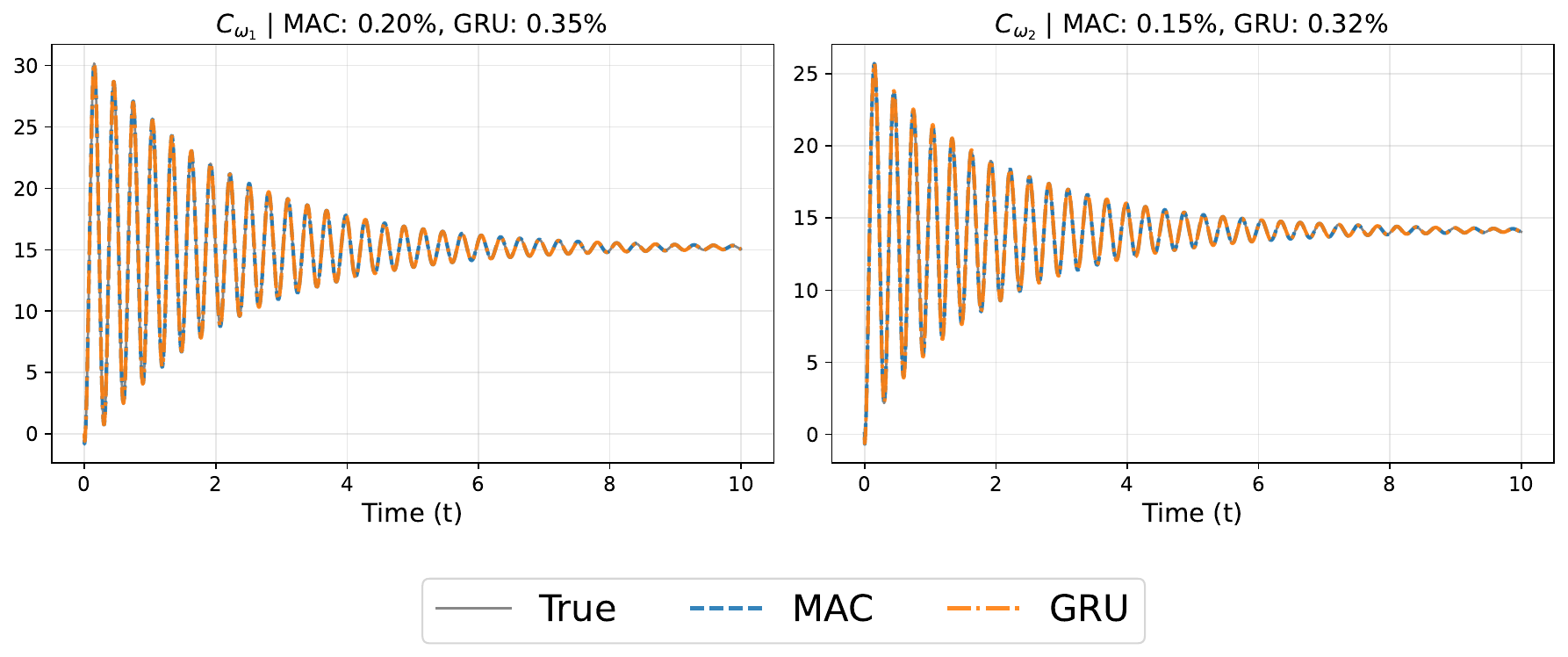}
    \caption{\label{fig_pg_closure}Comparison of closure-term predictions from the MAC model, the GRU-based model, and the true closure terms on the same representative in-distribution test initial condition as in \Cref{fig_pg_resolved} over the time interval $[0,10]$. For each closure term, the relative $L^2$ error over the time interval $[0,10]$ is also reported.}
\end{figure}

\begin{figure}[htbp]
    \centering
    \includegraphics[width=0.9\textwidth]{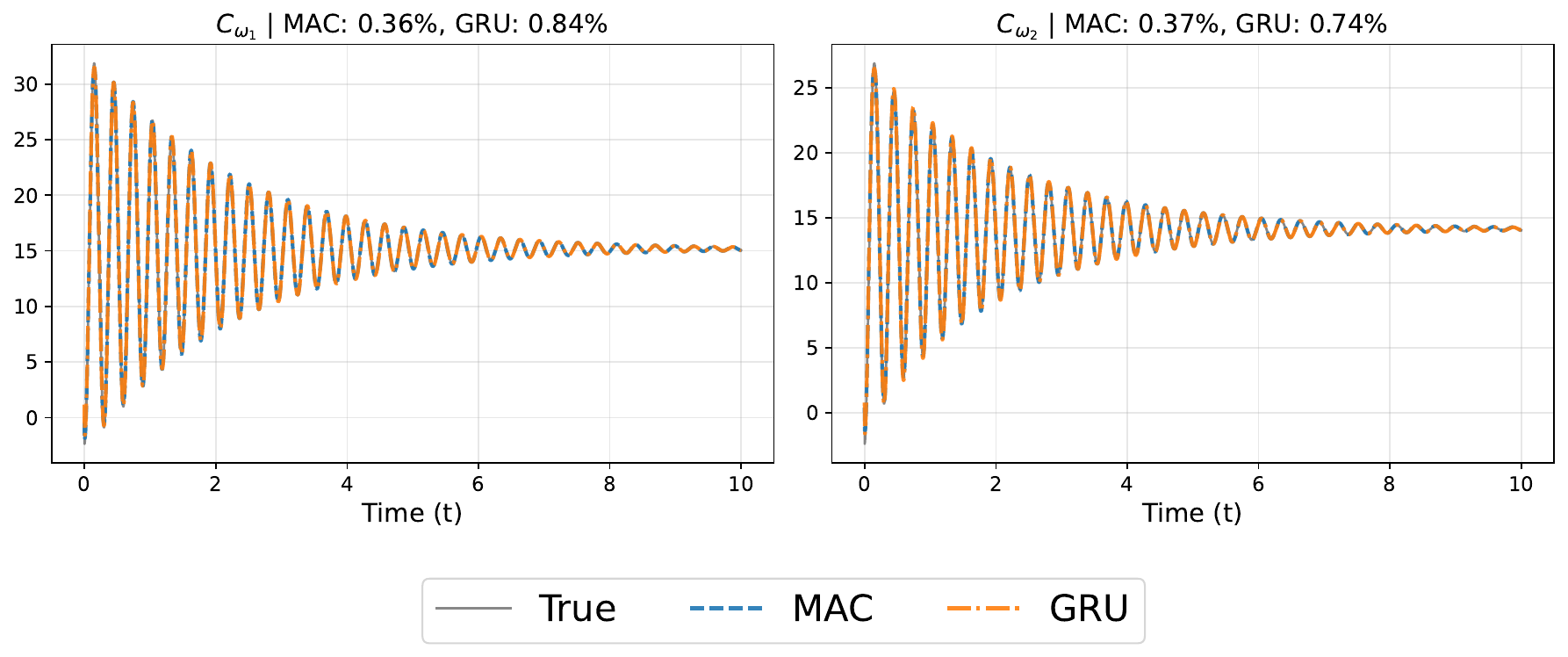}
    \caption{\label{fig_pg_ood_closure}Comparison of closure-term predictions from the MAC model, the GRU-based model, and the true closure terms on the same representative out-of-distribution test initial condition as in \Cref{fig_pg_ood_resolved} over the time interval $[0,10]$. For each closure term, the relative $L^2$ error over the time interval $[0,10]$ is also reported.}
\end{figure}

For the power-grid system, the main text reports resolved-variable rollout results for both the in-distribution and out-of-distribution test sets in \Cref{sec_results_pg}. 
Here, in \Cref{fig_pg_closure,fig_pg_ood_closure}, we show the corresponding closure-term predictions for the representative test initial conditions used in the main text. 
These plots compare the predicted closure terms produced by the MAC model and the GRU-based model with the true closure terms.

To complement the closure-term predictions for the representative test initial conditions, we also report summary statistics for the power-grid closure prediction errors. \Cref{tab_pg_closure_error} lists the mean relative $L^2$ errors of the closure terms predicted by the MAC model and the GRU-based model on the in-distribution and out-of-distribution test sets. These values are computed across all initial conditions in the corresponding test dataset and averaged over 5 random seeds. From \Cref{tab_pg_closure_error}, the MAC model consistently achieves lower closure prediction errors than the GRU-based model for both closure terms on both test sets. The error increase from the in-distribution test set to the out-of-distribution test set is also smaller for the MAC model, which is consistent with the improved resolved-variable rollout accuracy reported in the main text.

\begin{table}[htbp]
\centering
\caption{\label{tab_pg_closure_error}Relative $L^2$ error statistics for each closure term in the
power-grid system. The mean and standard deviation are computed across all initial
conditions in the corresponding test dataset and over 5 random seeds for the
MAC model and the GRU-based model.}
\begin{tabular}{c|c|cc}
\hline
Test set & Closure term & MAC & GRU \\
\hline
In-distribution
& $C_{\omega_1}$ & $0.1812\% \pm 0.0319\%$ & $0.3813\% \pm 0.1255\%$ \\
& $C_{\omega_2}$ & $0.1465\% \pm 0.0240\%$ & $0.3544\% \pm 0.1054\%$ \\
\hline
Out-of-distribution
& $C_{\omega_1}$ & $0.3396\% \pm 0.2264\%$ & $0.7315\% \pm 0.4453\%$ \\
& $C_{\omega_2}$ & $0.2988\% \pm 0.2104\%$ & $0.6529\% \pm 0.3740\%$ \\
\hline
\end{tabular}
\end{table}

\subsubsection{Korteweg--de Vries Equation}
\label{apdx_closure_kdv}

\begin{figure}[htbp]
    \centering
    \includegraphics[width=0.9\textwidth]{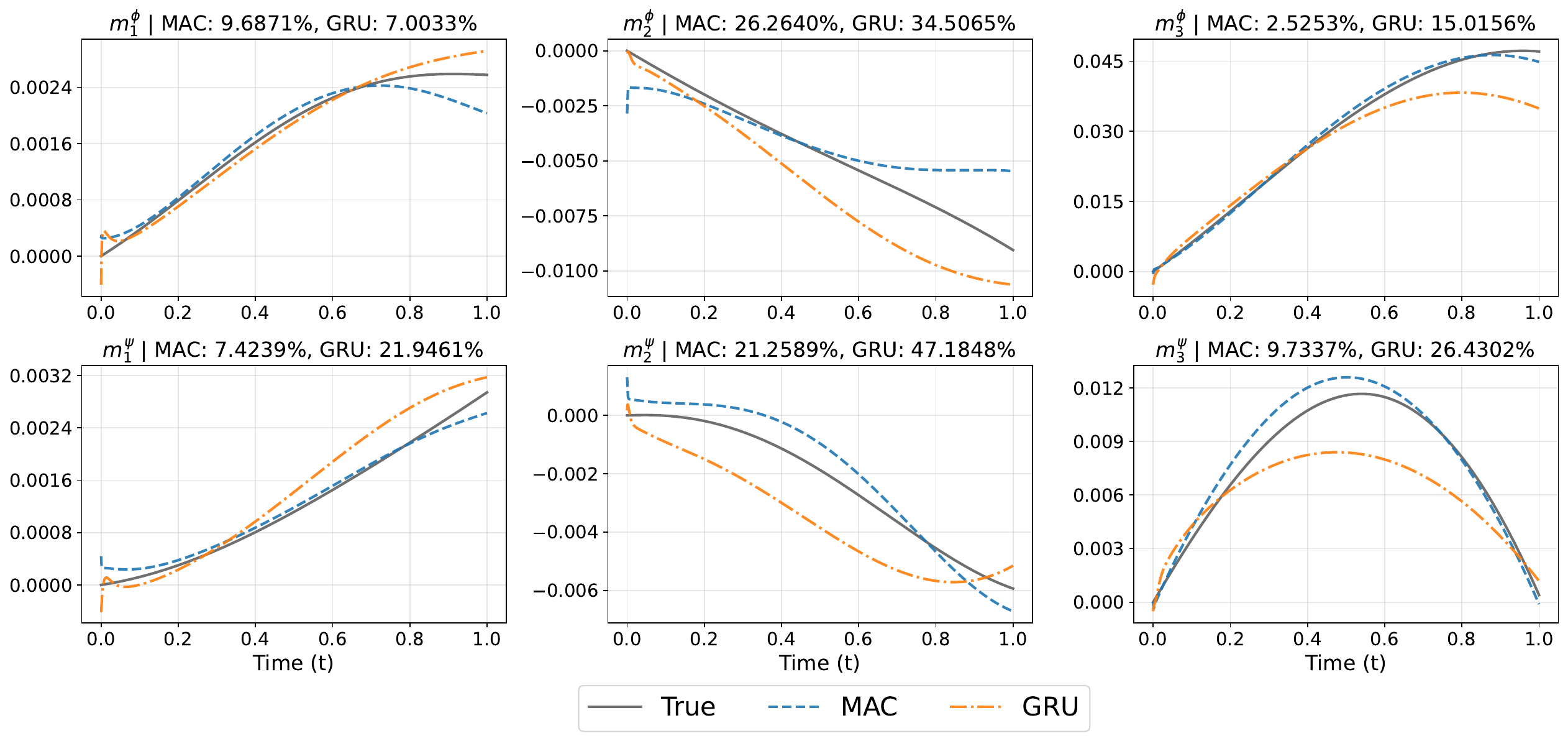}
    \caption{\label{fig_kdv_test_closure}Comparison of the closure-term predictions from the MAC model, the GRU-based model, and the true closure terms on the same representative test initial condition as in \Cref{fig_kdv_interp_resolved} over the temporal interpolation regime $[0,1]$. For each closure term, the relative $L^2$ error over the time interval $[0,1]$ is also reported.}
\end{figure}

For the Korteweg--de Vries equation, the main text reports autoregressive rollout results for the resolved Fourier modes in \Cref{sec_results_kdv}.
Here, in \Cref{fig_kdv_test_closure}, we show the corresponding closure-term predictions for the same representative test initial condition as in \Cref{fig_kdv_interp_resolved}.
The plot directly compares the predicted closure terms produced by the MAC model and the GRU-based model with the true closure data.

To complement the closure-term predictions for the representative test initial condition, we also report summary statistics for the KdV closure prediction errors.
\Cref{tab_kdv_closure_error} lists the relative $L^2$ error statistics of the closure terms predicted by the MAC model and the GRU-based model over the temporal interpolation interval $[0,1]$.
The mean and standard deviation are computed across $50$ test cases and over 5 random seeds.
Although the closure-term relative errors are larger than the resolved-variable relative errors reported in the main text, the predicted closure trajectories still capture the main trends of the true closure data.
As in the Burgers experiment, the relative errors can become large when the true closure terms have small magnitudes.
From \Cref{tab_kdv_closure_error}, the MAC model achieves lower mean closure prediction errors than the GRU-based model for all six closure components.

\begin{table}[htbp]
    \centering
    \caption{\label{tab_kdv_closure_error}Relative $L^2$ error statistics for each KdV closure term over the temporal interpolation regime $[0,1]$. The mean and standard deviation are computed across $50$ test cases and over 5 random seeds for the MAC model and the GRU-based model.}
    \begin{tabular}{c|cc}
    \hline
    Closure & MAC (\%) & GRU (\%) \\
    \hline
    $m_1^\phi$ & $60.6974 \pm 170.5109$ & $70.5424 \pm 198.9212$ \\
    $m_2^\phi$ & $47.5963 \pm 142.1611$ & $56.9540 \pm 174.5903$ \\
    $m_3^\phi$ & $23.3127 \pm 57.7234$ & $32.1758 \pm 89.9244$ \\
    \hline
    $m_1^\psi$ & $79.5410 \pm 276.9713$ & $109.7062 \pm 424.1927$ \\
    $m_2^\psi$ & $33.6720 \pm 73.8725$ & $43.0003 \pm 119.6127$ \\
    $m_3^\psi$ & $34.6505 \pm 121.0902$ & $43.4467 \pm 128.9046$ \\
    \hline
    \end{tabular}
\end{table}

\subsection{Additional Generalization Tests for the Korteweg--de Vries Equation}
\label{apdx_kdv_extrapolation}

As a complementary evaluation for the KdV case, we examine temporal extrapolation beyond the training time interval.
The MAC and GRU-based models are trained using trajectory data over the time interval $[0,1]$ and are subsequently rolled out over the extended time interval $[0,2]$ without further training.
Thus, the time interval $[1,2]$ provides a direct assessment of temporal extrapolation beyond the regime encountered during training.

\begin{figure}[htbp]
    \centering
    \includegraphics[width=16cm]{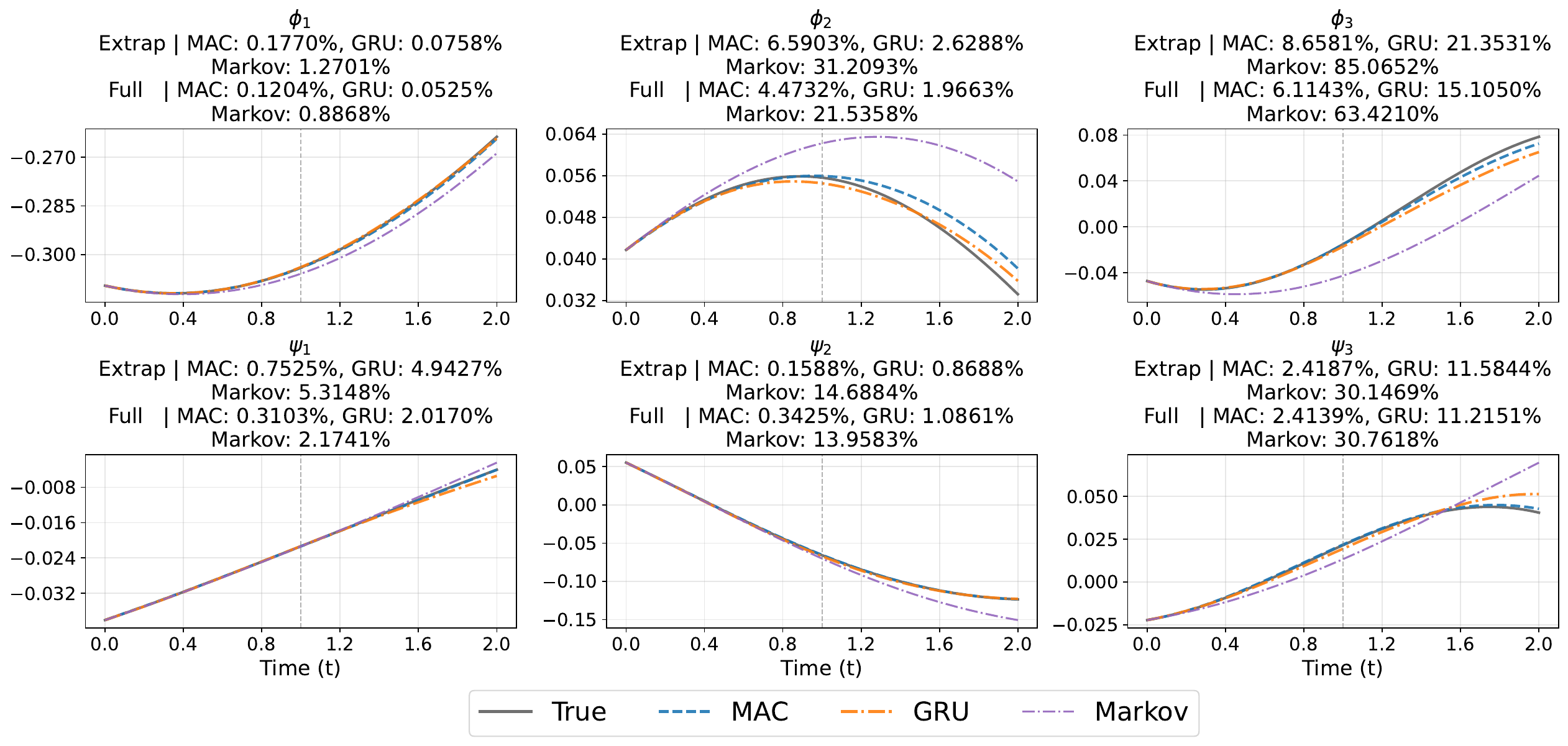}
    \caption{\label{fig_kdv_extrap_resolved}
    Comparison of resolved-mode predictions from the MAC model, the GRU-based model, and the Markovian reduced-order model on the same representative test initial condition as in \Cref{fig_kdv_interp_resolved} over the temporal extrapolation regime $[0,2]$.
    The sequence models are trained using data only over the time interval $[0,1]$, and the vertical dashed line at $t=1$ separates the training time interval from the pure temporal extrapolation interval.
    For each resolved Fourier mode, the relative $L^2$ errors over the pure extrapolation interval $[1,2]$ and the full rollout interval $[0,2]$ are also reported.}
\end{figure}

\Cref{fig_kdv_extrap_resolved} compares the predicted trajectories of the six resolved Fourier modes against the true solutions over the extended time interval $[0,2]$ for the same representative test initial condition as in \Cref{fig_kdv_interp_resolved}.
The vertical dashed line at $t=1$ marks the end of the training time interval and the beginning of temporal extrapolation.
For this representative test case, both the MAC model and the GRU-based model yield substantially lower extrapolation errors than the Markovian reduced-order model for all six resolved modes.
MAC also achieves lower extrapolation errors than the GRU-based model for most resolved Fourier modes.

To quantify the temporal extrapolation performance across the test dataset, \Cref{tab_kdv_extrap_resolved} reports the relative $L^2$ error statistics for all six resolved Fourier modes over both the full rollout interval $[0,2]$ and the pure extrapolation interval $[1,2]$.
The mean and standard deviation are computed across all initial conditions in the test dataset and over 5 random-seed runs.
Compared with prediction within the training time interval, all three models exhibit increased prediction errors on the extrapolation interval $[1,2]$.
Nevertheless, MAC achieves the lowest mean relative $L^2$ error for every resolved Fourier mode over both reported intervals.
The advantage over the GRU-based model is relatively modest for some modes but is consistent across all six resolved modes, while both the MAC model and the GRU-based model substantially outperform the Markovian reduced-order model, particularly for the higher Fourier modes.

\begin{table}[htbp]
\centering
\caption{\label{tab_kdv_extrap_resolved}
Relative $L^2$ error statistics for each resolved Fourier mode over the temporal extrapolation regime.
Results are reported on both the full rollout interval $[0,2]$ and the pure extrapolation interval $[1,2]$.
The mean and standard deviation are computed across all initial conditions in the test dataset and over 5 random seeds for the MAC model, the GRU-based model, and the Markovian reduced-order model.}
\resizebox{\textwidth}{!}{
\begin{tabular}{c|ccc|ccc}
\hline
\multirow{2}{*}{Mode}
& \multicolumn{3}{c|}{$[0,2]$}
& \multicolumn{3}{c}{$[1,2]$} \\
& MAC (\%) & GRU (\%) & Markov (\%)
& MAC (\%) & GRU (\%) & Markov (\%) \\
\hline

$\phi_1$
& $0.6790 \pm 0.7986$
& $0.7434 \pm 0.8858$
& $2.2052 \pm 2.3215$
& $1.0610 \pm 1.3492$
& $1.1554 \pm 1.4535$
& $3.3610 \pm 3.5280$ \\

$\phi_2$
& $2.4679 \pm 2.2184$
& $2.6089 \pm 2.2723$
& $13.2857 \pm 8.0747$
& $4.1080 \pm 4.1628$
& $4.3280 \pm 4.3514$
& $21.1367 \pm 14.5632$ \\

$\phi_3$
& $9.3209 \pm 9.2614$
& $9.9614 \pm 9.0988$
& $76.5371 \pm 66.3794$
& $13.3632 \pm 15.6678$
& $14.1550 \pm 15.6938$
& $108.1928 \pm 104.4421$ \\

\hline

$\psi_1$
& $0.9583 \pm 1.4749$
& $1.0766 \pm 1.5894$
& $2.5743 \pm 3.4968$
& $1.6170 \pm 2.5227$
& $1.8472 \pm 2.8238$
& $4.4082 \pm 6.6035$ \\

$\psi_2$
& $3.4541 \pm 4.4956$
& $4.1172 \pm 4.9785$
& $15.6181 \pm 17.6138$
& $6.3020 \pm 9.5878$
& $7.4894 \pm 10.7961$
& $27.4932 \pm 47.9034$ \\

$\psi_3$
& $9.0151 \pm 8.7408$
& $11.4494 \pm 10.1087$
& $88.5854 \pm 96.7406$
& $12.5615 \pm 13.5291$
& $15.5899 \pm 14.8691$
& $117.2969 \pm 141.3014$ \\

\hline
\end{tabular}
}
\end{table}

These results indicate that the advantage of the sequence-based closure models over the Markovian reduced-order model persists beyond the training time horizon, despite the increased difficulty of temporal extrapolation for the nonlinear and dispersive KdV dynamics.
Long-time prediction of nonlinear dispersive systems is known to be susceptible to accumulated prediction and phase errors~\autocite{Ranocha-2021,Lei-2026}, providing context for the increase in error beyond the training time interval.
Nevertheless, MAC maintains a consistent advantage over the GRU-based model across the resolved Fourier modes, while both the MAC model and the GRU-based model remain substantially more accurate than the Markovian reduced-order model.
Consistent with the temporal extrapolation results for Burgers' equation, these findings provide further evidence that MAC can maintain accurate closure modeling beyond the time horizon encountered during training across PDEs with distinctly different dynamical characteristics.

\begin{figure}[htbp]
    \centering
    \includegraphics[width=10cm]{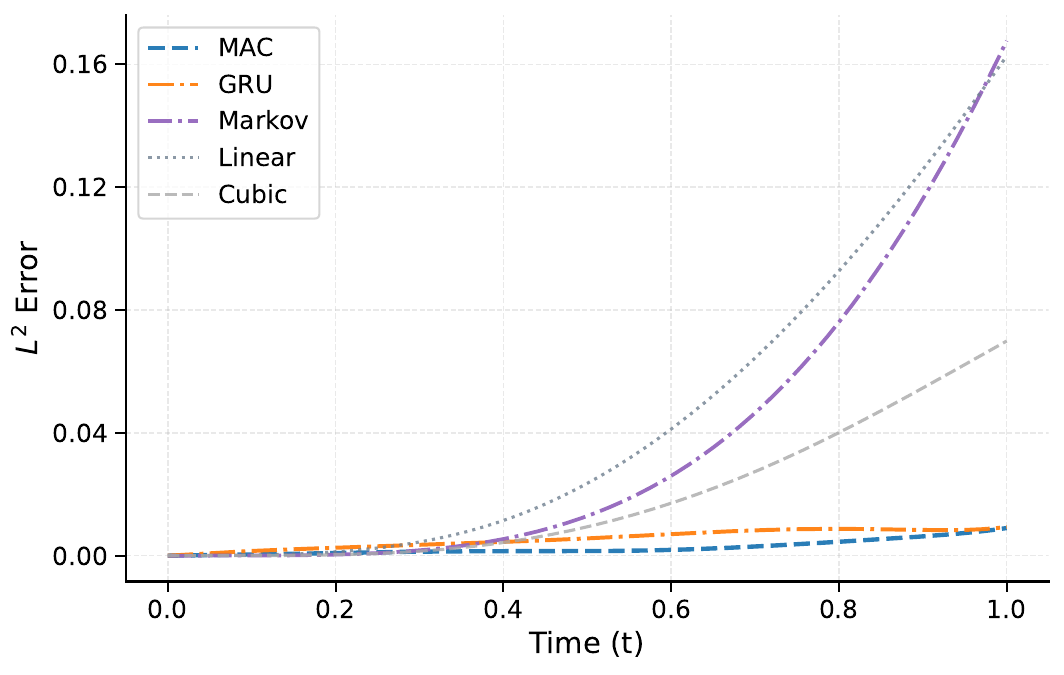}
    \caption{\label{fig_kdv_ood_case0}Evolution of the physical-space $L^2$ error for the projected $\ue^{\sin x}$ out-of-distribution initial condition over the rollout interval $[0,1]$, comparing the MAC model, the GRU-based model, the Markovian reduced-order model, and the linear and cubic memory models in~\autocite{Saad-2025}.}
\end{figure}

\begin{figure}[htbp]
    \centering
    \includegraphics[width=10cm]{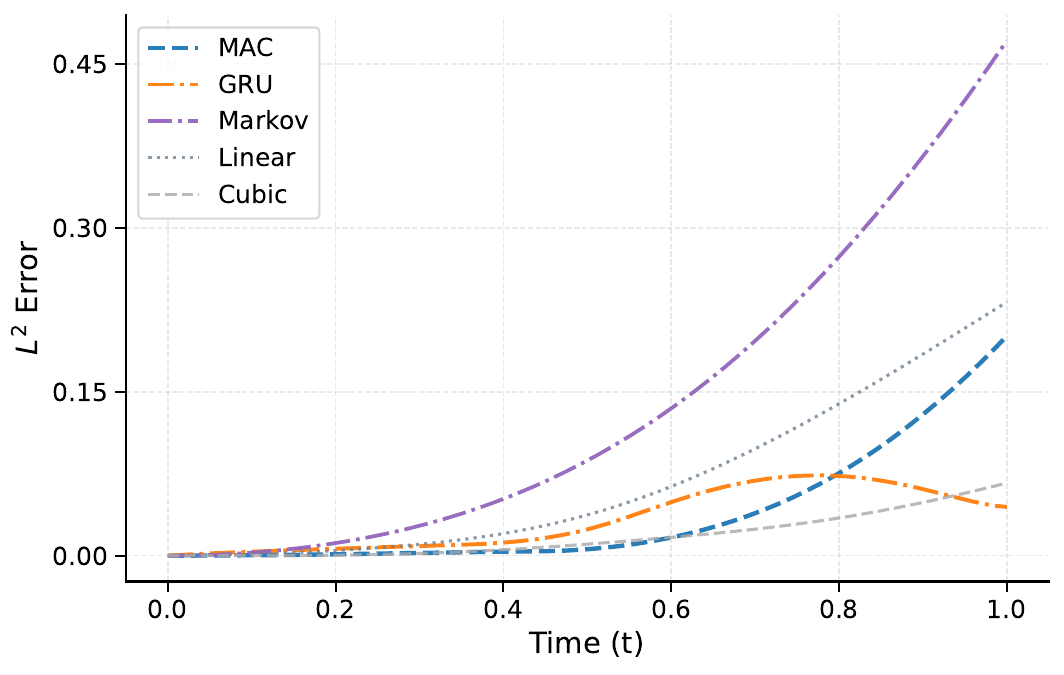}
    \caption{\label{fig_kdv_ood_case1}Evolution of the physical-space $L^2$ error for the out-of-distribution initial condition $u_0(x)=\ue^{\sin x}$ over the rollout interval $[0,1]$, comparing the MAC model, the GRU-based model, the Markovian reduced-order model, and the linear and cubic memory models in~\autocite{Saad-2025}.}
\end{figure}

In addition to temporal extrapolation, we further examine generalization with respect to the initial condition by considering two out-of-distribution initial conditions,
\begin{displaymath}
u_0(x)=\sin x
\qquad\text{and}\qquad
u_0(x)=\ue^{\sin x}.
\end{displaymath}
\Cref{fig_kdv_ood_case0,fig_kdv_ood_case1} compare the evolution of the physical-space $L^2$ error over the rollout interval $[0,1]$ for the MAC model, the GRU-based model, the Markovian reduced-order model, and the linear and cubic memory models in~\autocite{Saad-2025}.

From \Cref{fig_kdv_ood_case0,fig_kdv_ood_case1}, we observe that the MAC model maintains good prediction accuracy for both out-of-distribution initial conditions and achieves smaller errors than the competing models over most of the rollout interval. For the initial condition $u_0(x)=\sin x$, the MAC model provides the most accurate predictions throughout nearly the entire interval. For the initial condition $u_0(x)=\ue^{\sin x}$, the MAC model also performs well over most of the interval, although the cubic memory model achieves smaller errors near the end.

The cubic memory model provides highly accurate predictions, but the cost of evaluating its memory term grows rapidly as the autoregressive rollout progresses, whereas the MAC model maintains an essentially constant per-step inference cost. Taken together with the computational cost comparison in \Cref{apdx_runtime}, these results suggest that the MAC model can maintain accurate predictions for these out-of-distribution initial conditions while providing a favorable balance between accuracy and computational cost.

\subsection{Computational Cost Comparison}
\label{apdx_runtime}

The out-of-distribution experiments in \Cref{sec_results_burgers} compare the predictive accuracy of the MAC model with the linear and cubic memory models of~\autocite{Saad-2025}. Here, we complement those comparisons with a computational cost analysis that highlights how differently the inference cost of each time step grows as the autoregressive rollout progresses for these approaches.

\paragraph{Experimental setup.}
The linear and cubic memory models of~\autocite{Saad-2025} employ a time step of $\Delta t = 0.001$. Because these methods require the evaluation of a convolution integral whose support grows with the number of elapsed steps, using a finer time step would render their rollout prohibitively expensive. To enable a direct runtime comparison under identical conditions, we retrain the MAC model on the viscous Burgers' equation using the same time step $\Delta t = 0.001$. The training data are generated over the interval $[0,1]$, yielding training sequences of length $1000$.

\begin{figure}[htbp]
\centering
\includegraphics[width=0.6\textwidth]{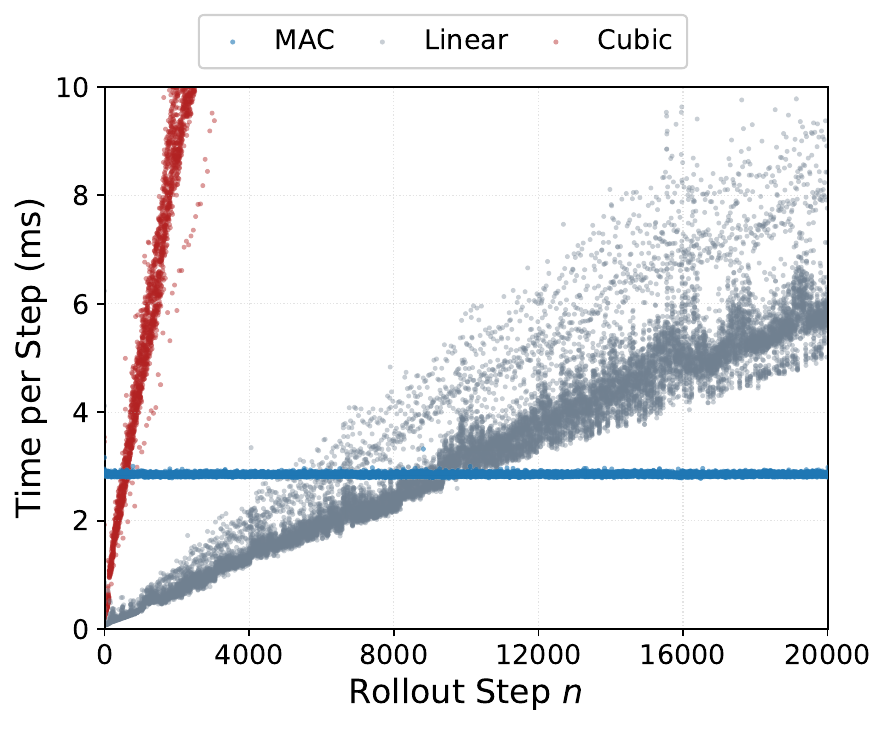}
\caption{Wall-clock time of each individual inference step plotted against the rollout step index for the MAC model and the linear and cubic memory models of~\autocite{Saad-2025} on the out-of-distribution initial condition $u_0(x)=\sin x$ over $[0,20]$ ($20,000$ steps at $\Delta t = 0.001$). The MAC model maintains an essentially constant per-step cost via recurrent state updates. The linear memory model exhibits linearly growing cost, while the cubic memory model grows even more rapidly, exceeding the plotted range within the first few thousand steps.}
\label{fig_runtime_comparison}
\end{figure}

\paragraph{Results.}
\Cref{fig_runtime_comparison} plots the wall-clock time of each individual inference step against the rollout step index for the MAC model and the linear and cubic memory models on the first out-of-distribution initial condition $u_0(x) = \sin x$, rolled out over the time interval $[0,20]$ ($20{,}000$ steps at $\Delta t = 0.001$).

The MAC model maintains an essentially constant per-step cost (approximately $3$\,ms) throughout the entire rollout. 
In contrast, the convolution-based methods exhibit growing per-step cost, since the memory integral evaluated at each step extends from the initial time to the current time. The linear memory model exhibits approximately linear growth, reaching roughly twice the per-step cost of the MAC model by step $20{,}000$. The cubic memory model grows even more rapidly, with its per-step cost exceeding the plotted range within the first few thousand steps. 

These results highlight the computational advantage of the MAC model's recurrent formulation over explicit convolution-based methods for long-time autoregressive rollout. 
By compressing the resolved history into a fixed-dimensional recurrent state, the MAC model avoids the growing cost of explicit convolution evaluation and maintains constant per-step inference cost regardless of the rollout horizon.

\clearpage
\section{Implementation Details}
\label{apdx_implementation}

This appendix provides additional implementation details for the numerical experiments reported in \Cref{sec_results}. 
While the main text focuses on the modeling framework and empirical findings, this appendix collects the practical choices needed to reproduce the experiments, including 
data generation, neural network architectures, model training hyperparameters, data normalization, noise injection schedules, and training loss design.

Unless otherwise stated, the same preprocessing, training protocol, and evaluation procedure are used across all sequence models considered in the comparison. 
In particular, the MAC model and the GRU-based sequence model are trained under comparable settings, with the number of trainable parameters controlled to be of the same order whenever possible. 
This setup ensures a fair comparison by avoiding differences in model capacity or optimization budget.

\subsection{Data Generation}
\label{apdx_data_generation}

\subsubsection{Viscous Burgers' equation}
\label{apdx_data_generation_burgers}

For the viscous Burgers' equation, the data generation follows the Fourier-space reduced-order construction described in \Cref{sec_results_burgers}. 
The trajectories of the full-order system are generated by solving the Fourier ODE system for the coefficients $\{\theta_k(t)\}_{k\in\mathbb{Z}}$ using a pseudospectral method. 
The spatial interval $[0,2\pi)$ is discretized using $N_x=2^{10}$ equispaced grid points, and the full Fourier system is advanced using a fourth-order Runge--Kutta method with time step $\Delta t=10^{-4}$. 
The viscosity coefficient used in our experiments is $\nu=0.1$.

For trajectories generated from random initial conditions, the resolved Fourier modes are initialized as described in the main text: both $\phi_k(0)$ and $\psi_k(0)$ are sampled uniformly from $(-\ue^{-k},\ue^{-k})$ for $k=1,2,\ldots,M$, while all other positive modes are set to zero. 
The corresponding negatively indexed modes are determined by the reality constraint on $u$, and the zero Fourier mode is taken to vanish.
For these random initial conditions, the full-order trajectories are generated over the time interval $[0,5]$.
For model training and validation, only the portion of these trajectories over the time interval $[0,1]$ is used.
The generated resolved trajectories are saved together with the corresponding closure data for the subsequent experiments.
In total, we generate $500$ trajectories from random initial conditions, of which $400$ are used for training, $50$ for validation, and $50$ for testing.

For the out-of-distribution tests, we use \(u_0(x)=\sin x\) and a projected \(\ue^{\sin x}\) initial condition. For the latter, only the first three positively indexed Fourier modes and their corresponding negatively indexed modes are retained, with the zero Fourier mode taken to vanish.
To evaluate long-time rollout performance under these out-of-distribution initial conditions, we solve the full Fourier system on the longer time interval $[0,20]$ and save the resulting trajectories of the resolved modes together with the corresponding closure trajectories for the subsequent experiments.

\subsubsection{Two-scale Lorenz '96 system}
\label{apdx_data_generation_l96}

For the two-scale Lorenz '96 system studied in \Cref{sec_results_l96}, the full slow--fast dynamics are simulated directly using a fourth-order Runge--Kutta method. 
The slow variables $\{X_k\}_{k=1}^{N}$ are treated as resolved variables, while the fast variables $\{Y_j\}_{j=1}^{JN}$ are regarded as unresolved degrees of freedom. 
The parameters used in the experiments are
\begin{displaymath}
    N=8,
    \qquad
    J=32,
    \qquad
    F=10,
    \qquad
    h=1,
    \qquad
    c=10,
    \qquad
    b=10.
\end{displaymath}
The interaction between the fast and slow variables enters the slow dynamics through the coupling term
\begin{displaymath}
    U_k(t)
    \coloneq
    -
    \frac{hc}{b}
    \sum_{j=J(k-1)+1}^{kJ}Y_j(t),
    \qquad
    k=1,2,\ldots,N.
\end{displaymath}

The full two-scale Lorenz '96 system is first simulated using the fine time step
\begin{displaymath}
    \Delta t_{\mathrm{fine}}=0.005.
\end{displaymath}
The recorded fine trajectory is then downsampled by a factor of $2$, giving the coarse time step
\begin{displaymath}
    \Delta t=0.01.
\end{displaymath}

A single long coarse trajectory of length $12001$ over the time interval $[0,120]$ is generated. 
This long trajectory is initialized randomly, with the initial slow variables sampled around $F/10$ and the fast variables initialized with small random values.
Before recording data, the system is advanced for $5000$ fine time steps as a warm-up period. 
After this warm-up, the first $10001$ coarse time steps over the time interval $[0,100]$ are used to construct the training dataset. 
More specifically, sliding windows of length $64$ with stride $1$ are extracted from the training segment and used as training sequences for the sequence models considered in this study. 
The segment over $[100,120]$ is used for validation during training.

In addition to the single long trajectory, $100$ independent test trajectories are generated from unseen random initial conditions. 
Similarly, for each unseen initial condition, the trajectory is initialized randomly, with the initial slow variables sampled around $F/10$ and the fast variables initialized with small random values.
Each test trajectory contains $2001$ coarse time steps over the time interval $[0,20]$.
The saved Lorenz '96 datasets contain the resolved slow variables $X$ and the corresponding coupling terms $U$, which are used in the subsequent autoregressive rollout experiments.

\subsubsection{Power-Grid System}
\label{apdx_data_generation_pg}

For the 3-bus DeMarco--Zheng power-grid system studied in \Cref{sec_results_pg}, the data generation follows the reduced-order construction described in the main text. The full five-dimensional system is simulated directly using a fourth-order Runge--Kutta method.
In the implementation, the full state is ordered as
\begin{displaymath}
    (\omega_1,\omega_2,\alpha_2,\alpha_3,V_3),
\end{displaymath}
with $(\omega_1,\omega_2,\alpha_2)$ retained as the resolved variables and $(\alpha_3,V_3)$ treated as unresolved variables. The right-hand side is evaluated using the explicit formula form of the power-grid equations. This is the same form used later during autoregressive rollout, so that the data-generation and rollout implementations are consistent.

The numerical parameter values are those of the 3-bus DeMarco--Zheng example in \autocite{Li-2022}. In particular, we use
\begin{displaymath}
\begin{split}
    &M_1=0.052,\qquad M_2=0.0531,\qquad
    D_1=D_2=0.05,\qquad D_3=0.005,\\
    &b_1=b_2=b_3=10,\qquad
    P_2=-2,\qquad P_3=3,\qquad Q_3=0.1,\\
    &\epsilon=5,\qquad V_1=V_2=0.9.
\end{split}
\end{displaymath}
For all generated trajectories, the unresolved variables are initialized at
\begin{displaymath}
    \alpha_3(0)=-0.3,
    \qquad
    V_3(0)=0.8.
\end{displaymath}
These same values are used as the fixed unresolved reference values in the Markovian reduced-order model.

The full system is integrated on the time interval $[0,10]$ using the fine time step
\begin{displaymath}
    \Delta t_{\rm fine}=5\times 10^{-5}.
\end{displaymath}
The simulated trajectories are then downsampled every $20$ fine steps, giving the coarse time step
\begin{displaymath}
    \Delta t=10^{-3}.
\end{displaymath}
Thus each complete trajectory contains $10001$ saved coarse time points for the resolved variables. At each saved time instant, we store the resolved variables, the unresolved variables, and the two nontrivial closure terms. The saved resolved-variable array has shape
\begin{displaymath}
    (n_\mathrm{traj},10001,3),
\end{displaymath}
corresponding to $(\omega_1,\omega_2,\alpha_2)$, while the saved closure array has shape
\begin{displaymath}
    (n_\mathrm{traj},10001,2),
\end{displaymath}
corresponding to $(C_{\omega_1},C_{\omega_2})$. Here, $n_\mathrm{traj}$ denotes the number of trajectories in the respective dataset. The closure target is computed as the difference between the resolved part of the full right-hand side and the Markovian resolved right-hand side obtained by fixing $\alpha_3=-0.3$ and $V_3=0.8$. Since the equation for $\alpha_2$ depends only on $\omega_1$ and $\omega_2$, only the closure terms associated with the $\omega_1$ and $\omega_2$ equations are used as learning targets.

For the in-distribution datasets, the resolved initial conditions are sampled independently
from
\begin{displaymath}
    \omega_1(0),\omega_2(0)\in[-0.05,0.05],
    \qquad
    \alpha_2(0)\in[-0.20,-0.12],
\end{displaymath}
while the unresolved initial conditions are fixed as above. We generate $512$ training trajectories, $64$ validation trajectories, and $100$ in-distribution test trajectories. The training trajectories are used for model optimization, the validation trajectories are used for checkpoint selection, and the in-distribution test trajectories are reserved for final evaluation.

For the out-of-distribution test, we use a wider range of resolved initial conditions,
\begin{displaymath}
    \omega_1(0),\omega_2(0)\in[-0.10,0.10],
    \qquad
    \alpha_2(0)\in[-0.24,-0.08],
\end{displaymath}
again with $\alpha_3(0)=-0.3$ and $V_3(0)=0.8$. This out-of-distribution test set contains $50$ trajectories and is used only for evaluation.

\subsubsection{Korteweg--de Vries equation}
\label{apdx_data_generation_kdv}

The data generation for the Korteweg--de Vries equation follows the Fourier-space reduced-order construction described in \Cref{sec_results_kdv}.  
The trajectories of the full-order system are generated by solving the Fourier ODE system for the coefficients $\{\theta_k(t)\}_{k\in\mathbb{Z}}$ using a pseudospectral method. 
The spatial interval $[0,2\pi)$ is discretized using $N_x=2^{10}$ equispaced grid points, and the full Fourier system is advanced using a Cox--Matthews exponential time-differencing fourth-order Runge--Kutta method with time step $\Delta t=10^{-4}$. 
The dispersion coefficient used in our experiments is $\delta=0.1$.

For trajectories generated from random initial conditions, the resolved Fourier modes are initialized as described in the main text: both $\phi_k(0)$ and $\psi_k(0)$ are sampled uniformly from $(-\ue^{-k},\ue^{-k})$ for $k=1,2,\ldots,M$, while all other positive modes are set to zero. 
The corresponding negatively indexed modes are determined by the reality constraint on $u$, and the zero Fourier mode is taken to vanish.
For these random initial conditions, the full-order trajectories are generated over the time interval $[0,2]$. Only the portion over the time interval $[0,1]$ is used for training, validation, and the temporal interpolation experiments in the main text, while the continuation over $[1,2]$ is used solely for the temporal extrapolation evaluation reported in \Cref{apdx_kdv_extrapolation}.
The generated resolved trajectories are saved together with the corresponding closure data for the subsequent experiments.
In total, we generate $500$ trajectories from random initial conditions, of which $400$ are used for training, $50$ for validation, and $50$ for testing.

For the tests on out-of-distribution initial conditions in \Cref{apdx_kdv_extrapolation}, we use \(u_0(x)=\sin x\) and a projected \(\ue^{\sin x}\) initial condition. For the latter, only the first three positively indexed Fourier modes and their corresponding negatively indexed modes are retained, with the zero Fourier mode taken to vanish.
To evaluate long-time rollout performance under these out-of-distribution initial conditions, we solve the full Fourier system on the time interval $[0,1]$ and save the resulting trajectories of the resolved modes together with the corresponding closure trajectories for the subsequent experiments.

\subsection{Neural Network Architectures}
\label{apdx_neural_network_architectures}

In the Mamba-Assisted Closure (MAC) model, the Mamba-based sequence model predicts closure terms from the resolved variables. These predicted closure terms are then incorporated into the reduced-order dynamics, which are advanced in time to generate autoregressive resolved trajectories. When the MAC model is applied to different benchmark problems, the reduced-order dynamics differ, but the same general neural-network architecture is used for the Mamba-based sequence model.

For each benchmark problem, we use problem-dependent input and output dimensions, as well as problem-dependent choices of depth and width for the Mamba-based and GRU-based sequence models. These choices are reported in \Cref{apdx_model_training_hyperparameters}. For Burgers' equation, the input consists of the resolved Fourier modes and has dimension $2M$, corresponding to the real and imaginary parts of the resolved modes, where $M=3$ in our study. For the Lorenz '96 system, the input consists of the resolved slow variables and has dimension $N=8$. For the power-grid system, the input consists of the resolved variables $(\omega_1,\omega_2,\alpha_2)$ and has dimension $3$, while the output consists of the two nontrivial closure components $(C_{\omega_1},C_{\omega_2})$.
For the Korteweg--de Vries equation, as for Burgers' equation, the input and output dimensions are both $2M=6$.
The output dimension is chosen to match the dimension of the closure term being learned.

In this architecture, the Mamba-based sequence model first applies a linear input projection from the input dimension to a hidden dimension $d_{\mathrm{model}}$, followed by a $\operatorname{SiLU}$ activation.  The resulting hidden sequence is then passed through a stack of residual Mamba blocks. Each residual Mamba block consists of a normalization layer, a Mamba layer, and a residual connection:
\begin{displaymath}
    h \mapsto h + \operatorname{Mamba}\bigl(\operatorname{Norm}(h)\bigr).
\end{displaymath}
Here, the Mamba layer is the module implemented using the Python package \texttt{mamba-ssm}, available from \href{https://pypi.org/project/mamba-ssm/}{PyPI}.
After the stacked residual Mamba blocks, a final normalization layer is applied, followed by a two-layer output projection:
\begin{displaymath}
    h \mapsto W_2\,\sigma(W_1 h),
\end{displaymath}
where $\sigma$ denotes the $\operatorname{SiLU}$ activation.

For comparison, the GRU-based sequence model uses the same input and output projection structure. The main difference is that the residual Mamba blocks are replaced by residual GRU blocks. Each residual GRU block consists of a normalization layer, a GRU layer, and a residual connection:
\begin{displaymath}
    h \mapsto h + \operatorname{GRU}\bigl(\operatorname{Norm}(h)\bigr).
\end{displaymath}
Here, the GRU layer is the standard module available in PyTorch. Thus, the GRU-based sequence model preserves the same overall sequence-to-sequence structure as the Mamba-based sequence model.

In both architectures, the normalization layer is applied inside each residual block before the Mamba or GRU layer, and another normalization layer is applied before the final output projection. 
The specific choices of depth, model hidden dimension, Mamba state dimension, convolution width, and GRU width are given in \Cref{apdx_model_training_hyperparameters}. 
These choices are selected so that the Mamba-based sequence model and the GRU-based sequence model have comparable numbers of trainable parameters whenever possible.

\subsection{Model and Training Hyperparameters}
\label{apdx_model_training_hyperparameters}

Each experiment was conducted using a single NVIDIA H100 GPU on the PNNL Deception HPC cluster, without multi-GPU parallelization or distributed training. All models are implemented in PyTorch~2.8.0 (CUDA~12.9), with the Mamba layers implemented using the \texttt{mamba-ssm} package (version~2.2.5).

\subsubsection{Burgers' Equation}
\label{apdx_model_training_hyperparameters_burgers}

For the Burgers experiments, the Mamba-based sequence model takes the resolved Fourier modes as input. 
Since $M=3$ complex resolved positively indexed Fourier modes are retained, the input and output dimensions are both $2M=6$, corresponding to the real and imaginary parts of the resolved modes. 
The model hyperparameters used in the full training runs are summarized in \Cref{tab_burgers_model_hyperparameters}. 
The main training hyperparameters are summarized in \Cref{tab_burgers_training_hyperparameters}. 
The noise injection schedule and loss design are described separately in \Cref{apdx_noise_injection} and \Cref{apdx_training_losses}.

\begin{table}[htbp]
    \centering
    \caption{\label{tab_burgers_model_hyperparameters}Model hyperparameters for the Burgers experiments.}
    \begin{tabular}{lll}
        \hline
        Hyperparameter & Mamba-based model & GRU-based model \\
        \hline
        Input dimension & $2M=6$ & $2M=6$ \\
        Output dimension & $2M=6$ & $2M=6$ \\
        Number of residual blocks & $4$ & $4$ \\
        Hidden dimension $d_{\mathrm{model}}$ & $22$ & $28$ \\
        Mamba expansion factor & $2$ & -- \\
        Mamba convolution width & $4$ & -- \\
        Mamba state dimension & $12$ & -- \\
        GRU layers per residual block & -- & $1$ \\
        Linear bias in Mamba layer & no & -- \\
        Convolution bias in Mamba layer & yes & -- \\
        Normalization type & LayerNorm & LayerNorm \\
        Activation function & $\operatorname{SiLU}$ & $\operatorname{SiLU}$ \\
        Number of trainable parameters & \num{20906} & \num{20950} \\
        \hline
    \end{tabular}
\end{table}

\begin{table}[htbp]
    \centering
    \caption{\label{tab_burgers_training_hyperparameters}Training hyperparameters for the Burgers experiments.}
    \begin{tabular}{ll}
        \hline
        Hyperparameter & Value \\
        \hline
        Number of training steps & \num{30000} \\
        Batch size & $96$ \\
        Optimizer & AdamW \\
        Weight decay & $1\times 10^{-4}$ \\
        Learning-rate scheduler & OneCycleLR \\
        Maximum learning rate & $1.5\times 10^{-3}$ \\
        Warm-up fraction in scheduler & $0.12$ \\
        Initial learning-rate divisor & $50$ \\
        Final learning-rate divisor & $800$ \\
        Annealing strategy & Cosine \\
        \hline
    \end{tabular}
\end{table}
\clearpage

\subsubsection{Lorenz '96 System}
\label{apdx_model_training_hyperparameters_l96}

For the Lorenz '96 experiments, the sequence model takes the resolved slow variables as input. 
Since $N=8$ slow variables are treated as resolved variables, the input and output dimensions are both $N=8$. 
The model hyperparameters used in the full training runs are summarized in \Cref{tab_l96_model_hyperparameters}. 

\begin{table}[htbp]
    \centering
    \caption{\label{tab_l96_model_hyperparameters}Model hyperparameters for the Lorenz '96 experiments.}
    \begin{tabular}{lll}
        \hline
        Hyperparameter & Mamba-based model & GRU-based model \\
        \hline
        Input dimension $d_{\mathrm{in}}$ & $N=8$ & $N=8$ \\
        Output dimension & $N=8$ & $N=8$ \\
        Number of residual blocks & $3$ & $4$ \\
        Hidden dimension $d_{\mathrm{model}}$ & $63$ & $62$ \\
        Mamba expansion factor & $2$ & -- \\
        Mamba convolution width & $4$ & -- \\
        Mamba state dimension & $16$ & -- \\
        GRU layers per residual block & -- & $1$ \\
        Linear bias in Mamba layer & no & -- \\
        Convolution bias in Mamba layer & yes & -- \\
        Normalization type & LayerNorm & LayerNorm \\
        Activation function & $\operatorname{SiLU}$ & $\operatorname{SiLU}$ \\
        Number of trainable parameters & \num{100871} & \num{99332} \\
        \hline
    \end{tabular}
\end{table}

The main training hyperparameters are summarized in \Cref{tab_l96_training_hyperparameters}. 
The noise injection schedule and loss design are described separately in \Cref{apdx_noise_injection} and \Cref{apdx_training_losses}.

\begin{table}[htbp]
    \centering
    \caption{\label{tab_l96_training_hyperparameters}Training hyperparameters for the Lorenz '96 experiments.}
    \begin{tabular}{ll}
        \hline
        Hyperparameter & Value \\
        \hline
        Number of training steps & \num{1000} \\
        Batch size & $512$ \\
        Optimizer & AdamW \\
        Weight decay & $1\times 10^{-2}$ \\
        Learning-rate scheduler & OneCycleLR \\
        Maximum learning rate & $1\times 10^{-3}$ \\
        Warm-up fraction in scheduler & $0.1$ \\
        Initial learning-rate divisor & $25$ \\
        Final learning-rate divisor & $500$ \\
        Annealing strategy & Cosine \\
        \hline
    \end{tabular}
\end{table}

\subsubsection{Power-Grid System}
\label{apdx_model_training_hyperparameters_pg}

For the power-grid experiments, the sequence model takes the resolved variables $(\omega_1,\omega_2,\alpha_2)$ as input and predicts the two nontrivial closure components $(C_{\omega_1},C_{\omega_2})$. Therefore, the input dimension is $3$ and the output dimension is $2$. The model hyperparameters used in the full training runs are summarized in \Cref{tab_pg_model_hyperparameters}.

\begin{table}[htbp]
\centering
\caption{\label{tab_pg_model_hyperparameters}Model hyperparameters for the power-grid experiments.}
\begin{tabular}{c|cc}
\hline
Hyperparameter & Mamba-based model & GRU-based model \\
\hline
Input dimension $d_{\rm in}$ & $3$ & $3$ \\
Output dimension $d_{\rm out}$ & $2$ & $2$ \\
Number of residual blocks & $3$ & $3$ \\
Hidden dimension $d_{\rm model}$ & $10$ & $16$ \\
Mamba expansion factor & $2$ & -- \\
Mamba convolution width & $4$ & -- \\
Mamba state dimension & $16$ & -- \\
GRU layers per residual block & -- & $1$ \\
Linear bias in Mamba layer & no & -- \\
Convolution bias in Mamba layer & yes & -- \\
Normalization type & LayerNorm & LayerNorm \\
Activation function & $\operatorname{SiLU}$ & $\operatorname{SiLU}$\\
Number of trainable parameters & \num{5472} & \num{5394} \\
\hline
\end{tabular}
\end{table}

The main training hyperparameters are summarized in \Cref{tab_pg_training_hyperparameters}. 
The noise injection schedule is described in \Cref{apdx_noise_injection}.

\begin{table}[htbp]
\centering
\caption{\label{tab_pg_training_hyperparameters}Training hyperparameters for the power-grid experiments.}
\begin{tabular}{c|c}
\hline
Hyperparameter & Value \\
\hline
Number of training steps & \num{10000} \\
Batch size & $96$ \\
Optimizer & AdamW \\
Weight decay & $1\times 10^{-4}$ \\
Learning-rate scheduler & OneCycleLR \\
Maximum learning rate & $1.5\times 10^{-3}$ \\
Warm-up fraction in scheduler & $0.12$ \\
Initial learning-rate divisor & $50$ \\
Final learning-rate divisor & $800$ \\
Annealing strategy & Cosine \\
\hline
\end{tabular}
\end{table}

\subsubsection{Korteweg--de Vries Equation}
\label{apdx_model_training_hyperparameters_kdv}

For the Korteweg--de Vries experiments, the Mamba-based sequence model takes the resolved Fourier modes as input. 
Since $M=3$ complex resolved positively indexed Fourier modes are retained, the input and output dimensions are both $2M=6$, corresponding to the real and imaginary parts of the resolved modes. 
The model hyperparameters used in the full training runs are summarized in \Cref{tab_kdv_model_hyperparameters}. 
The main training hyperparameters are summarized in \Cref{tab_kdv_training_hyperparameters}. 

\begin{table}[htbp]
    \centering
    \caption{\label{tab_kdv_model_hyperparameters}Model hyperparameters for the Korteweg--de Vries experiments.}
    \begin{tabular}{lll}
        \hline
        Hyperparameter & Mamba-based model & GRU-based model \\
        \hline
        Input dimension & $2M=6$ & $2M=6$ \\
        Output dimension & $2M=6$ & $2M=6$ \\
        Number of residual blocks & $2$ & $2$ \\
        Hidden dimension $d_{\mathrm{model}}$ & $24$ & $30$ \\
        Mamba expansion factor & $2$ & -- \\
        Mamba convolution width & $4$ & -- \\
        Mamba state dimension & $12$ & -- \\
        GRU layers per residual block & -- & $1$ \\
        Linear bias in Mamba layer & no & -- \\
        Convolution bias in Mamba layer & yes & -- \\
        Normalization type & LayerNorm & LayerNorm \\
        Activation function & $\operatorname{SiLU}$ & $\operatorname{SiLU}$ \\
        Number of trainable parameters & \num{12486} & \num{12666} \\
        \hline
    \end{tabular}
\end{table}

\begin{table}[htbp]
    \centering
    \caption{\label{tab_kdv_training_hyperparameters}Training hyperparameters for the Korteweg--de Vries experiments.}
    \begin{tabular}{ll}
        \hline
        Hyperparameter & Value \\
        \hline
        Number of training steps & \num{30000} \\
        Batch size & $96$ \\
        Optimizer & AdamW \\
        Weight decay & $1\times 10^{-4}$ \\
        Learning-rate scheduler & OneCycleLR \\
        Maximum learning rate & $1.5\times 10^{-3}$ \\
        Warm-up fraction in scheduler & $0.12$ \\
        Initial learning-rate divisor & $50$ \\
        Final learning-rate divisor & $800$ \\
        Annealing strategy & Cosine \\
        \hline
    \end{tabular}
\end{table}

\subsection{Data Normalization}
\label{apdx_data_normalization}

Different normalization strategies are used for different benchmark problems, following the scale and structure of the corresponding data. 
In all cases, the normalization statistics are computed only from the training data and then reused for validation, testing, and autoregressive rollout experiments.

For Burgers' equation, we normalize the resolved Fourier modes
$\{\phi_k,\psi_k\}_{k=1}^{M}$ and the closure data 
$\{m_k^\phi,m_k^\psi\}_{k=1}^{M}$ separately using component-wise max-absolute scaling. 
More specifically, for each resolved mode $k=1,\ldots,M$, the scaling factors are computed over all samples and all time points in the training dataset as
\begin{displaymath}
    s_{\phi,k} = \max_{\text{training data}} \abs{\phi_k(t)},
    \qquad
    s_{\psi,k} = \max_{\text{training data}} \abs{\psi_k(t)},
\end{displaymath}
and
\begin{displaymath}
    s_{\phi,k}^{(m)} = \max_{\text{training data}} \abs{m_k^\phi(t)},
    \qquad
    s_{\psi,k}^{(m)}  = \max_{\text{training data}} \abs{m_k^\psi(t)}.
\end{displaymath}
The normalized quantities used by the sequence model at time $t$ are then
\begin{displaymath}
    \biggl\{
    \frac{\phi_k(t)}{s_{\phi,k}},
    \frac{\psi_k(t)}{s_{\psi,k}}
    \biggr\}_{k=1}^M,
\end{displaymath}
and
\begin{displaymath}
    \biggl\{
    \frac{m_k^\phi(t)}{s^{(m)}_{\phi,k}},
    \frac{m_k^\psi(t)}{s^{(m)}_{\psi,k}}
    \biggr\}_{k=1}^M.
\end{displaymath}

For the Lorenz '96 system, we normalize the resolved slow variables 
$\{X_k\}_{k=1}^{N}$ and the closure terms $\{U_k\}_{k=1}^{N}$ component-wise. 
For each $k=1,\ldots,N$, the means $\mu_{X,k}$ and $\mu_{U,k}$, as well as the standard deviations $\sigma_{X,k}$ and $\sigma_{U,k}$, are computed over all samples and all time points in the training dataset.
The normalized quantities used by the sequence model at time $t$ are then
\begin{displaymath}
    \biggl\{
    \frac{X_k(t)-\mu_{X,k}}{\sigma_{X,k}}
    \biggr\}_{k=1}^{N},
\end{displaymath}
and
\begin{displaymath}
    \biggl\{
    \frac{U_k(t)-\mu_{U,k}}{\sigma_{U,k}}
    \biggr\}_{k=1}^{N}.
\end{displaymath}

The power-grid experiment uses the same component-wise mean-standard-deviation normalization as the Lorenz '96 system, with statistics computed from the training trajectories for the resolved variables $(\omega_1,\omega_2,\alpha_2)$ and the closure terms $(C_{\omega_1},C_{\omega_2})$.
The Korteweg--de Vries experiment uses the same component-wise max-absolute scaling strategy as Burgers' equation, with separate scaling factors computed from the training data for the resolved Fourier modes and the corresponding closure terms.

\subsection{Noise Injection}
\label{apdx_noise_injection}

During training, we inject small additive noise into the normalized input sequences to improve the robustness of the trained sequence model during autoregressive rollout. 
Unless otherwise stated, the same noise-injection schedule is used in the experiments reported in this work.

Let $\xi$ denote the normalized resolved input sequence used by the sequence model. 
When noise injection is active, the model input is replaced by
\begin{displaymath}
    \xi_{\mathrm{noisy}} = \xi + \varepsilon,
    \qquad
    \varepsilon \sim \mathcal{N}(0,\sigma_{\mathrm{noise}}^2 I),
\end{displaymath}
where the noise is sampled independently with the same shape as $\xi$.
In the experiments, we use
\begin{displaymath}
    \sigma_{\mathrm{noise}} = 5\times 10^{-4}.
\end{displaymath}

The noise level is scheduled over the training process in three phases. 
Let $S$ be the total number of training steps. 
For the first $10\%$ of training steps, no noise is injected. 
For the middle phase, from $0.1S$ to $0.8S$, the above Gaussian noise is added to the normalized input sequence. 
For the final $20\%$ of training steps, the noise is turned off again. 
Thus the schedule is
\begin{displaymath}
    \sigma_{\mathrm{noise}}(s)
    =
    \begin{cases}
        0, & s \leqslant 0.1S, \\
        5\times 10^{-4}, & 0.1S < s \leqslant 0.8S, \\
        0, & 0.8S < s \leqslant S.
    \end{cases}
\end{displaymath}

This schedule is used to separate the early fitting stage, the robustness-enhancing stage, and the final refinement stage. 
The noise is applied only to the model input during training; the training targets are not modified. 

\subsection{Training Loss Choices}
\label{apdx_training_losses}

When training the Mamba-based sequence model within the MAC framework, the choice of training loss can have a significant effect on the quality of the learned closure and the resulting autoregressive rollout. Across the four benchmark problems, we use two training-loss formulations. The Burgers', power-grid, and KdV experiments employ direct supervision on the closure terms, while the Lorenz '96 experiment uses a multi-step resolved-variable rollout loss. In this subsection, we discuss Burgers' equation and the Lorenz '96 system as representative examples of these two formulations and examine alternative loss designs tested for each problem. These examples illustrate that the appropriate loss design can depend on the dynamics of the benchmark problem.

\subsubsection{Burgers' Equation}

When training the Mamba-based sequence model for Burgers' equation, the input sequence consists of the normalized resolved Fourier modes.
The sequence model outputs the (normalized) predicted closure terms at the corresponding time points.
These predicted closure terms are compared with the true (normalized) closure terms using a mean squared error loss, which is used as the training objective.
This direct closure loss is sufficient to obtain accurate predictions for both the resolved Fourier modes and the closure terms, as shown in \Cref{sec_results_burgers,apdx_closure_burgers}.

\begin{figure}[htbp]
    \centering
    \includegraphics[width=0.95\textwidth]{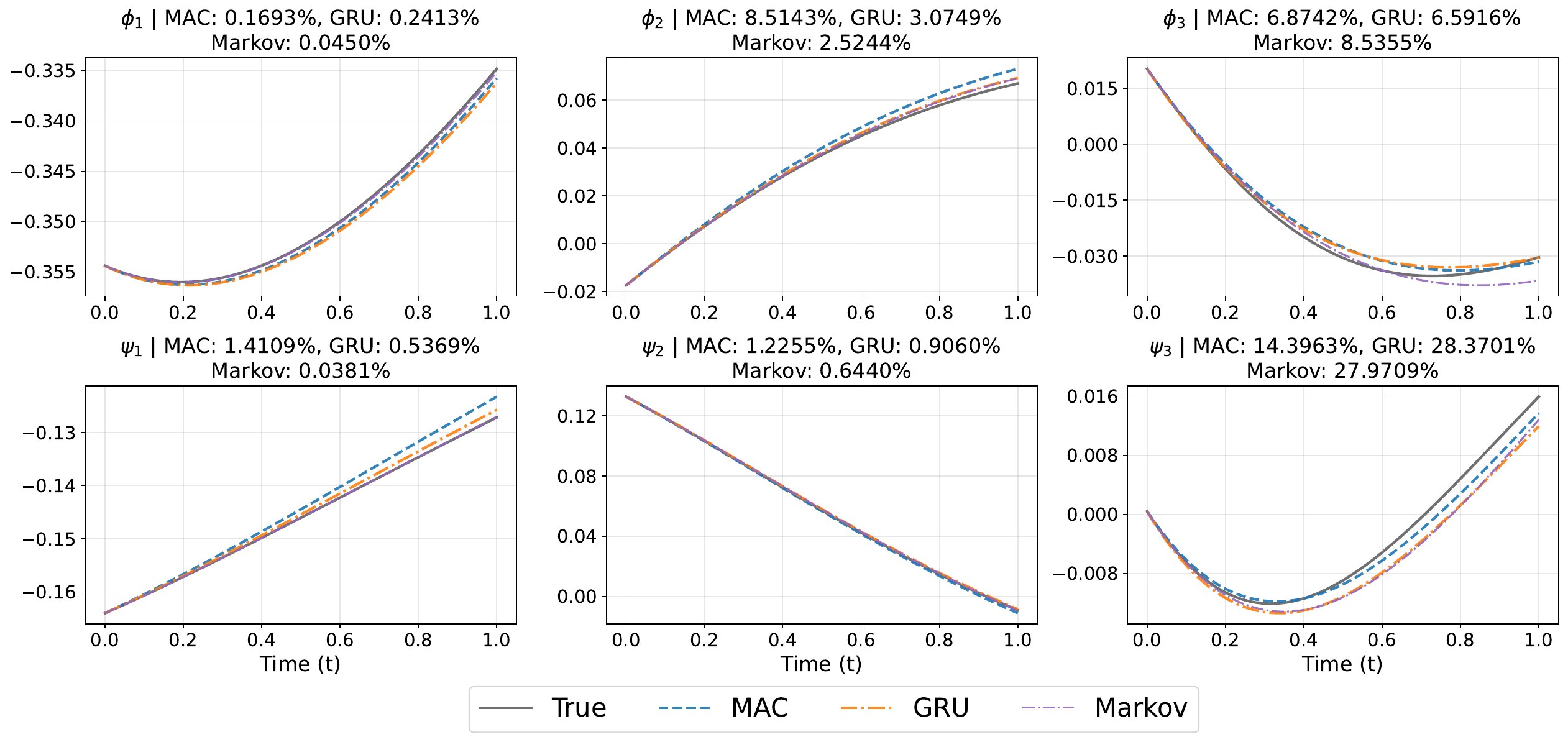}
    \caption{\label{fig_burgers_test_resolved_loss}Comparison of resolved-mode predictions from the MAC model, the GRU-based model, and the Markovian reduced-order model on the same representative test initial condition as in \Cref{fig_burgers_interp_resolved} over the temporal interpolation regime $[0,1]$. For each resolved Fourier mode, the relative $L^2$ error over the time interval $[0,1]$ is also reported. Here the MAC model and the GRU-based model are trained using the alternative single-step resolved-mode loss.}
\end{figure}

\begin{figure}[htbp]
    \centering
    \includegraphics[width=0.95\textwidth]{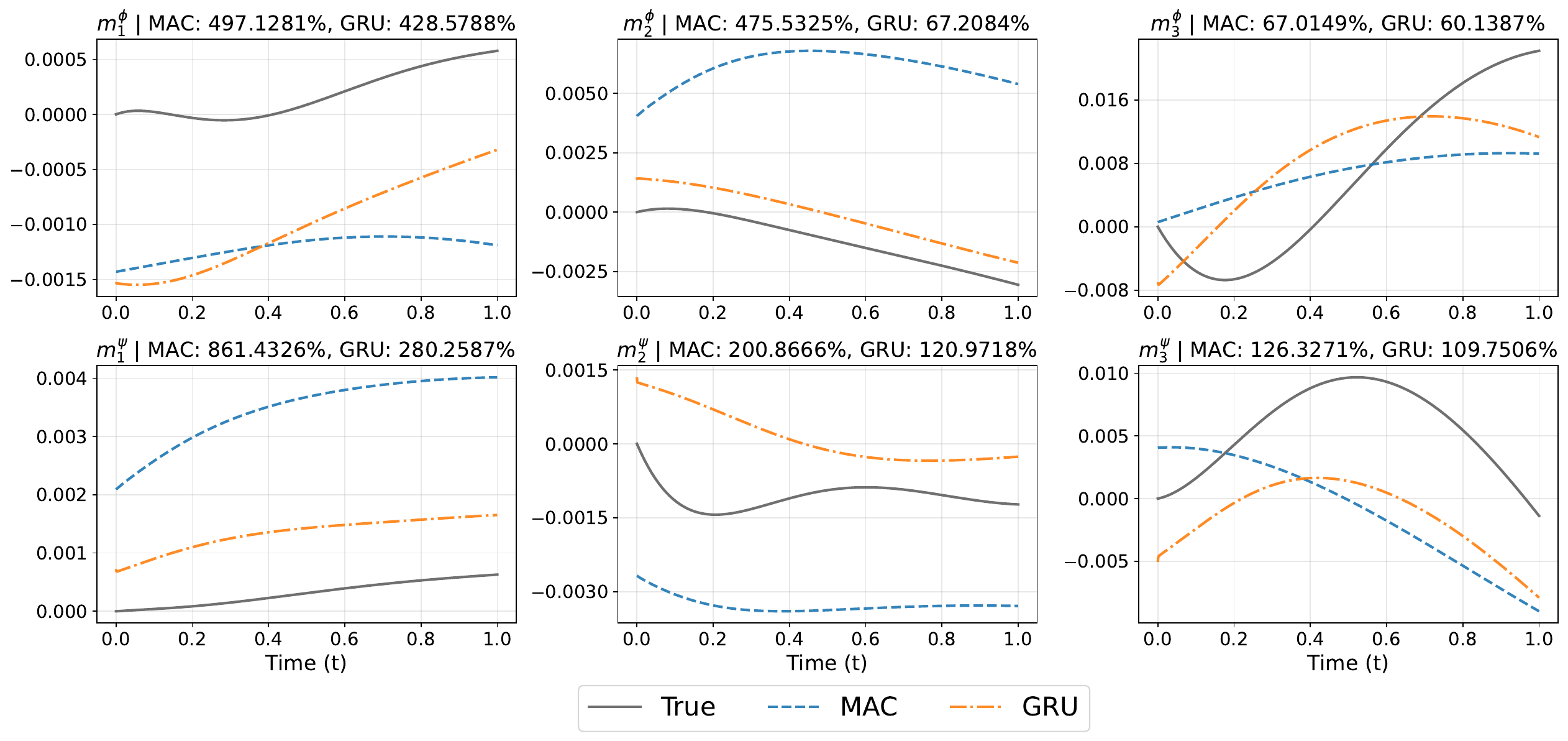}
    \caption{\label{fig_burgers_test_closure_loss}Comparison of the closure-term predictions from the MAC model and the GRU-based model against the true closure terms on the same representative test initial condition as in \Cref{fig_burgers_interp_resolved} over the temporal interpolation regime $[0,1]$. For each closure term, the relative $L^2$ error over the time interval $[0,1]$ is also reported. Here the MAC model and the GRU-based model are trained using the alternative single-step resolved-mode loss.
    }
\end{figure}

In our exploration, we also tested an alternative loss design for Burgers' equation.
In this alternative approach, the closure term predicted by the Mamba-based sequence model is inserted into the resolved dynamics, and the reduced-order system is advanced by one time step.
The predicted resolved Fourier modes at the next time step are then compared with the true resolved modes using a mean squared error loss.
This single-step resolved-mode loss is therefore applied to train the Mamba-based sequence model through the MAC framework.
However, this alternative loss design gives worse prediction accuracy in our experiments.
The results for the predicted resolved modes and closure terms with this alternative loss design, on the same representative case shown in the main text, are presented in \Cref{fig_burgers_test_resolved_loss,fig_burgers_test_closure_loss}.
We can see that, under this alternative loss design, the prediction accuracy on the resolved Fourier modes becomes worse, and the prediction accuracy on the closure terms also deteriorates substantially.

One possible reason is that the single-step resolved-mode loss provides only a weak and indirect training signal for the closure term. For Burgers' equation, the time step is very small, $\Delta t=10^{-4}$, so the contribution of the closure term to a one-step resolved-mode update is also small. As a result, different closure predictions may lead to similar one-step resolved-mode errors, and the single-step resolved-mode loss alone may not strongly constrain the closure term.

Another possible reason is the potential non-uniqueness of the single-step resolved-mode loss objective for Burgers' equation. Since the closure term enters the resolved dynamics through nonlinear spectral interactions and only a subset of Fourier modes is retained, multiple closure configurations may produce similar resolved trajectories. Consequently, the single-step resolved-mode loss may not uniquely identify the true closure term, and the model may instead converge to an inaccurate closure solution that fails to capture the true closure behavior.

Both of these factors are consistent with our empirical observation that training with the single-step resolved-mode loss yields substantially degraded prediction accuracy, as shown in \Cref{fig_burgers_test_resolved_loss,fig_burgers_test_closure_loss}. In contrast, the direct closure loss provides a stronger supervised signal for learning the closure term than the alternative single-step resolved-mode loss.

\subsubsection{Lorenz '96 System}

For the Lorenz '96 system, we use a ten-step resolved-variable rollout loss to train the Mamba-based sequence model.
Unlike the Burgers case, the loss is not computed directly on the predicted closure terms.
Instead, the predicted closure terms are inserted into the resolved dynamics, and the resulting MAC framework is advanced autoregressively for multiple time steps.
The loss is then computed on the predicted resolved variables.

More specifically, each training input is a normalized resolved sequence of length $64$,
\begin{displaymath}
    X_0, X_1, \ldots, X_{63}.
\end{displaymath}
Since the rollout loss uses ten steps, the first $64-10=54$ states,
\begin{displaymath}
    X_0, X_1, \ldots, X_{53},
\end{displaymath}
are used as starting states for the rollout.
Starting from these states, the MAC framework is advanced for ten steps.
At each step, the Mamba-based sequence model predicts the closure terms, which are then inserted into the resolved Lorenz '96 dynamics before the resolved dynamics are advanced by one time step.

The loss is computed only on the advanced resolved variables.
The one-step predictions are compared with
\begin{displaymath}
    X_1, X_2, \ldots, X_{54},
\end{displaymath}
the two-step predictions are compared with
\begin{displaymath}
    X_2, X_3, \ldots, X_{55},
\end{displaymath}
and this continues up to the ten-step predictions, which are compared with
\begin{displaymath}
    X_{10}, X_{11}, \ldots, X_{63}.
\end{displaymath}
The resulting training objective is the mean squared error over these ten sets of predicted resolved variables.

This loss design is motivated by the chaotic nature of the Lorenz '96 system, in which small errors in the closure term can be rapidly amplified through the resolved dynamics, leading to trajectory divergence during autoregressive rollout. A direct closure loss penalizes only the pointwise error in the predicted closure terms, without accounting for how these errors propagate and accumulate in the subsequent resolved-variable evolution. By contrast, the multi-step resolved-variable rollout loss trains the Mamba-based sequence model through the MAC framework so that the predicted closure terms are optimized according to their effect on the resolved-variable trajectory. This makes the training objective more closely aligned with the autoregressive rollout task and helps suppress error accumulation in the chaotic regime.

\begin{figure}[htbp]
    \centering
    \includegraphics[width=15cm]{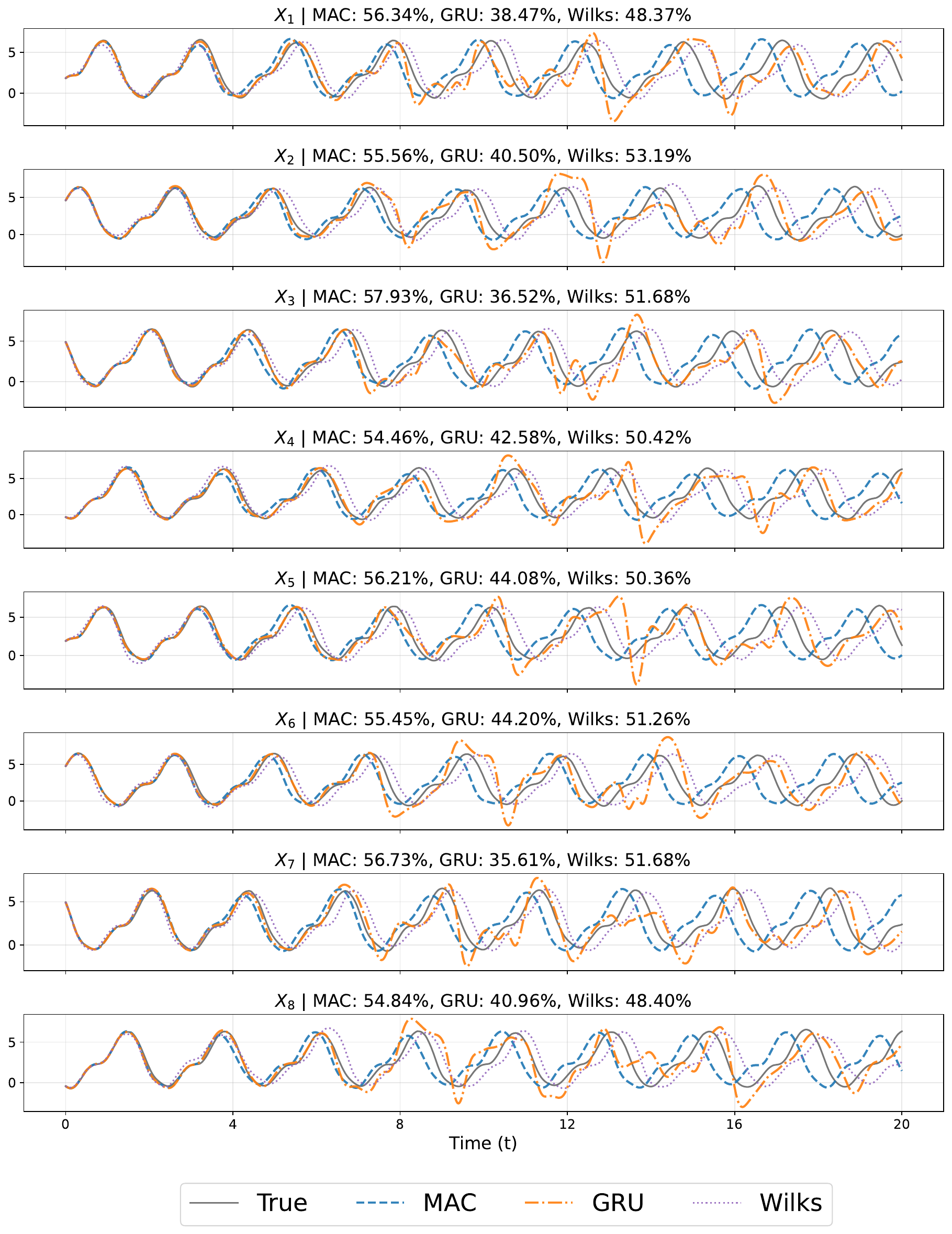}
    \caption{\label{fig_l96_unseen_resolved_loss}
    Comparison of resolved slow-variable predictions from the MAC model, the GRU-based model, and the Wilks method for the same representative unseen initial condition as in \Cref{fig_l96_unseen_resolved} over the time interval $[0,20]$. For each resolved slow variable, the relative $L^2$ error is also reported. Here the MAC model and the GRU-based model are trained using the alternative direct closure loss.
    }
\end{figure}

\begin{figure}[htbp]
    \centering
    \includegraphics[width=15cm]{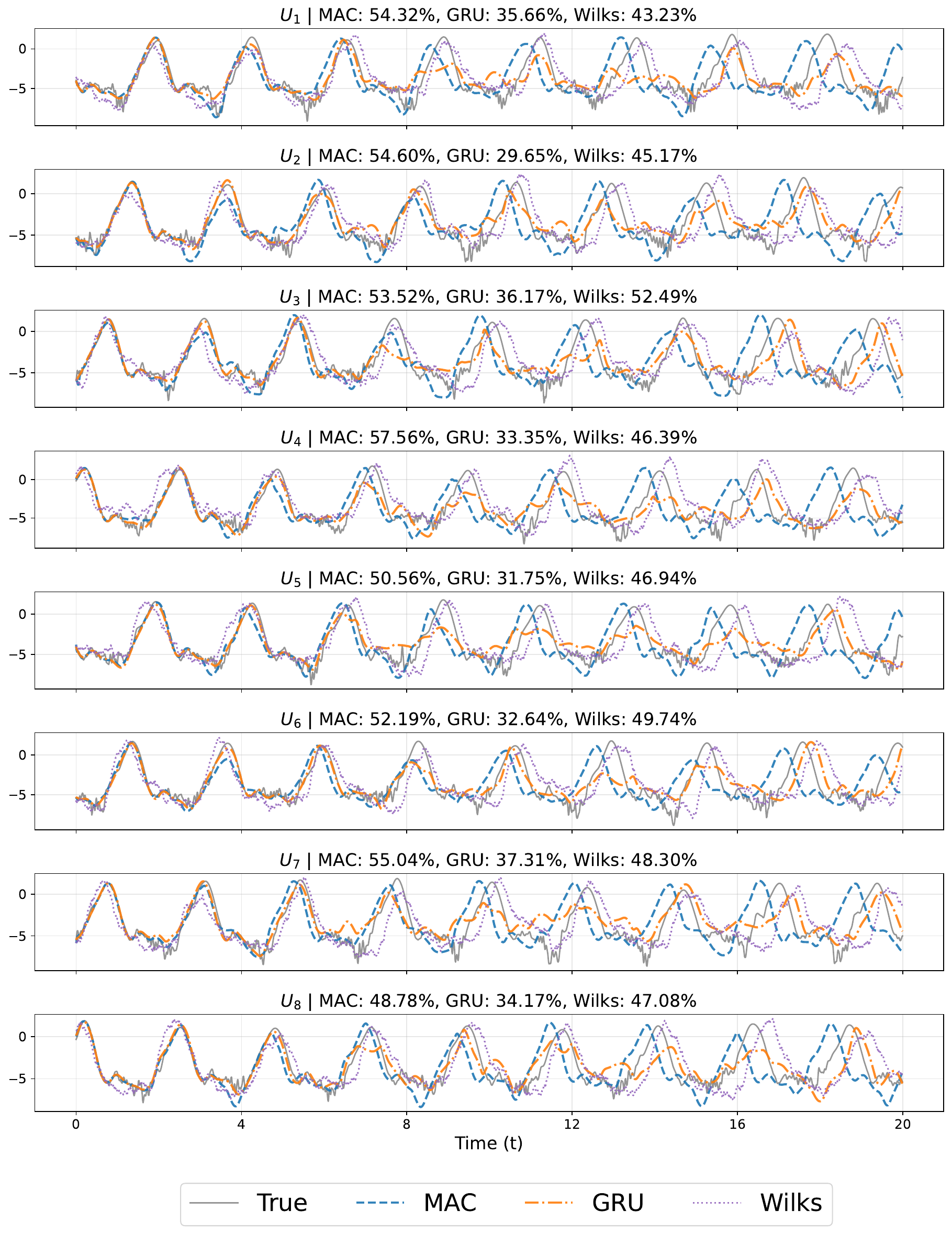}
    \caption{\label{fig_l96_unseen_closure_loss}
    Comparison of closure-term predictions from the MAC model, the GRU-based model, and the Wilks method for the same representative unseen initial condition as in \Cref{fig_l96_unseen_resolved} over the time interval $[0,20]$. For each closure term, the relative $L^2$ error is also reported. Here the MAC model and the GRU-based model are trained using the alternative direct closure loss.
    }
\end{figure}

To compare this choice with a direct closure loss, we also trained the Mamba-based sequence model by minimizing the mean squared error between the predicted closure terms and the true closure terms.
The resulting temporal interpolation results are shown in \Cref{fig_l96_unseen_resolved_loss,fig_l96_unseen_closure_loss}.
In this setting, training with the direct closure loss leads to substantially worse predictions for both the resolved variables and the closure terms than training with the multi-step resolved-variable rollout loss.
This comparison supports the use of the multi-step resolved-variable rollout loss for the Lorenz '96 system.

\subsection{Runge--Kutta Time Stepping with Zero-Order Holding of the Closure}
\label{apdx_rk4_zero_order_holding}

We briefly describe the time-stepping procedure used in the autoregressive rollout experiments. 
Consider an abstract reduced-order system with a learned closure term,
\begin{displaymath}
    \frac{\ud y}{\ud t}=f(y)+\widehat{g}(y),
\end{displaymath}
where $f$ denotes the known reduced-order dynamics and $\widehat{g}$ denotes the closure predicted by the sequence model.

A standard fourth-order Runge--Kutta discretization of this full learned right-hand side would evaluate both $f$ and $\widehat{g}$ at each intermediate stage:
\begin{align*}
    k_1 &= f(y_n)+\widehat{g}(y_n),\\
    k_2 &= f\bigl(y_n+\tfrac{\Delta t}{2}k_1\bigr)
        + \widehat{g}\bigl(y_n+\tfrac{\Delta t}{2}k_1\bigr),\\
    k_3 &= f\bigl(y_n+\tfrac{\Delta t}{2}k_2\bigr)
        + \widehat{g}\bigl(y_n+\tfrac{\Delta t}{2}k_2\bigr),\\
    k_4 &= f(y_n+\Delta t k_3)
        + \widehat{g}(y_n+\Delta t k_3),
\end{align*}
followed by
\begin{displaymath}
    y_{n+1}
    =
    y_n
    +
    \frac{\Delta t}{6}
    \bigl(k_1+2k_2+2k_3+k_4\bigr).
\end{displaymath}
This would require evaluating the learned closure at the RK4 half-step stage states, and also at the final stage state.

In our implementation, we instead use a zero-order holding treatment for the learned closure. 
At the beginning of each time step, the sequence model predicts the closure once,
\begin{displaymath}
    \widehat{g}_n = \widehat{g}(y_n),
\end{displaymath}
and this predicted closure is held fixed throughout the RK4 stages:
\begin{align*}
    k_1 &= f(y_n)+\widehat{g}_n,\\
    k_2 &= f\bigl(y_n+\tfrac{\Delta t}{2}k_1\bigr)+\widehat{g}_n,\\
    k_3 &= f\bigl(y_n+\tfrac{\Delta t}{2}k_2\bigr)+\widehat{g}_n,\\
    k_4 &= f(y_n+\Delta t k_3)+\widehat{g}_n.
\end{align*}
The update to $y_{n+1}$ is then computed using the same RK4 weighted average formula.

This zero-order holding treatment is used to balance numerical accuracy and computational cost. The known reduced-order part $f$ is evaluated at the RK4 stage states, while the learned closure is evaluated only once at the beginning of each time step and held fixed over all four stages, thereby acting as a piecewise constant term over each time step. Compared with advancing the complete reduced system using a first-order Euler method, this treatment retains the stage-wise evaluation of the known dynamics without requiring repeated sequence-model evaluations. A standard stage-wise RK4 treatment of the complete learned right-hand side would require four sequence-model evaluations per time step, substantially increasing the computational cost. This time-stepping procedure is used for all predictive results reported in \Cref{sec_results} and leads to stable autoregressive rollouts. 

\clearpage
\section{Symbols \& Notation}

\Cref{tab_notation} summarizes the main symbols and notation used in this work.
\begin{table}[htbp]
\centering
\caption{Summary of the main symbols and notation used in this work.}
\label{tab_notation}
\begin{tabular}{l p{0.58\linewidth}}
\toprule
\textbf{Notation} & \textbf{Description} \\
\midrule
ROM & Reduced-order model \\
MAC & Mamba-Assisted Closure framework \\
GRU & Gated recurrent unit \\
RK4 & Fourth-order Runge--Kutta method \\
\midrule
$t$ & Time variable \\
$\Delta t$ & Time step used in numerical experiments \\
$u(t,x)$ & Solution of Burgers' equation in physical space \\
$x$ & Spatial coordinate for Burgers' equation on $[0,2\pi)$ \\
$\nu$ & Viscosity coefficient in Burgers' equation \\
$M$ & Number of positively indexed resolved Fourier modes in the Burgers experiments \\
$\theta_k(t)$ & $k$-th Fourier mode of the Burgers solution \\
$\phi_k(t),\psi_k(t)$ & Real and imaginary parts of the Fourier mode $\theta_k(t)$ \\
$\mathcal{M}_k(t)$ & Closure term for the $k$-th resolved Fourier mode in Burgers' equation \\
$m_k^\phi(t),m_k^\psi(t)$ & Real and imaginary parts of the Burgers closure term\\
\midrule
$X_k(t)$ & $k$-th slow variable in the two-scale Lorenz '96 system \\
$Y_j(t)$ & $j$-th fast variable in the two-scale Lorenz '96 system \\
$U_k(t)$ & Closure term associated with the $k$-th slow variable in the Lorenz '96 system \\
$N$ & Number of slow variables in the Lorenz '96 system \\
$J$ & Number of fast variables coupled to each slow variable \\
\bottomrule
\end{tabular}
\end{table} 

\end{appendices}
\clearpage
\section*{References}
\addcontentsline{toc}{section}{\protect\numberline {}{References}}
\printbibliography[heading=none]  

\end{document}